\pdfoutput=1
\documentclass[10pt,twocolumn,letterpaper]{article}

\usepackage{cvpr}
\usepackage{times}
\usepackage{epsfig}
\usepackage{graphicx}
\usepackage{amsmath}
\usepackage{amssymb}
\usepackage{float}
\usepackage{amsmath,diagbox,tabu,stackengine}
\usepackage[]{algorithm2e}
\usepackage[utf8]{inputenc} 
\usepackage[T1]{fontenc}    
\usepackage{url}            
\usepackage{booktabs}       
\usepackage{amsfonts}       
\usepackage{nicefrac}       
\usepackage{microtype}      

\usepackage{caption}
\usepackage{multirow}
\usepackage{xspace}
\usepackage{mathtools}
\usepackage{wasysym}
\usepackage{marvosym}
\usepackage[table,xcdraw]{xcolor}
\usepackage{pdfcomment}
\usepackage{lipsum}
\usepackage{subcaption}

\usepackage{hyperref}
\hypersetup{breaklinks=true,colorlinks}

\cvprfinalcopy 

\def\cvprPaperID{3409} 
\def\httilde{\mbox{\tt\raisebox{-.5ex}{\symbol{126}}}}

\ifcvprfinal\fi

\makeatother

\title{\vspace{-0mm}Efficient Interactive Annotation of Segmentation Datasets with Polygon-RNN++}

\author{David Acuna$^{1,3,}$\thanks{authors contributed equally}\hspace{1cm} Huan Ling$^{1,2}$\footnotemark[1] \hspace{1cm} Amlan Kar$^{1,2,}$\footnotemark[1] \hspace{1cm} Sanja Fidler$^{1,2}$\\
$^1$University of Toronto \hspace{1em} $^2$Vector Institute \hspace{1em} $^3$NVIDIA\thanks{work done when D.A. was at UofT} \\
{\tt\small \{davidj, linghuan, amlan, fidler\}@cs.toronto.edu}}

\begin{document}

\maketitle

\begin{abstract}
Manually labeling datasets with object masks is extremely time consuming. In this work, we follow the idea of Polygon-RNN~\cite{polyrnn} to produce polygonal annotations of objects interactively using humans-in-the-loop. We introduce several important improvements to the model: 1) we design a new CNN encoder architecture, 2) show how to effectively train the model with Reinforcement Learning, and 3) significantly increase the output resolution using a Graph Neural Network, allowing the model to accurately annotate high-resolution objects in images. Extensive evaluation on the Cityscapes dataset~\cite{cityscapes} shows that our model, which we refer to as Polygon-RNN++, significantly outperforms the original model in both automatic (10\% absolute and 16\% relative improvement in mean IoU) and interactive modes (requiring 50\% fewer clicks by annotators). We further analyze the cross-domain scenario in which our model is trained on one dataset, and used out of the box on datasets from varying domains. The results show that Polygon-RNN++ exhibits powerful generalization capabilities, achieving significant improvements over existing pixel-wise methods. 
Using simple online fine-tuning we further achieve a high reduction in annotation time for new datasets, moving a step closer towards an interactive annotation tool to be used in practice.
\end{abstract}

\vspace{-8mm}
\section{Introduction}
\label{sec:intro}

Detailed reasoning about structures in images is a necessity for numerous computer vision applications. For example, it is crucial in the domain of autonomous driving to localize and outline all cars, pedestrians, and miscellaneous static and dynamic objects~\cite{DWT17,SGN17,maskrcnn}. For mapping, there is a need to obtain detailed footprints of buildings and roads from aerial/satellite imagery~\cite{TCity2017}, while medical/healthcare domains require automatic methods to precisely outline cells, tissues and other relevant structures~\cite{medical1,medical2}. 

Neural networks have proven to be an effective way of inferring semantic~\cite{JayICLR2015,LongCVPR2015} and object instance segmentation information~\cite{maskrcnn,SGN17} in challenging imagery. It is well known that the amount and variety of data that the networks see during training drastically affects their performance at run time. Collecting ground truth instance masks, however, is an extremely time consuming task, typically requiring human annotators to spend 20-30 seconds per object in an image. 

To this end, in~\cite{polyrnn}, the authors introduced Polygon-RNN, a conceptual model for semi-automatic and interactive labeling to help speed up object annotation. Instead of producing pixel-wise segmentation of an object as in existing interactive tools such as Grabcut~\cite{Rother2004SIGGRAPH},~\cite{polyrnn} predicts the vertices of a polygon that outlines the object. The benefits of using a polygon representation are three-fold, {\bf 1)} it is sparse (only a few vertices represent regions with a large number of pixels), {\bf 2)} it is easy for an annotator to interact with, and {\bf 3)} it allows for efficient interaction, typically requiring only a few corrections from the annotator~\cite{polyrnn}.  
 Using their model, the authors have shown high annotation speed-ups on two autonomous driving datasets~\cite{cityscapes,kitti}.

\begin{figure}[t!]
\includegraphics[width=\linewidth,trim=0 5 0 0,clip]{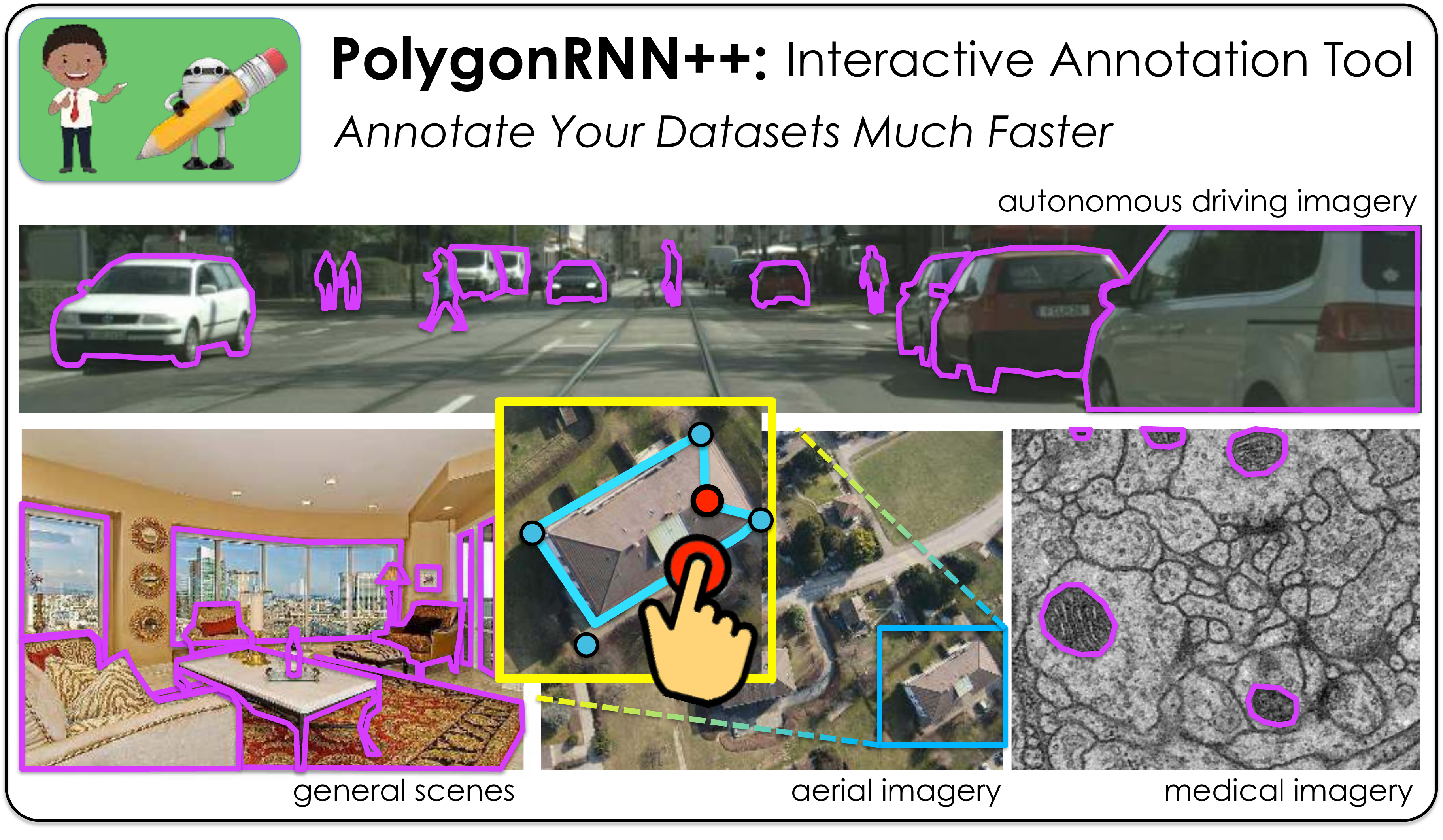} 
\vspace{-7mm}
\caption{\small We introduce Polygon-RNN++, an interactive object annotation tool. We make several advances over~\cite{polyrnn}, allowing us to annotate objects faster and more accurately. Furthermore, we exploit a simple online fine-tuning method to adapt our model from one dataset to efficiently annotate novel, out-of-domain datasets.}
\label{fig:intro}
\vspace{-2mm}
\end{figure}

In this work, we introduce several improvements to the Polygon-RNN model. In particular, we {\bf 1)} make a few changes to the neural network architecture, {\bf 2)} propose a better learning algorithm to train the model using reinforcement learning, and {\bf 3)} show how to significantly increase the output resolution of the polygon (one of the main limitations of the original model) using a Graph Neural Network~\cite{scarselli2009graph,GGNN}. We analyze the robustness of our approach to noise, and its generalization capabilities to out-of-domain imagery. 

In the fully automatic mode (no annotator in the loop), our model achieves significant improvements over the original approach, outperforming it by 10\% mean IoU on the Cityscapes dataset~\cite{cityscapes}. %
In interactive mode, our approach requires 50\% fewer clicks as compared to \cite{polyrnn}. To demonstrate generalization, we use a model trained on the Cityscapes dataset to annotate a subset of a scene parsing dataset~\cite{ade20k}, aerial imagery~\cite{Sun14}, and two medical datasets~\cite{medical1,medical2}. The model significantly outperforms strong pixel-wise labeling baselines, showcasing that it inherently learns to follow object boundaries, thus generalizing better. We further show that a simple online fine-tuning approach achieves high annotation speed-ups on out-of-domain dataset annotation.
Our model is released online: {\small\url{http://www.cs.toronto.edu/polyrnn/}}. 

\section{Related Work}
\label{sec:related}

%

{\bf Interactive annotation}.  Since object instance segmentation is time consuming to annotate manually, several works have aimed at speeding up this process using interactive techniques. In seminal work,~\cite{Boykov2001ICCV} used scribbles to model the appearance of foreground/background, and performed segmentation via graph-cuts~\cite{Boykov2004PAMI}. This idea was extended by~\cite{Nagaraja} to use multiple scribbles on both the object and background, and was demonstrated in annotating objects in videos. 
GrabCut~\cite{Rother2004SIGGRAPH}  exploited 2D bounding boxes provided by the annotator, and performed pixel-wise foreground/background labeling  using EM.~\cite{DeepCut} combined GrabCut with CNNs to annotate structures in medical imagery. 
Most of these works operate on the pixel level, and typically have difficulties in cases where foreground and background have similar color. 

In~\cite{polyrnn}, the authors used polygons instead. The main power of using such a representation is that it is sparse; only a few vertices of a polygon represent large image regions. This allows the user to easily introduce corrections, by simply moving the wrong vertices. An RNN also effectively captures typical shapes of objects as it forms a non-linear sequential representation of shape. This is particularly important in ambiguous regions, ie shadows and saturation, where boundaries cannot be observed. We follow this line of work, and introduce several important modifications to the architecture and training. Furthermore, the original model was only able to make prediction at a low resolution ($28\times 28$), thus producing blocky polygons for large objects. Our model significantly increases the output resolution ($112\times 112$).  


{\bf Object instance segmentation}. Most approaches to object instance segmentation~\cite{IIS16,torr16,ZhangICCV15,ZhangCVPR16,deepmask,sharpmask,maskrcnn,DWT17,SGN17} operate on the pixel-level. Many rely on object detection, and use a convnet over a box proposal to perform the labeling~\cite{deepmask,sharpmask,maskrcnn}. In~\cite{ZhangCVPR12,Sun14}, the authors produce a polygon around an object. These approaches first detect boundary fragments, followed by finding an optimal cycle  linking the boundaries into object regions.~\cite{Duan16} produce superpixels in the form of small polygons which are further combined into an object. %
Here, as in~\cite{polyrnn} we use neural networks to produce polygons, and in particular tackle the interactive labeling scenario which has not been explored in these works.

\section{Polygon-RNN++}
In this section, we introduce Polygon-RNN++. Following~\cite{polyrnn}, our model expects an annotator to provide a bbox around the object of interest. We extract an image crop enclosed by the 15\% enlarged box. We use a CNN+RNN architecture as in~\cite{polyrnn}, with a CNN serving as an image feature extractor, and the RNN decoding one polygon vertex at a time. Output vertices are represented as locations in a grid.

\begin{figure*}[htb!]
\vspace{-3mm}
\begin{minipage}{0.58\linewidth}
\includegraphics[width=1\linewidth,trim=0 70 0 30,clip]{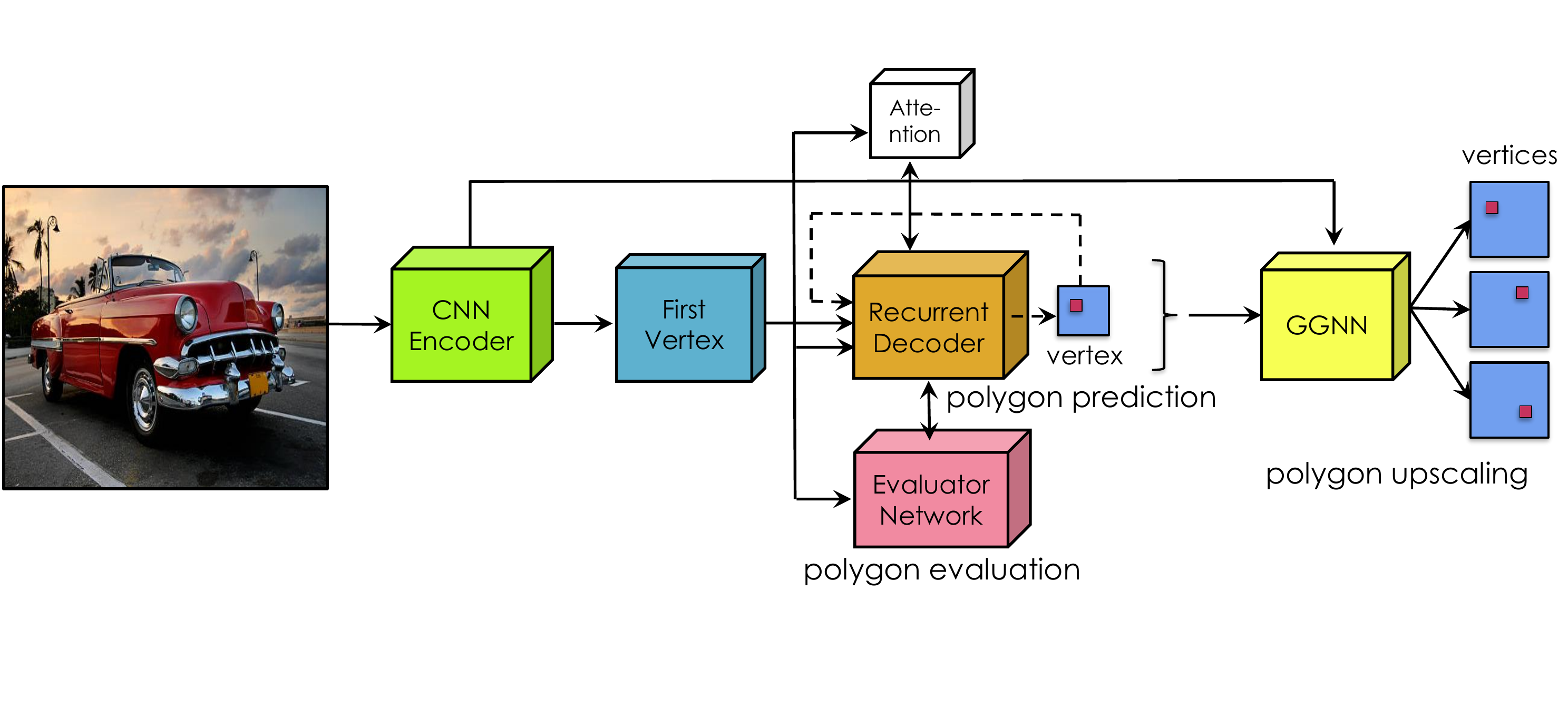} 
\vspace{-6.5mm}
\caption{Polygon-RNN++ model \footnotesize{(figures best viewed in color)}}
\label{fig:full_model}
\end{minipage}
\hspace{10mm}
\begin{minipage}{0.34\linewidth}
\vspace{1mm}
\includegraphics[width=1\linewidth,trim=2 10 2 0,clip]{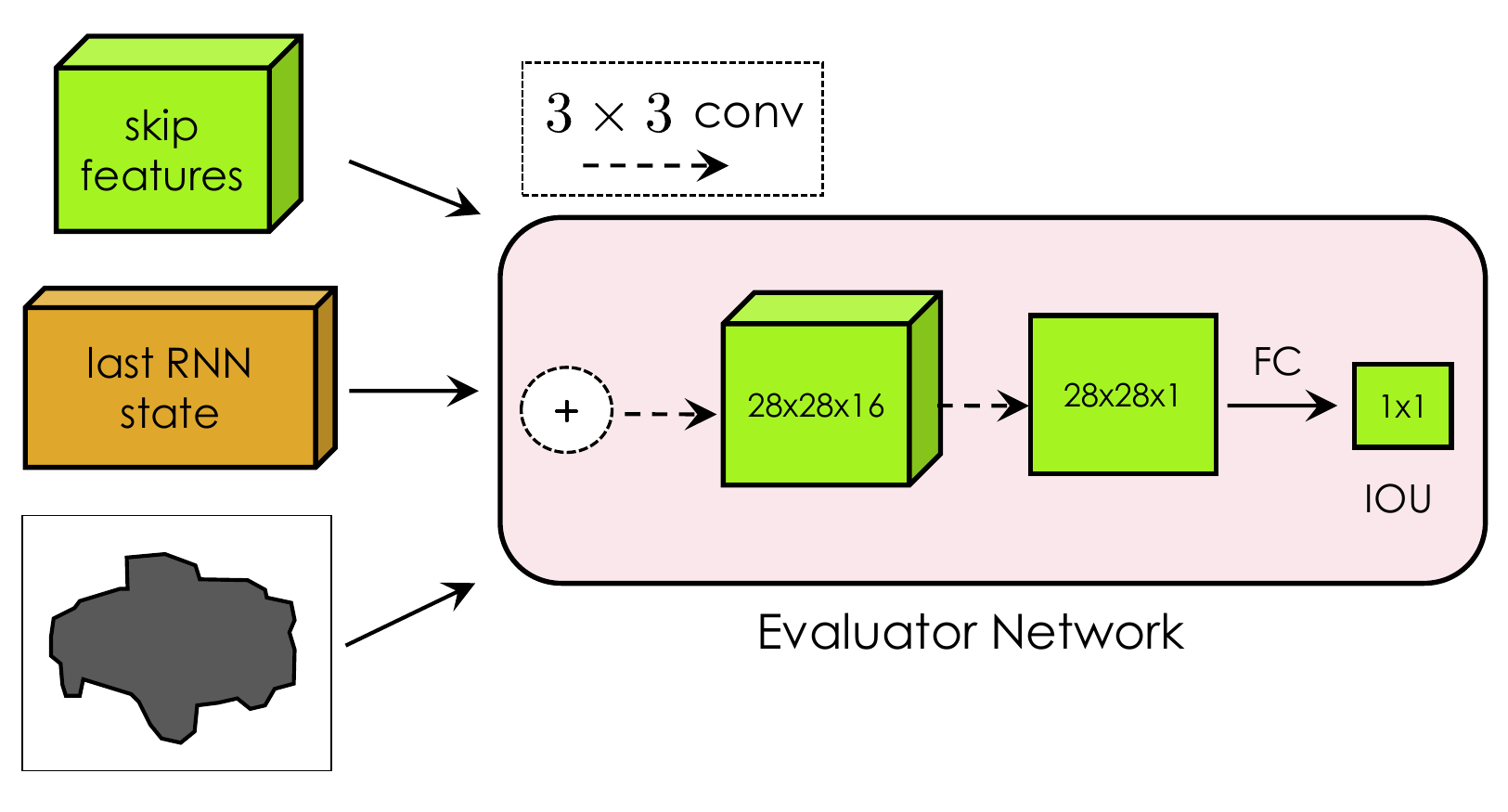} 
\vspace{-5.5mm}
\caption{Evaluator Network predicting the quality of a polygon output by the RNN decoder}
\label{fig:evaluator_network}
\end{minipage}
\vspace{-4mm}
\end{figure*}

The full model is depicted in Fig.~\ref{fig:full_model}. Our redesigned encoder produces image features that are used to predict the first vertex. The first vertex and the image features are then fed to the recurrent decoder. Our RNN exploits visual attention at each time step to produce polygon vertices. A learned evaluator network selects the best polygon from a set of candidates proposed by the decoder. Finally, a graph neural network re-adjusts polygons, augmented with additional vertices, at a higher resolution.

This model naturally incorporates a human in the loop, allowing the annotator to correct an erroneously predicted vertex. This vertex is then fed back to the model, helping the model to correct its prediction in the next time steps. 

\subsection{Residual Encoder with Skip Connections}
Most networks perform repeated down-sampling operations at consecutive layers of a CNN, which impacts the effective output resolution in tasks such as image segmentation~\cite{JayICLR2015,pohlen2016full}. In order to alleviate this issue, we follow \cite{deeplab} and modify the ResNet-50 architecture~\cite{he15deepresidual} by reducing the stride of the network and introducing dilation factors. This allows us to increase the resolution of the output feature map without reducing the receptive field of individual neurons. We also remove the original average pooling and FC layers.

\begin{figure}[t!]
\vspace{-1mm}
\includegraphics[width=\linewidth]{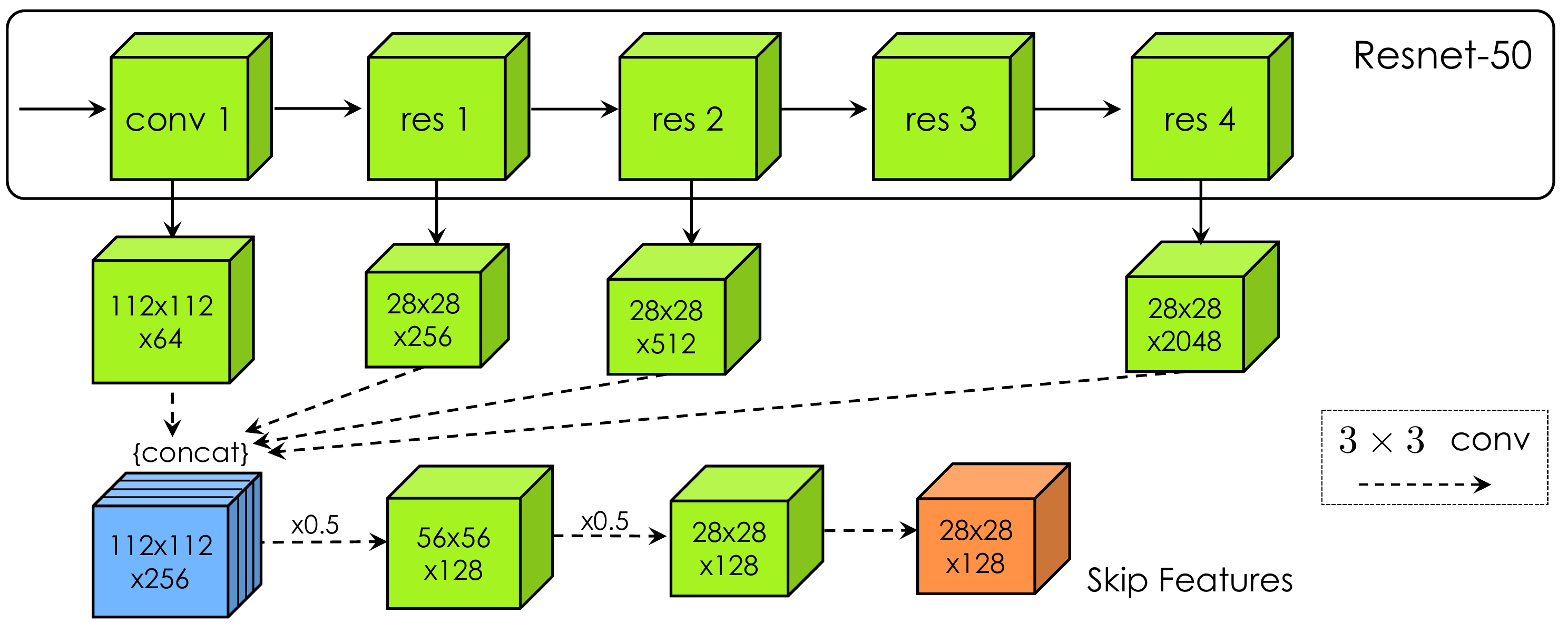}
\vspace{-7mm}
\caption{Residual Encoder architecture. Blue tensor is fed to GNN, while the orange tensor is input to the RNN decoder. }
\label{fig:enc_arch}
\vspace{-2mm}
\end{figure}

We further add a skip-layer architecture~\cite{LongCVPR2015, pspnet} which aims to capture both, low-level details such as edges and corners, as well as high-level semantic information.
In~\cite{polyrnn}, the authors perform down-sampling in the skip-layer architecture, built on top of VGG, before concatenating the features from different layers. Instead, we concatenate all the outputs of the skip layers at the highest possible resolution, and use a combination of conv layers and max-pooling operations to obtain the final feature map. We employ conv filters with a kernel size of $3 \times 3$, batch normalization~\cite{ioffe2015batch} and ReLU non-linearities. In cases where the skip-connections have different spatial dimensions, 
we use bilinear upsampling before concatenation. The architecture is shown in Fig.~\ref{fig:enc_arch}. We refer to the final feature map as the \emph{skip features}. 

\vspace{-0mm}
\subsection{Recurrent Decoder}
As in~\cite{polyrnn}, we use a Recurrent Neural Network to model the sequence of 2D vertices of the polygon outlining an object. In line with previous work, we also found that the use of Convolutional LSTM~\cite{convlstm} is essential: {\bf 1)} to preserve spatial information and {\bf 2)} to reduce the number of parameters to be learned. In our RNN, we further add an attention mechanism, as well as predict the first vertex within the same network (unlike~\cite{polyrnn} which has two separate networks).

We use a two-layer ConvLTSM with a $3 \times 3$ kernel with 64 and 16 channels, respectively. 
We apply batch norm~\cite{ioffe2015batch} at each time step, without sharing mean/variance estimates across time steps. We represent our output at time step $t$ as a one-hot encoding of $(D \times D) + 1$ elements, where $D$ is the resolution at which we predict. In our experiments, $D$ is set to 28. The first $D \times D$ dimensions represent the possible vertex positions and the last dimension corresponds to the end-of-seq token that signals that the polygon is closed. 

\vspace{-3mm}
\paragraph*{Attention Weighted Features:}
In our RNN, we exploit a mechanism akin to attention. In particular, at time step $t$, we compute the weighted feature map as,
\begin{equation}
\begin{aligned}
& {\alpha}_{t} = \mathrm{softmax}(f_{att}(x, f_1(h_{1,t-1}) , f_2(h_{2,t-1}) ))\\
& F_t = x \circ \alpha_{t}
 \end{aligned}
\end{equation}
where $\circ$ is the Hadamard product, $x$ is the \emph{skip feature} tensor, and $h_{1,t}$, $h_{2,t}$ are the hidden state tensors from the two-layer ConvLSTM. Here, $f_{1}$ and $f_{2}$ map $h_{1,t}$ and $h_{2,t}$ to $\mathbb{R}^{D\times D \times 128}$ using one fully-connected layer. $f_{att}$ takes the sum of its inputs and maps it to $D \times D$ through a fully connected layer, giving one ``attention'' weight per location. 

Intuitively, we use the previous RNN hidden state to gate certain locations in the image feature map, allowing the RNN to focus only on the relevant information in the next time step. The gated feature map $F_t$ is then concatenated with one-hot encodings of the two previous vertices $y_{t-1}, y_{t-2}$ and the first vertex $y_0$, and passed to the RNN at time step $t$.

\vspace{-4mm}
\paragraph*{First Vertex:} Given a previous vertex and an implicit direction, the next vertex of a polygon is always uniquely defined, except for the first vertex. To tackle this problem, the authors in~\cite{polyrnn} treated the first vertex as a special case and used an additional architecture (trained separately) to predict it. In our model, we add another branch from the skip-layer architecture,  constituting of two layers each of dimension $D\times D$. Following~\cite{polyrnn}, the first layer predicts edges, while the second predicts the vertices of the polygon. At test time, the first vertex is sampled from the final layer of this branch.

\subsection{Training using Reinforcement Learning }
In~\cite{polyrnn}, the authors trained the model using the cross-entropy loss at each time step. However, such training has two major limitations: {\bf 1)} MLE over-penalizes the model (for example when the predicted vertex is on an edge of the GT polygon but is not one of the GT vertices), and {\bf 2)} it optimizes a metric that is very different from the final evaluation metric (i.e. IoU).
Further, the model in~\cite{polyrnn} was trained following a typical training regime where the GT vertex is fed to the next time step instead of the model's prediction. This training regime, called teacher forcing creates a mismatch between training and testing known as the exposure bias problem~\cite{ranzato2015sequence}.

In order to mitigate these problems, we only use MLE training as an initialization stage. We then reformulate the polygon prediction task as a reinforcement learning problem and fine-tune the network using RL. During this phase, we let the network discover policies that optimize the desirable, yet non-differentiable evaluation metric (IoU) while also exposing it to its own predictions during training. 

\vspace{-3mm}
\subsubsection{Problem formulation}
\vspace{-2mm}
We view our recurrent decoder as a sequential decision making agent. The parameters $\theta$ of our encoder-decoder architecture  define its policy $p_\theta$ for selecting the next vertex $v_t$. At the end of the sequence, we obtain a reward $r$. We compute our reward as the IoU between the mask enclosed by the generated polygon and the ground-truth mask $m$. To maximize the expected reward, our loss function becomes
\begin{align}
L(\theta)= -\mathbb{E}_{v^s \sim p_\theta}[r(v^s,m)]
\end{align}
where $v^s = (v_1^s,...,v_T^s)$, and $v_t^s$ is the vertex sampled from the model at time $t$. Here, $r=\mathrm{IoU}(\mathrm{mask}(v^s),m)$. 
\vspace{-2mm}
\subsubsection{Self-Critical Training with Policy Gradients}
\vspace{-2mm}
Using the REINFORCE trick~\cite{RL1992} to compute the gradients of the expectation, we have
\begin{align}
\nabla L(\theta)= -\mathbb{E}_{v^s \sim p_\theta}[r(v_s,m) \nabla \log p_\theta(v^s) ]
\end{align}
In practice, the expected gradient is computed using simple Monte-Carlo sampling with a single sample. This procedure is known to exhibit high variance and is highly unstable without proper context-dependent normalization. A natural way to deal with this is to use a learned \textit{baseline} which is subtracted from the reward. In this work, we follow the self-critical method~\cite{rennie2016self} and use the test-time inference reward of our model as the baseline. Accordingly, we reformulate the gradient of our loss function to be
\begin{align}
\nabla L(\theta) &= - [(r(v^s,m)- r(\hat v^s,m)) \nabla \log p_\theta(v^s) ]
\end{align}
where $r(\hat v^s,m)$ is the reward obtained by the model using greedy decoding.
To control the level of randomness in the vertices explored by the model, we introduce a temperature parameter $\tau$ in the softmax of the policy. This ensures that the sampled vertices lead to well behaved polygons. In our experiments, we set $\tau=0.6$.
%


%

\subsection{Evaluator Network}

Smart choice of the first vertex is crucial as it biases the initial predictions of the RNN, when the model does not have a strong history to reason about the object to annotate. This is particularly important in cases of occluding objects. It is desirable for the first vertex to be far from the occlusion boundaries so that the model follows the object of interest. 
In RNNs, beam search is typically used to prune off improbable sequences (such as when the model starts to follow an occluding object). However, since classical beam search uses log probabilities to evaluate beams, it does not directly apply to our model which aims to optimize IoU. A point on an occlusion boundary generally exhibits a strong edge and thus would have a high log probability during prediction, reducing the chances of it being pruned by beam search.

In order to solve this problem, we propose to use an \emph{evaluator network} at inference time, aiming to effectively choose among multiple candidate polygons. Our evaluator network takes as input the \emph{skip features}, the last state tensor of the ConvLSTM, and the predicted polygon, and tries to estimate its quality by predicting its IoU with GT.
The  network has two $3 \times 3$ convolutional layers followed by a FC layer, forming another branch in the model. Fig.~\ref{fig:evaluator_network} depicts its architecture. While the full model can be trained end-to-end during the RL step, we choose to train the evaluator network separately after the RL fine-tuning has converged.
 
During training, we  minimize the mean squared error
\begin{align}
L(\phi)= [p(\phi,v^s)-\mathrm{IoU}(m_{v^{s}},m)]^2
\end{align}
where $p$ is the network's predicted IoU, $m_{v^s}$ is the mask for the sampled vertices and $m$ is the ground-truth mask. To ensure diversity in the vertices seen, we sample polygons with $\tau=0.3$.
We emphasize that we do not use this network as a baseline estimator during the RL training step since we found that the self-critical method produced better results.

\vspace{-3.5mm}
\paragraph*{Inference:} At test time, we take $K$ top scoring first vertex predictions. For each of these, we generate polygons via classical beam-search (using log prob with a beam-width $B$). This yields $K$ different polygons, one for each first vertex candidate. We use the evaluator network to choose the best polygon. In our experiments, we use $K=5$. While one could use the evaluator network instead of beam-search at each time step, this would lead to impractically long inference times. Our faster full model (using $B=K=1$) runs at 
295ms per object instance on a Titan XP.

\vspace{-3.5mm}
\paragraph*{Annotator in the Loop:} We follow the same protocol as in~\cite{polyrnn}, where the annotator corrects the vertices in sequential order. Each correction is then fed back to the model, which re-predicts the rest of the polygon. 

\subsection{Upscaling with a Graph Neural Network}
The model described above produces polygons at a resolution of $D \times D$, where we set $D$ to be 28 to satisfy memory bounds and to keep the cardinality of the output space amenable. In this section, we exploit a Gated Graph Neural Network (GGNN)~\cite{GGNN}, in order to generate polygons at a much higher resolution. GNN has been proven efficient for semantic segmentation~\cite{3dggnnICCV17}, where it was used at pixel-level.

Note that when training the RNN decoder, the GT polygons are simplified at their target resolution (co-linear vertices are removed) to alleviate the ambiguity of the prediction task. 
Thus, at a higher resolution, the object may have additional vertices, thus changing the topology of the polygon. %

\begin{figure}[t!]
\vspace{-2mm}
\begin{minipage}{1\linewidth}
 \includegraphics[width=1\linewidth,trim=0 0 0 0,clip]{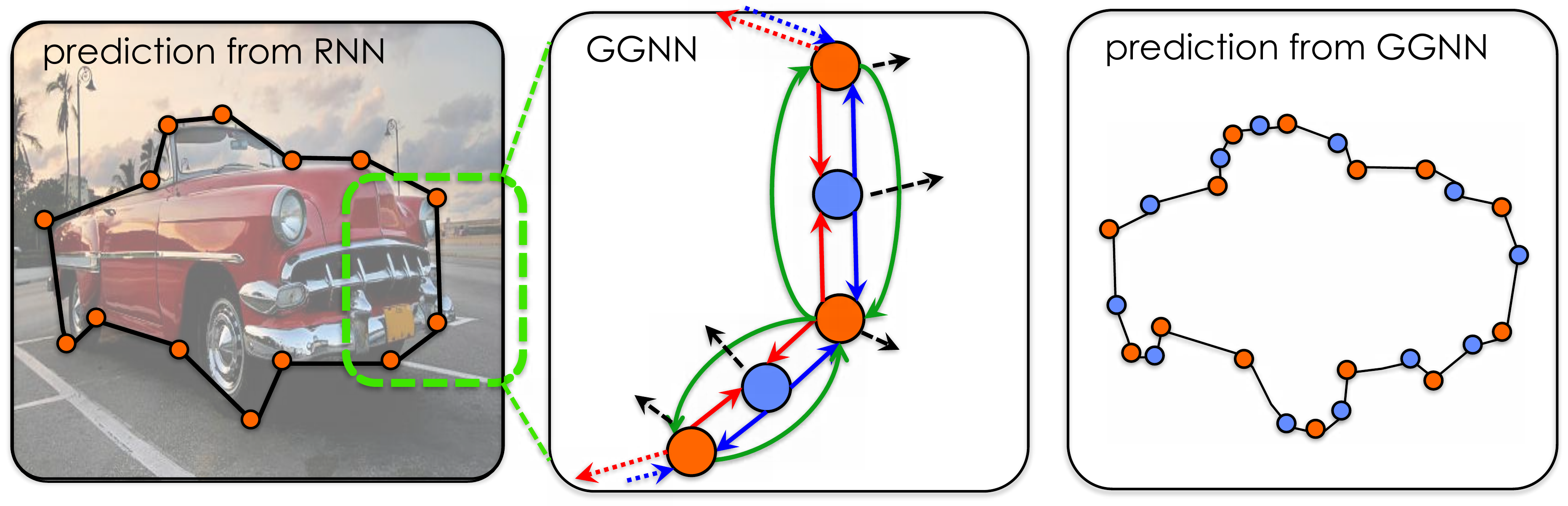}
  \end{minipage}
    \vspace{-3mm}
  \caption{GGNN model: We take predicted polygon from RNN (orange vertices), and add midpoints (in blue) between every pair of consecutive vertices (orange). Our GGNN has three types of edges (red, blue, green), each having its own weights for message propagation. Black dashed arrows pointing out of the nodes (middle diagram) indicate that the GGNN aims to predict the relative location for each of the nodes (vertices), after completing propagation. Right is the high resolution polygon output by the GGNN.}
  \label{fig:PolygonGGNN}
  \vspace{-2mm}
\end{figure}

Our upscaling model takes as input the sequence of vertices generated by the RNN decoder. We treat these vertices as nodes in a graph. %
To model finer details at a higher resolution, we add a node in between two consecutive nodes, with its location being in the middle of their corresponding edge. We also connect the last and the first vertex, effectively converting the sequence into a cycle. We connect neighboring nodes using $3$ different types of edges, as shown in Fig.~\ref{fig:PolygonGGNN}. 

GGNN defines a propagation model that extends RNNs to arbitrary graphs, effectively propagating information between nodes, before producing an output at each node.  Here, we aim to predict the relative offset of each node (vertex) at a higher resolution. 
The model is visualized in Fig.~\ref{fig:PolygonGGNN}.

\vspace{-3mm}
\paragraph{Gated Graph Neural Network:}
For completeness, we briefly summarize the GGNN model~\cite{GGNN}. GGNN uses a graph $\{V, E\}$, where V and E are the sets of nodes and edges, respectively. It consists of a propagation model performing message passing in the graph, and an output model for prediction tasks. We represent the initial state of a node $v$ as $x_v$ and the hidden state of node $v$ at time step $t$ as $h_v^t$. 
The basic recurrence of the propagation model is
\begin{equation}
\begin{aligned}
& h_v^{0} = [x_v^{\top}, 0]^{\top}\\
& a_v^{t} = A_{v:}^{\top}\,[{h_1^{t-1}}^{\top}, ... , {h_{|V|}^{t-1}}^{\top} ]^{\top} + b\\
& h_v^{t} = f_{GRU}(h_v^{t-1}, a_v^{t})
 \end{aligned}
\end{equation}
The matrix $A\in\mathbb{R}^{|V| \times 2N|V|}$ determines how the nodes in the graph communicate with each other, where $N$ represents the number of different edge types.
Messages are propagated for  $T$ steps. The output for node $v$ is then defined as

\begin{equation}
\begin{aligned}
& h_v = \tan(f_1(h_v^{T}))\\
& \mathrm{out}_v = f_2(h_v)\\
\end{aligned}
\end{equation}
Here, $f_1$ and $f_2$ are MLP, and $\mathrm{out}_v$ is $v$'s desired output. 

\vspace{-3mm}
\paragraph{PolygonRNN++ with GGNN:} To get observations for our GGNN model, we add another branch on top of our skip-layer architecture, specifically, from the $112\times112\times 256$ feature map (marked in blue in Fig.~\ref{fig:enc_arch}). We exploit a conv layer with 256 filters of size $15 \times 15$, giving us a feature map of size $112\times112\times256$. %
For each node $v$ in the graph, we extract a $S \times S$ patch around the scaled $(v_x,v_y)$ location, giving us the observation vector $x_v$. 
After propagation, we predict the output of a node $v$ as a location in a $D' \times D'$ spatial grid. We make this grid relative to the location $(v_x,v_y)$, rendering the prediction task to be a relative displacement with respect to its initial position. 
This prediction is treated as a classification task and the model is trained with the cross entropy loss. In particular, in order to train our model, we first take predictions from the RNN decoder, and correct a wrong prediction if it deviates from the ground-truth vertex by more than a threshold. The targets for training our GGNN are then the relative displacements of each of these vertices with respect to their corresponding ground-truth vertices.

\vspace{-4mm}
\paragraph{Implementation details:} We set $S$ to 1 and $D'$ to 112. While our model supports much higher output resolutions, we found that larger $D'$ did not improve results. The hidden state of the GRU in the GGNN has 256 dimensions. We use $T=5$ propagation steps.  In the output model, $f_1$ is a $256 \times 256$ FC layer and $f_2$ is a $256 \times 15 \times 15$ MLP. In training, we take the predictions from the RNN, and replace vertices with GT vertices if they deviate by more than 3 cells.

\subsection{Annot. New Domains via Online Fine-Tuning}
We now also tackle the scenario in which our model is trained on one dataset, and is used to annotate a novel dataset. As the new data arrives, the annotator uses our model to annotate objects and corrects wrong predictions when necessary. We propose a simple approach to fine-tune our model in such a scenario, in an online fashion. 

Let us denote $C$ as the number of chunks the new data is divided into, $CS$ as the chunk size, $N_{EV}$ as the number of training steps for the evaluator network and $N_{MLE}, N_{RL}$ as the number of training steps for each chunk with MLE and RL, respectively. Our online fine-tuning is described in Algorithm~\ref{algo:active} where $PredictAndCorrect$ refers to the (simulated) annotator in the loop. Because we train on corrected data, we smooth our targets for MLE training with a manhattan distance transform truncated at distance 2.
\vspace{-2mm}
\RestyleAlgo{boxruled}
\SetAlFnt{\small}
\begin{algorithm}[h]
\SetAlgoLined
 bestPoly = cityscapesPoly\;
 \While{currChunk in ($1..C$)}{
  rawData = readChunk(currChunk)\;
  data = $PredictAndCorrect$(rawData, bestPoly)\;
  data += $SampleFromSeenData$($CS$)\;   
  newPoly = $Train_{MLE}$(data, $N_{MLE}$, bestPoly)\;
  newPoly = $Train_{RL}$(data, $N_{RL}$, newPoly)\;  
  newPoly = $Train_{EV}$(data, $N_{EV}$, newPoly)\;  
  bestPoly = newPoly\;
 }
 \caption{Online Fine Tuning on New Datasets}
 \label{algo:active}
\end{algorithm}
\begin{table*}[t!]
\vspace{-3mm}
\begin{center}
{\footnotesize
\addtolength{\tabcolsep}{3pt}
\begin{tabular}{|l|c|c|c|c|c|c|c|c|c|}
\hline
Model & Bicycle & Bus & Person & Train & Truck & Motorcycle & Car & Rider & {\bf Mean} \\
\hline\hline
Square Box & 35.41 & 53.44 & 26.36 & 39.34 & 54.75 & 39.47 & 46.04 & 26.09 & 40.11 \\
Dilation10 & 46.80 & 48.35 & 49.37 & 44.18 & 35.71 & 26.97 & 61.49 & 38.21 &43.89 \\
DeepMask \cite{deepmask} & 47.19 & 69.82 & 47.93 & 62.20 & 63.15 & 47.47 & 61.64 & 52.20 & 56.45 \\
SharpMask \cite{sharpmask} & 52.08 & 73.02 & 53.63 & 64.06 & 65.49 & 51.92 & 65.17 & 56.32 & 60.21 \\
Polygon-RNN  \cite{polyrnn} & 52.13 & 69.53 & 63.94 & 53.74  & 68.03 & 52.07 & 71.17 & 60.58 & 61.40 \\

\hline 
\hline
Residual Polygon-RNN &  54.86   &  69.56   &  67.05   & 50.20   &  66.80  & {55.37}   &  70.05    &  63.40    &  62.16   \\
\hspace{2mm}{+ Attention} &  {56.47}  &  {73.57}  &  {68.15}  & 53.31   &  {74.08}  &   {57.34}  &  {75.13}   &  {65.42}   & {65.43}  \\
\hspace{2mm}{+ RL} &  {57.38}  &  {75.99}  &  {68.45}  & 59.65   &  {76.31}  &  58.26  & {75.68}   & {65.65}   &  {67.17}  \\
\hspace{2mm}{+ Evaluator Network} & {62.34}  &  {79.63}  &   {70.80}   & {62.82}   &   {77.92}  &   {61.69}  &  {78.01}   &  {68.46}   & {70.21}  \\
\hspace{2mm}{+ GGNN} & \bf{63.06}  & \bf{81.38}   &  \bf{72.41}   & \bf{64.28}   &  \bf{78.90}    &  \bf{62.01}    & \bf{79.08}  &  \bf{69.95}   &   \bf{71.38} \\
\hline
\end{tabular}
\vspace{-2mm}
\caption{Performance (IoU in \% in val test) on all the Cityscapes classes in {\bf automatic mode}. All methods exploit GT boxes.}
\label{tbl:performance_cityscapes}
}
\end{center}
\end{table*}

\begin{figure*}[t!]
\vspace{-6mm}
\begin{minipage}{0.33\linewidth}
\includegraphics[width=1\linewidth,trim=55 0 70 30,height=7em,clip]{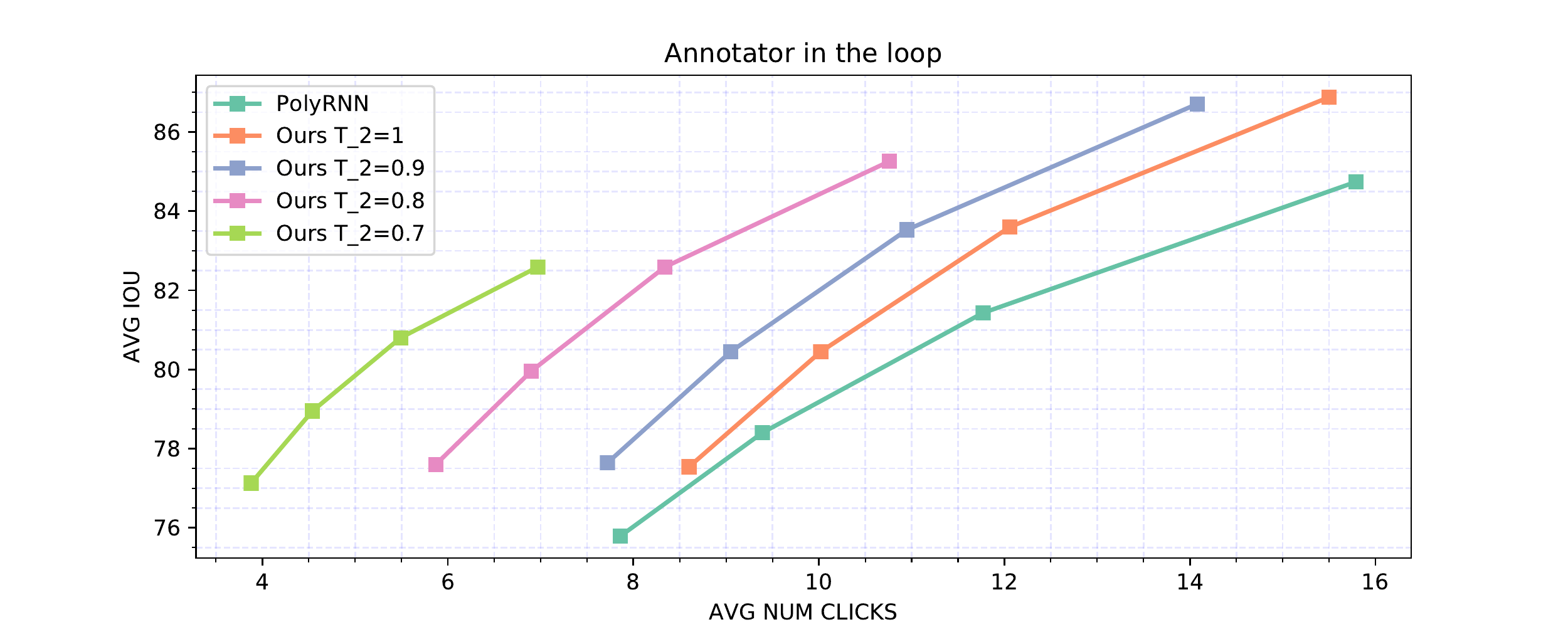} 
\vspace{-7mm}
\caption{Interactive mode on Cityscapes}
\label{fig:annotator_in_the_loop_cityscapes}
\end{minipage}
\hfill
\begin{minipage}{0.33\linewidth}
\includegraphics[width=1\linewidth,trim=35 0 70 31,clip,height=7em]{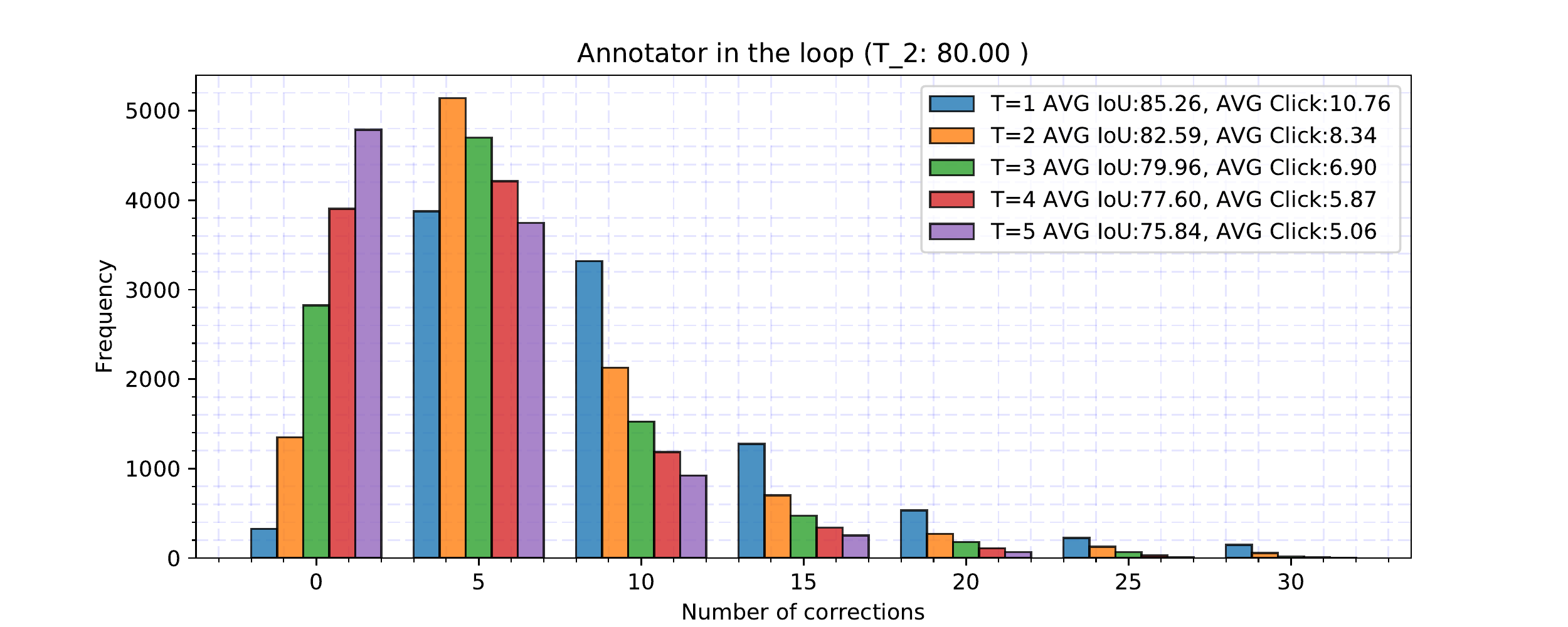} 
\vspace{-7mm}
\caption{Interactive mode on Citysc. $(T_2=0.8)$}
\label{fig:annotator_cityscapes_histogram}
\end{minipage}
\hfill
\begin{minipage}{0.33\linewidth}
\includegraphics[width=1\linewidth,trim=55 0 70 30,clip,height=7em]{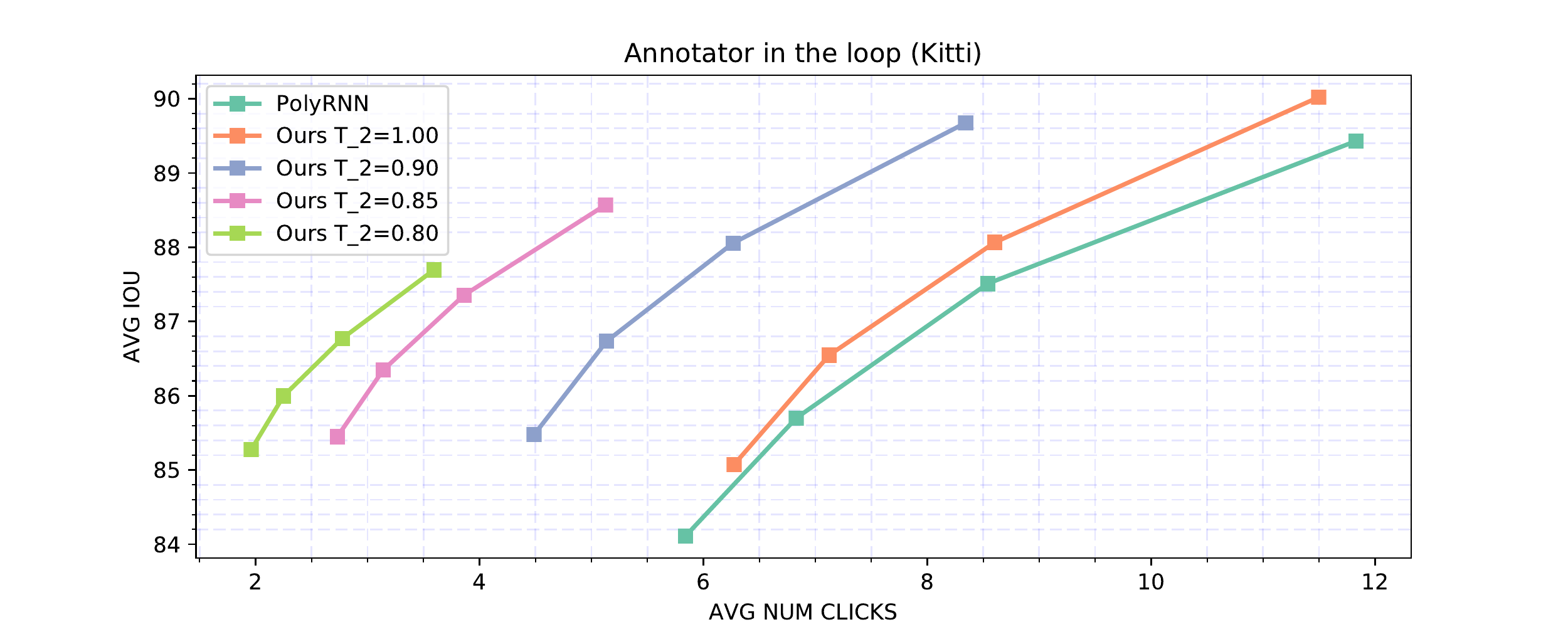} 
\vspace{-7mm}
\caption{Interactive mode on KITTI}
\label{fig:annotator_in_the_loop_kitti}
\end{minipage}
\vspace{-4mm}
\end{figure*}

\vspace{-6mm}
\section{Experimental Results}
\label{sec:results}

In this section, we provide an extensive evaluation of our model. We report both automatic and interactive instance annotation results on the challenging Cityscapes dataset~\cite{cityscapes} and compare with strong pixel-wise methods. We then characterize the generalization capability of our model with evaluation on the KITTI dataset~\cite{kitti} and four out-of-domain datasets spanning general scenes~\cite{ade20k}, aerial~\cite{Sun14}, and medical imagery~\cite{medical1,medical2}. Finally, we evaluate our online fine-tuning scheme, demonstrating significant decrease in annotation time for novel datasets. Note that as in~\cite{polyrnn}, we assume that user-provided ground-truth boxes around objects are given. We further analyze robustness of our model to noise with respect to these boxes, mimicking noisy annotators.  

\vspace{-2mm} 
\subsection{In-Domain Annotation}
\vspace{-1mm}

We first evaluate our approach in training and evaluating on the same domain. This mimics the scenario where one takes an existing dataset, and uses it to annotate novel images from the same domain. In particular, we use the
 Cityscapes dataset~\cite{cityscapes}, which is currently one of the most comprehensive benchmarks for instance segmentation. It contains  2975 training, 500 validation and 1525 test images with 8 semantic classes. %
To ensure a fair comparison, we follow the same alternative split proposed by~\cite{polyrnn}. %
As in~\cite{polyrnn}, we preprocess the ground-truth polygons according to depth ordering to obtain polygons for only the visible regions of each instance. 

\vspace{-3.5mm}
\paragraph*{Evaluation Metrics:} We utilize two quantitative measures to evaluate our model. {\bf 1)} We use the intersection over union (IoU) metric to evaluate the quality of the generated polygons and {\bf 2)} we calculate the number of annotator clicks required to correct the predictions made by the model. We describe the correction protocol in detail in a subsequent section.

\vspace{-3.5mm}
\paragraph*{Baselines: } 
Following~\cite{polyrnn}, we compare with DeepMask~\cite{deepmask}, SharpMask~\cite{sharpmask}, as well as Polygon-RNN~\cite{polyrnn} as state-of-the-art baselines. Note that the first two approaches are pixel-wise methods and errors in their output cannot easily be corrected by an annotator. To be fair, we only compare our automatic mode with their approaches. In their original approach, \cite{deepmask,sharpmask} exhaustively sample patches at different scales over the entire image. Here, we evaluate \cite{deepmask,sharpmask} by providing exact ground-truth boxes to their models. %

We also use two additional baselines, namely SquareBox from~\cite{polyrnn} and Dilation10 from~\cite{dilation}. %
SquareBox considers the provided bounding box as its prediction. Dilation10 is obtained from the segmentation results of \cite{dilation} from the model that was trained on the Cityscapes dataset.

\vspace{-3.5mm}
\paragraph*{Automatic Mode:} 
We compare Polygon-RNN++ to the baselines in Table~\ref{tbl:performance_cityscapes}, as well as ablate the use of each of the components in our model.  Here,  Residual Polygon-RNN  refers to the original model with our novel image architecture instead of VGG. 
Our full approach outperforms the top performer~\cite{polyrnn} by almost 10\% IoU, and achieves best performance for each class. Moreover, Polygon-RNN++ surpasses the reported human agreement~\cite{polyrnn} of 78.6\% IoU on \emph{cars}, on average. Using human agreement on cars as a proxy, we observe that the model also obtains human-level performance for the \emph{truck} and \emph{bus} classes. %

\begin{table*}[!htb]
\vspace{-1mm}
\begin{minipage}{.48\linewidth}
\begin{center}
{\footnotesize
\addtolength{\tabcolsep}{-0.6pt}
\begin{tabular}{|l|c|c|c|c|}
\hline
Model & ADE & Rooftop & Cardiac MR & ssTEM\\
\hline\hline
SquareBox (Expansion) & 42.95 & 40.71 & 62.10 & 42.24 \\
Ellipse (Expansion) & 48.53 & 47.51 & 73.63 & 51.04\\
\hline
Square Box (Perfect) & 69.35 & 62.11 & 79.11 & 66.53\\
Ellipse (Perfect) &  69.53 & 66.82 & 92.44 & 71.32 \\
\hline
DeepMask\cite{deepmask}  & 59.74 & 15.82 & 60.70 & 31.21 \\
SharpMask\cite{sharpmask}  & 61.66 & 18.53  & 69.33 & 46.67  \\
\hline
Ours w/o GGNN & 70.21 & 65.03 & 80.55 & 53.77\\
Ours w/ GGNN & 71.82 & 65.67 & 80.63 & 53.12\\
\hline
\end{tabular}
\vspace{-3mm}
\caption{Out-of-domain automatic mode performance}
\label{tbl:performance_cross_domain_all}
\vspace{0.5mm}
}
\end{center}
\end{minipage}
\hspace{3mm}
\begin{minipage}{.265\linewidth}
	\begin{center}
	{\footnotesize
	\addtolength{\tabcolsep}{-1pt}
    \begin{tabular}{|l|c|}
    \hline
    Model & IoU (\%) \\
    \hline\hline
    DeepMask \cite{deepmask} & 78.3 \\ 
    SharpMask \cite{sharpmask} & 78.8 \\ 
    Beat The MTurkers \cite{ChenCVPR14} & 73.9 \\ 
    Polygon-RNN~\cite{polyrnn} & 74.22 \\
    \hline
    Ours w/o GGNN & 81.40\\
    Ours w/ GGNN & \bf{83.14}\\
    \hline
    \end{tabular}
    \vspace{-2mm}
	\caption{Car annot. results on KITTI in automatic mode (no fine-tuning, 
	$0$ clicks)}    
    \label{tbl:cross_domain_kitti}
    }  
    \end{center}
\end{minipage}
\hspace{2mm}
\begin{minipage}{.20\linewidth}
	\begin{center}    
    {\footnotesize
    \addtolength{\tabcolsep}{-1pt}
    \begin{tabular}{|l|c|}
    \hline
    Bbox Noise & IoU (\%)\\    
    \hline \hline
    0\% &  71.38\\ \hline
    0-5\% &  70.54 \\ \hline%
    5-10\% &  68.07 \\ \hline %
    10-15\% &  64.80 \\%
    \hline
	\end{tabular}
    \vspace{-2mm}
    \caption{Robustness to Bounding Box noise on Cityscapes (in \% of side length at each vertex)}
    \label{tbl:robustness_to_noise}
	}	
	\end{center}
\end{minipage}
\end{table*}

\begin{figure*}
\vspace{-3mm}
\hfill\minipage{0.4\textwidth}
\includegraphics[width=1.0\textwidth]{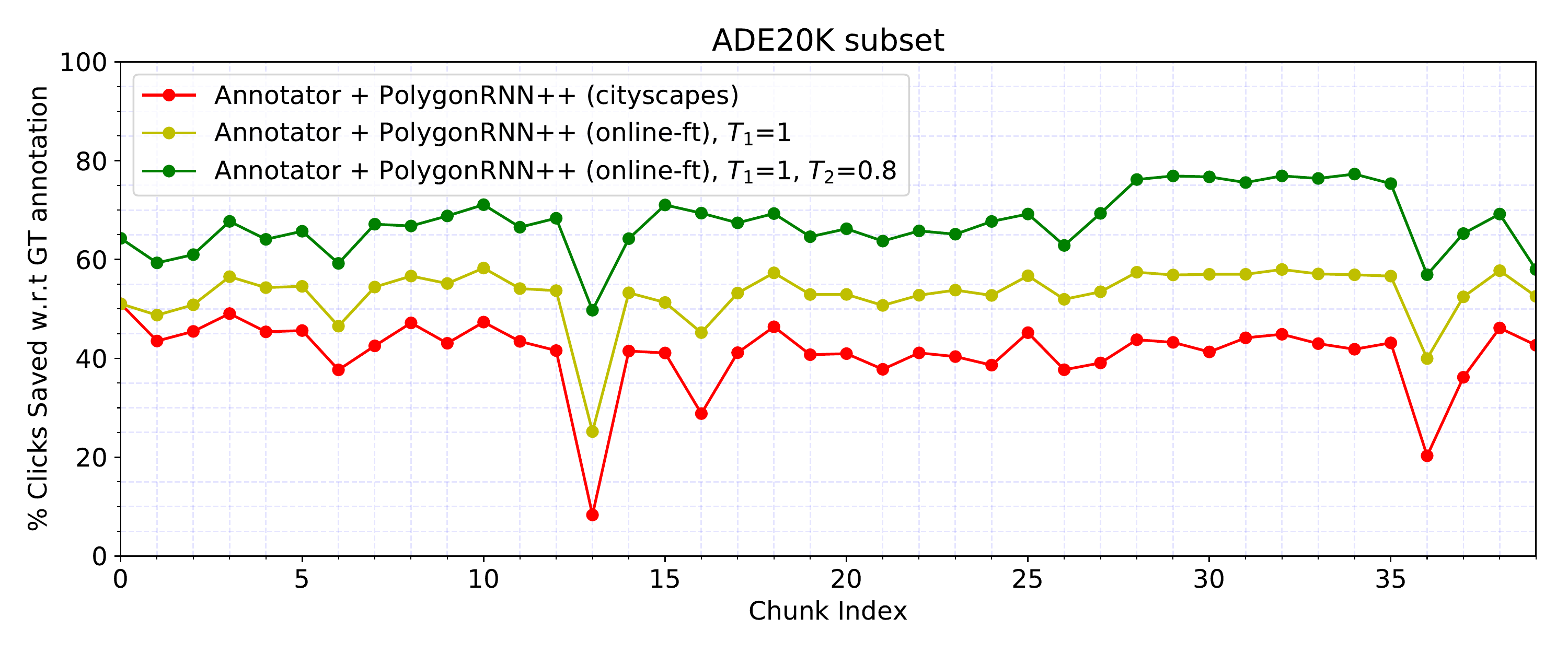}
\endminipage\hfill
\minipage{0.2\textwidth}
\includegraphics[width=1.0\textwidth]{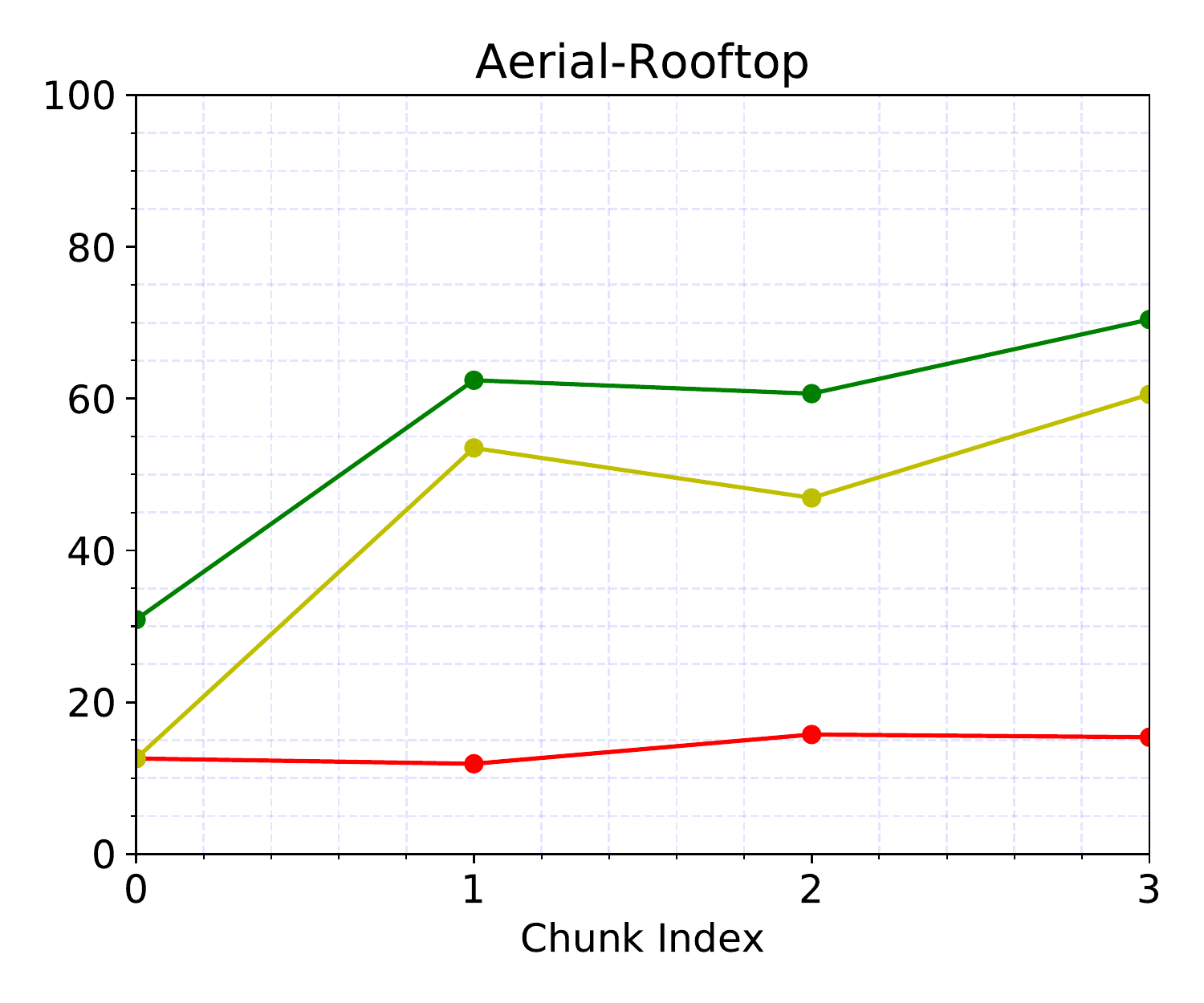}
\endminipage\hfill
\minipage{0.2\textwidth}
\includegraphics[width=1.0\textwidth]{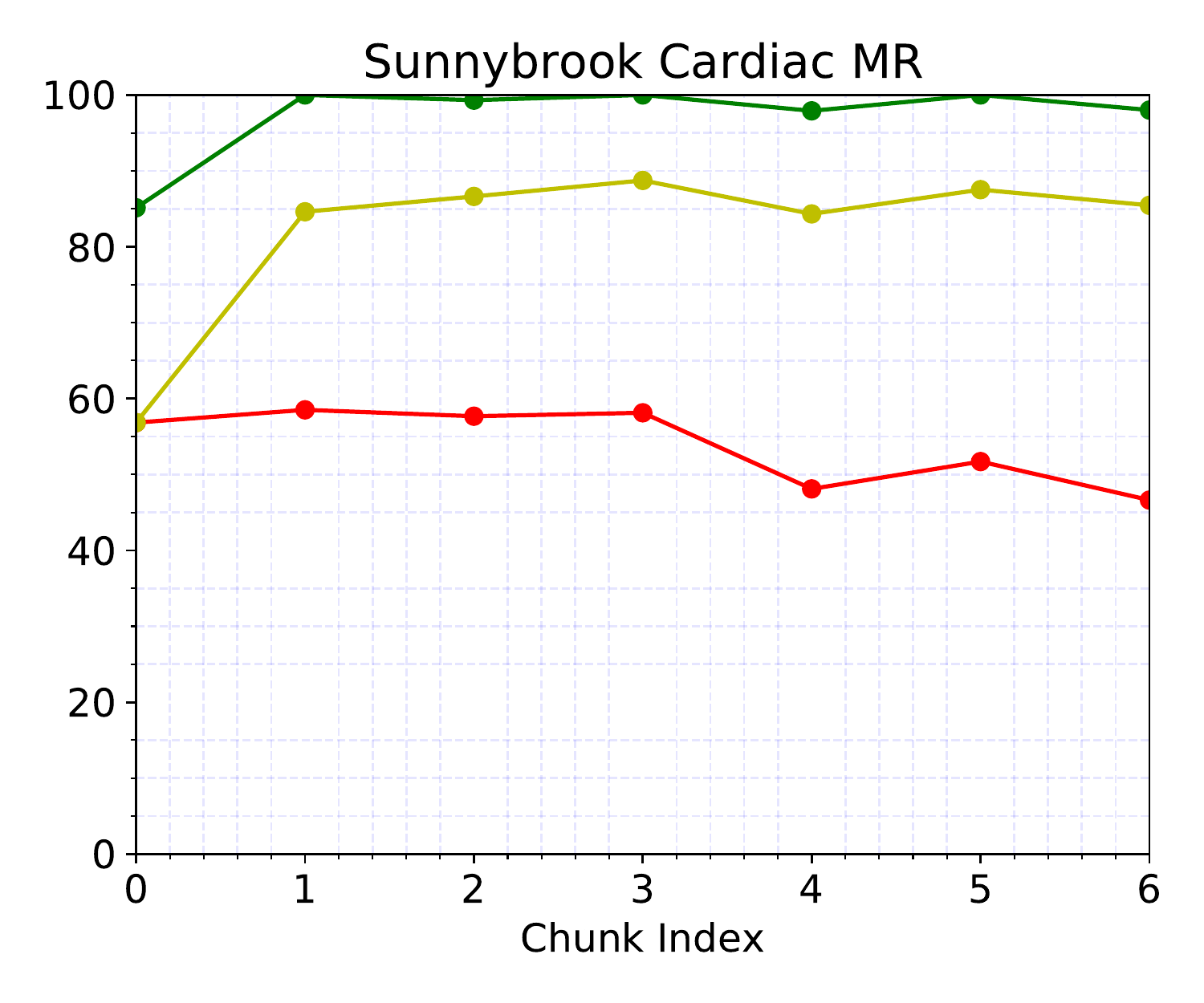}
\endminipage\hfill
\minipage{0.2\textwidth}
\includegraphics[width=1.0\textwidth]{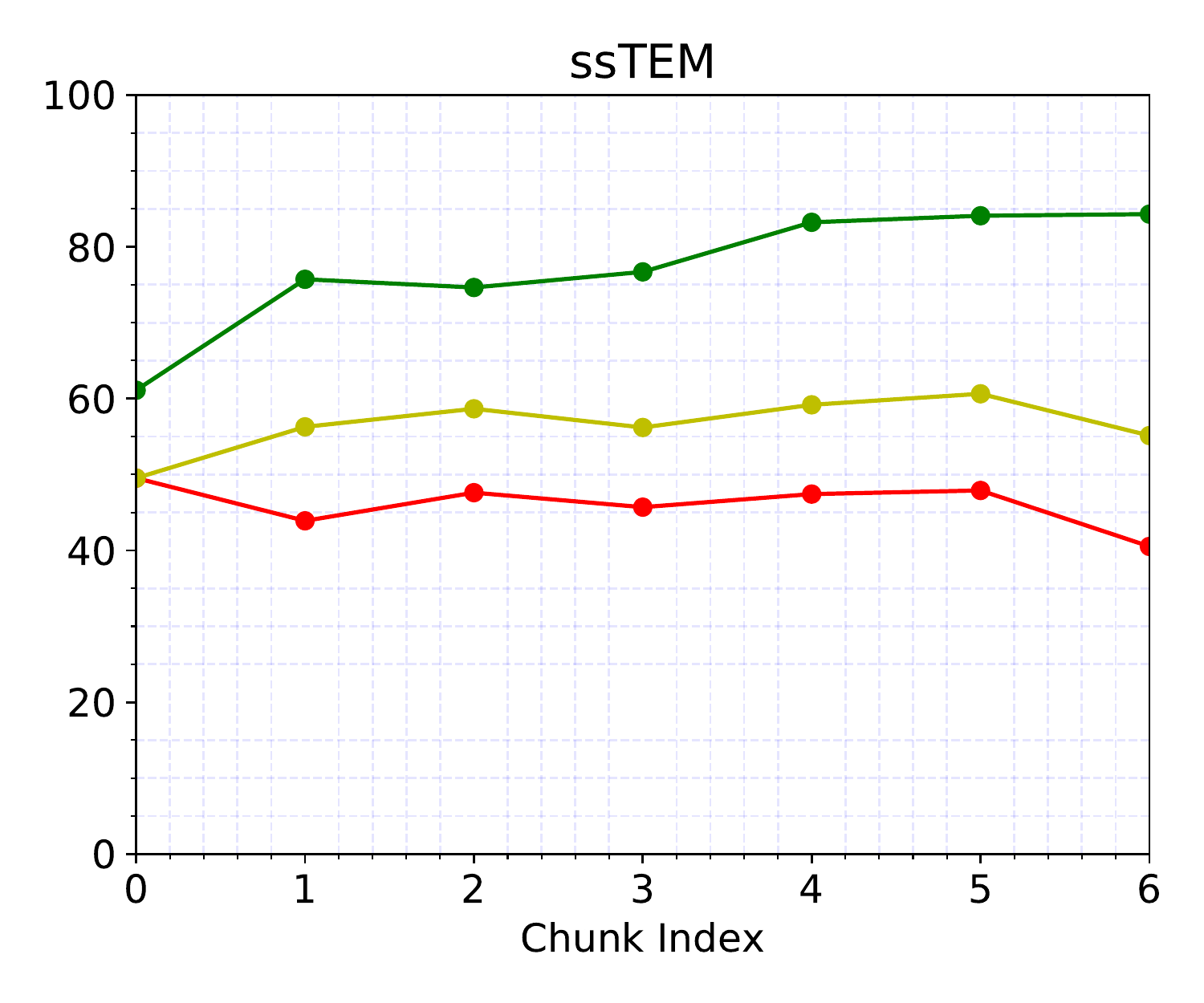}
\endminipage\hfill
\vspace{-3.5mm}
\caption{\small Percentage clicks saved with online fine-tuning on out-of-domain datasets (Plots share legend and y axis)}
\label{fig:active}
\vspace{-3mm}
\end{figure*}

\vspace{-3mm}
\paragraph*{Interactive Mode:}
The interactive mode aims to minimize annotation time while obtaining high quality annotations. Following the simulation proposed in \cite{polyrnn}, we calculate the number of annotator clicks required to correct predictions from the model. The annotator corrects a prediction if it deviates from the corresponding GT vertex by a min distance of $T$, where the hyperparameter $T$ governs the quality of the produced annotations. For fair comparison, distances are computed using manhattan distance at the model output resolution using distance thresholds $T \in [1, 2, 3, 4]$, as in~\cite{polyrnn}. 

Additionally, we introduce a second threshold $T_2$, which is defined as the IoU between the predicted polygon and the GT mask, where we consider polygons achieving agreement above $T_2$ unnecessary for the annotator to interfere. We exploit this threshold due to the somewhat unsatisfactory correction simulation above: for example, if the predicted vertex falls along a GT polygon edge, this vertex is in fact correct and should not be corrected. 
Note that, in the extreme case of $T_2=1$, our simulator assumes that corrections are necessary for every predicted polygon. In this case, the simulation is equivalent to the one presented in~\cite{polyrnn}. %

In Fig.~\ref{fig:annotator_in_the_loop_cityscapes}, we compare the average number of clicks per instance required to annotate {\bf all classes} on the Cityscapes \textit{val} set (500 images) with different values of $T_2$. Using $T_2=1$, we see that our model outperforms~\cite{polyrnn}, requiring fewer clicks to obtain the same IoU. It is particularly impressive that at $T_2=0.8$ our model is still more accurate than Polygon-RNN at $T_2=1.0$. At $T_2=0.7$, we achieve over $80\%$ mIoU with only 5 clicks per object on average, which is a reduction of more than 50\% over~\cite{polyrnn}. Fig.~\ref{fig:annotator_cityscapes_histogram} shows frequency of required corrections  for different $T$ at $T_2=0.8$. In~\ref{ss:human_annotator}, we show results with real human annotators.

\vspace{-3.5mm}
\paragraph*{Robustness to bounding box noise: } 
To simulate the effect of a lazy annotator, we analyze the effect of noise in the bbox provided to the model. We randomly expand the bbox by a percentage of its width and height. Results in Table~\ref{tbl:robustness_to_noise} illustrates that our model is very robust to some amount of noise (0-5\%). Even in the presence of moderate and large noise (5-10\%,10-15\%), it outperforms the reported performance of previous baselines which use perfect bboxes.

\vspace{-3.5mm}
\subsubsection{Instance-Level Segmentation}
\vspace{-1mm}

We evaluate our model on the task of (automatic) full-image object instance segmentation. %
Since PolygonRNN++ requires bounding boxes, we use FasterRCNN~\cite{fasterrcnn} to detect objects.
The predicted boxes are then fed to our model to produce polygonal instance segmentations. Evaluating Polygon-RNN++ with FasterRCNN on the {\bf Cityscapes test set} achieves $22.8 \%$  AP and $42.6\%$ $AP_{50}$. Following~\cite{SGN17}, we also add semantic segmentation~\cite{pspnet} to post-process the results. We simply perform a logical ``and" operation between the  predicted class-semantic map and our prediction. 
Following this scheme, we achieve $25.49\%$  AP and $45.47\%$ $AP_{50}$ on the {\bf test set}. More details are in the Appendix.

\begin{figure*}[t!]
\vspace{-1mm}
\includegraphics[width=0.496\linewidth, trim=180 200 150 250,clip]{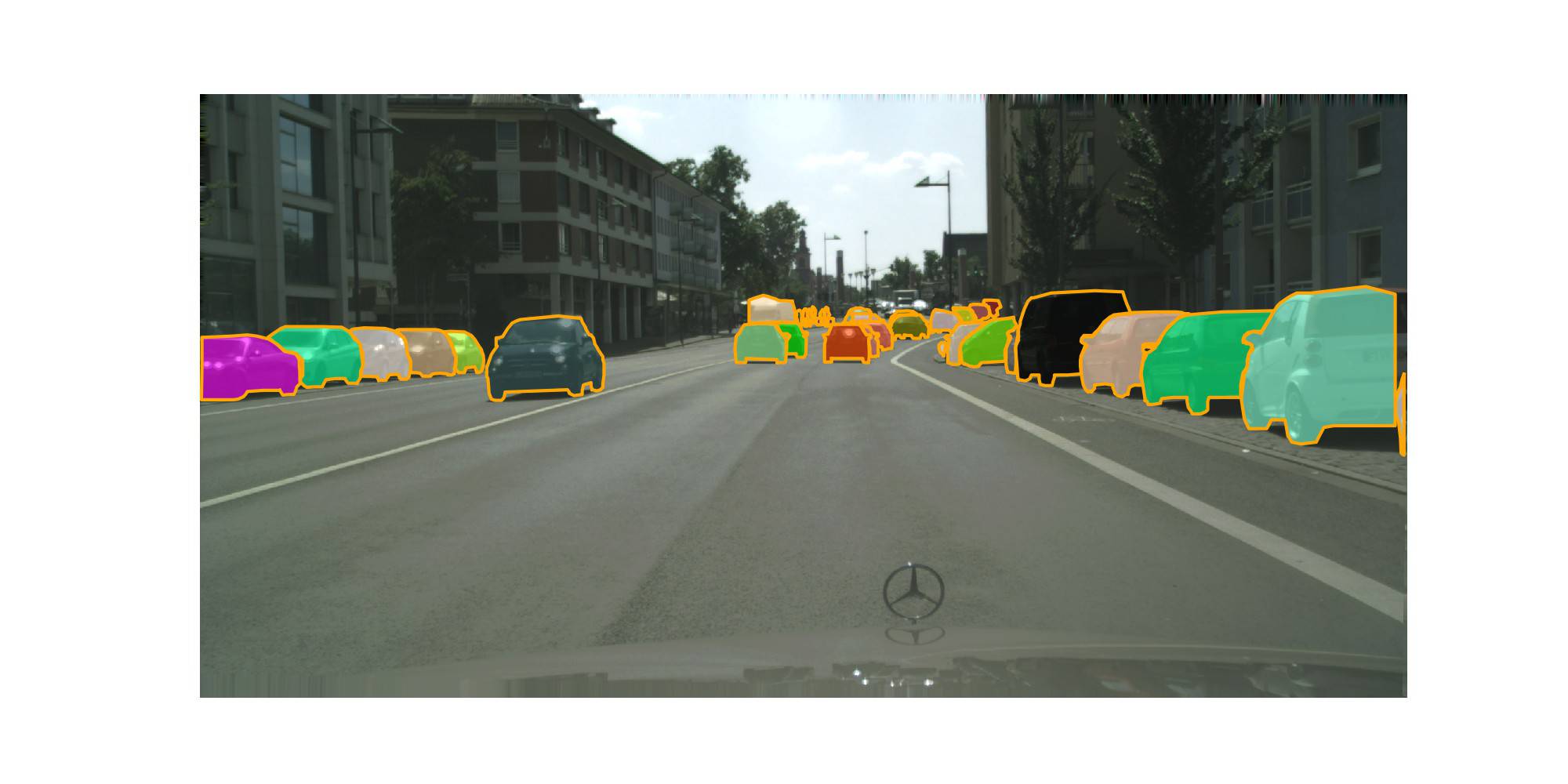}\hspace{0.7mm}\includegraphics[width=0.496\linewidth, trim=180 200 150 250,clip]{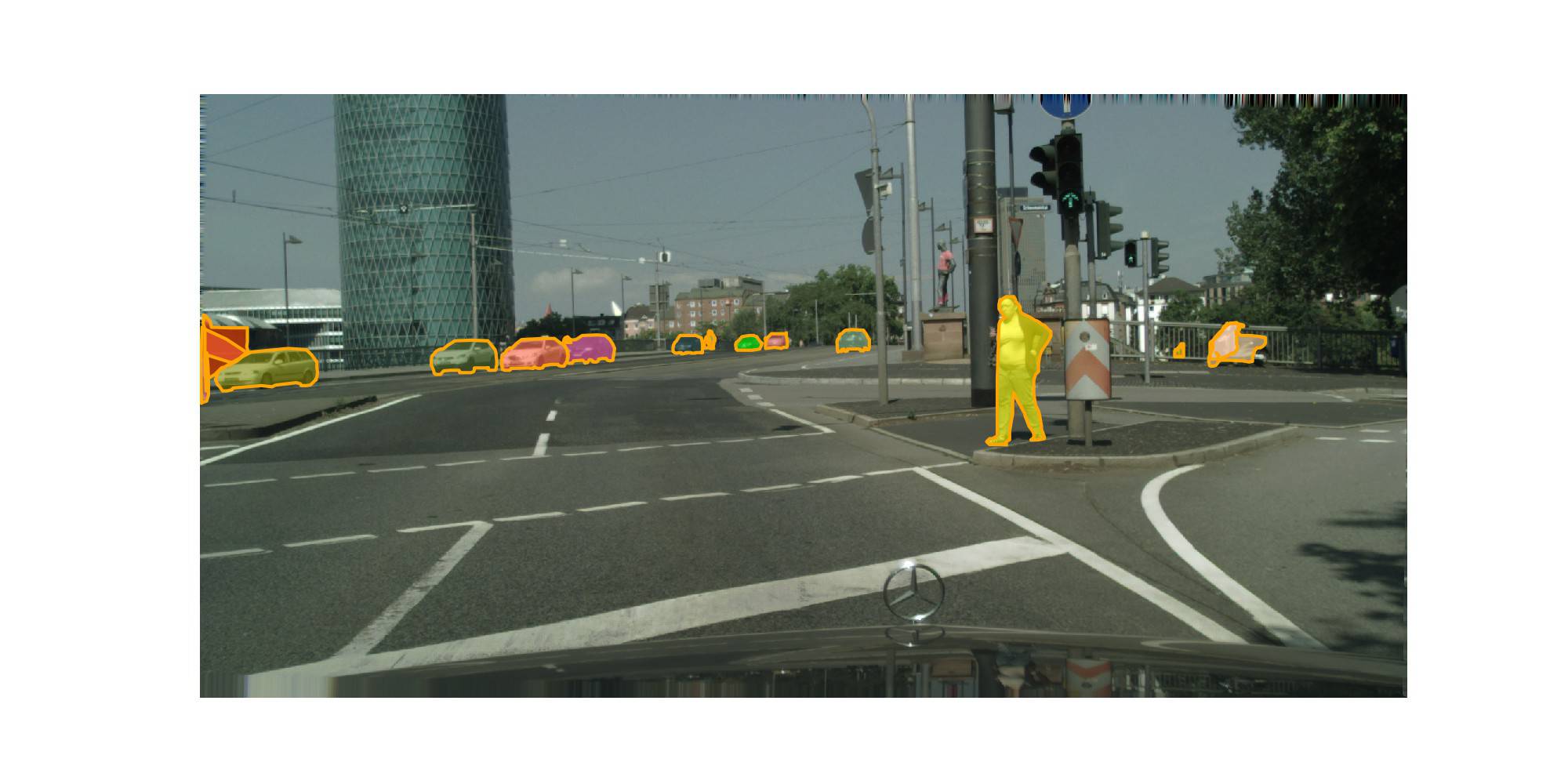}\\
\includegraphics[width=0.496\linewidth, trim=180 250 150 200,clip]{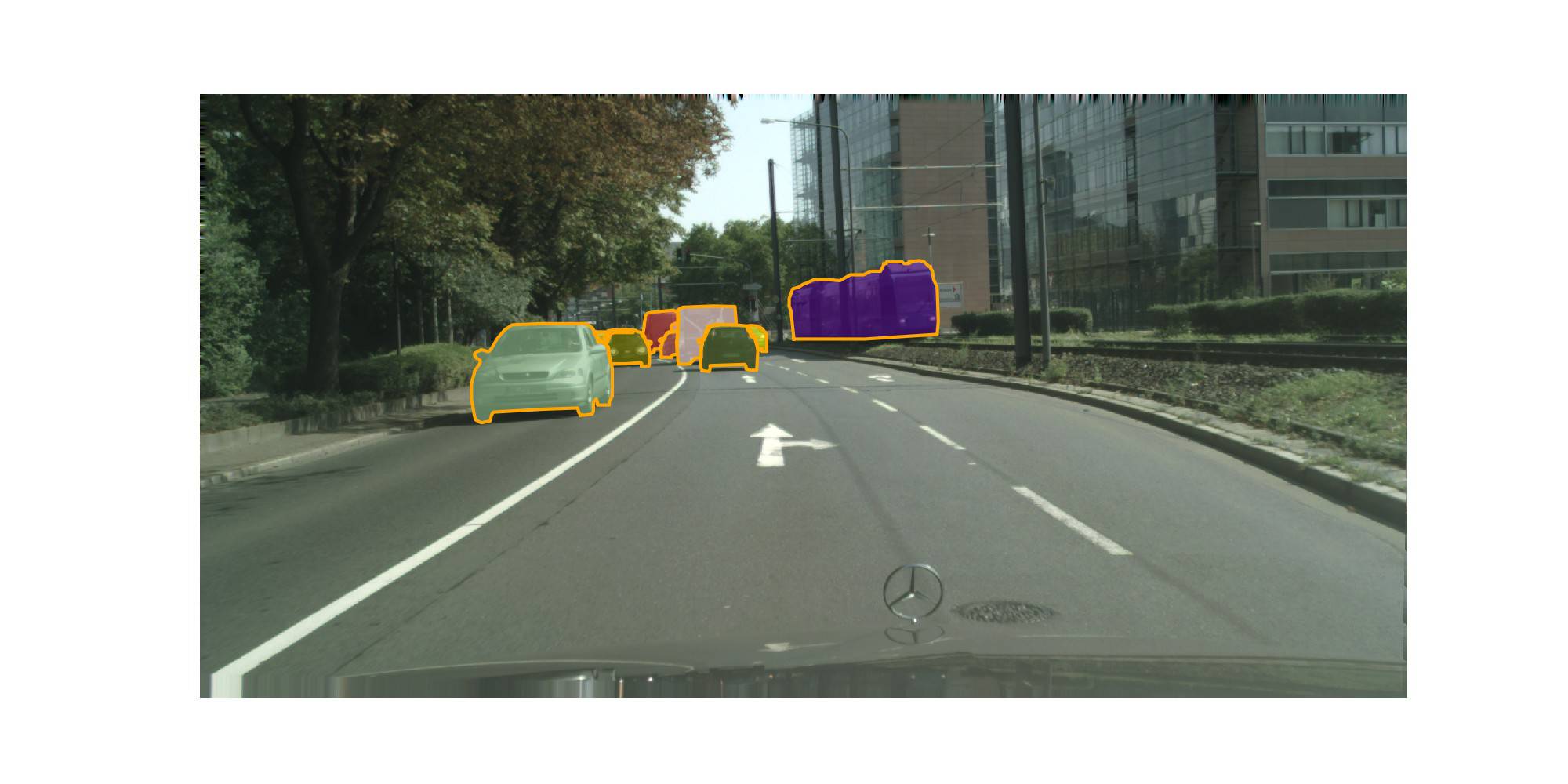}\hspace{0.7mm}\includegraphics[width=0.496\linewidth, trim=180 250 150 200,clip]{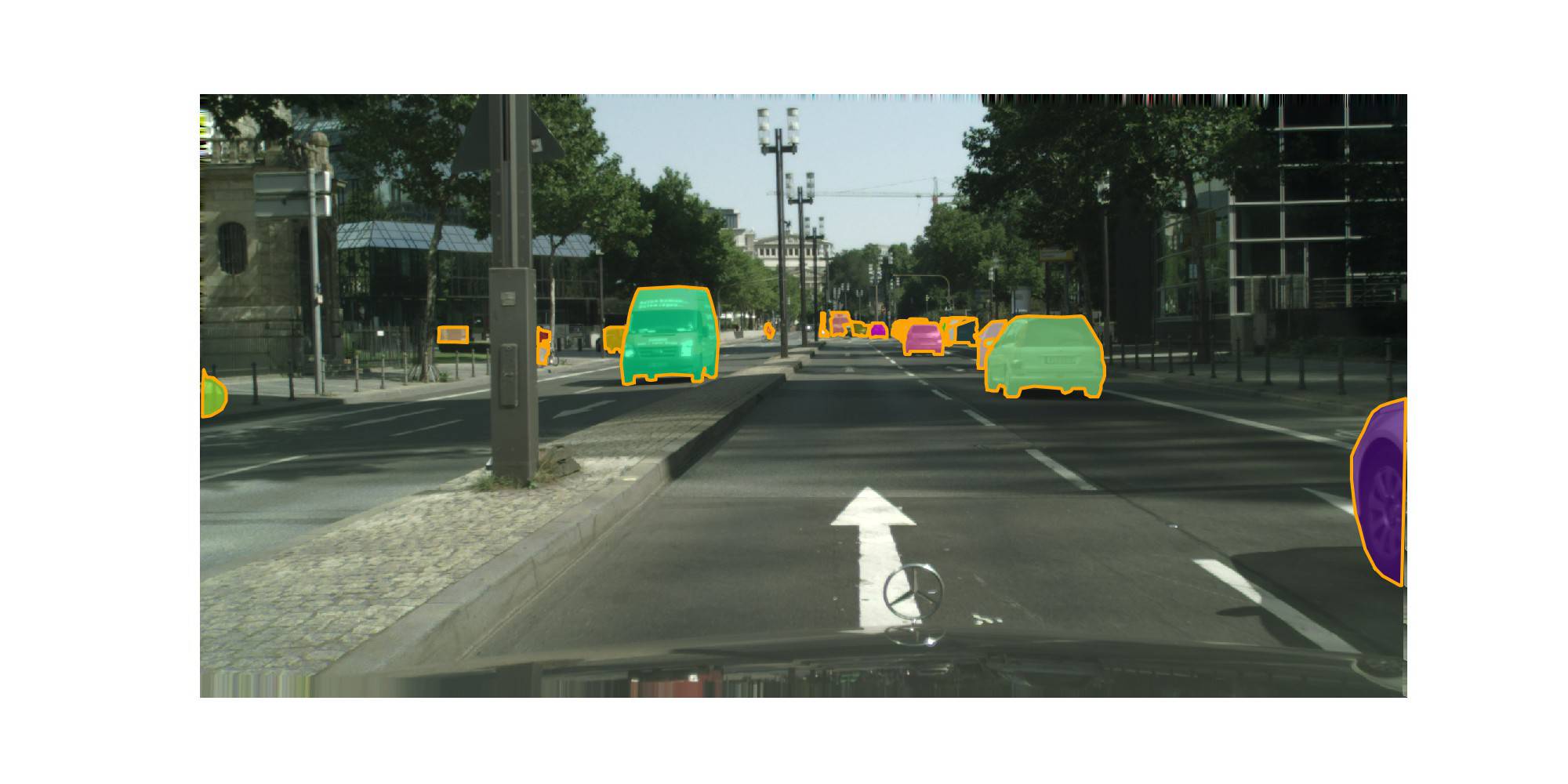}\\\includegraphics[width=0.496\linewidth, trim=180 250 150 200,clip]{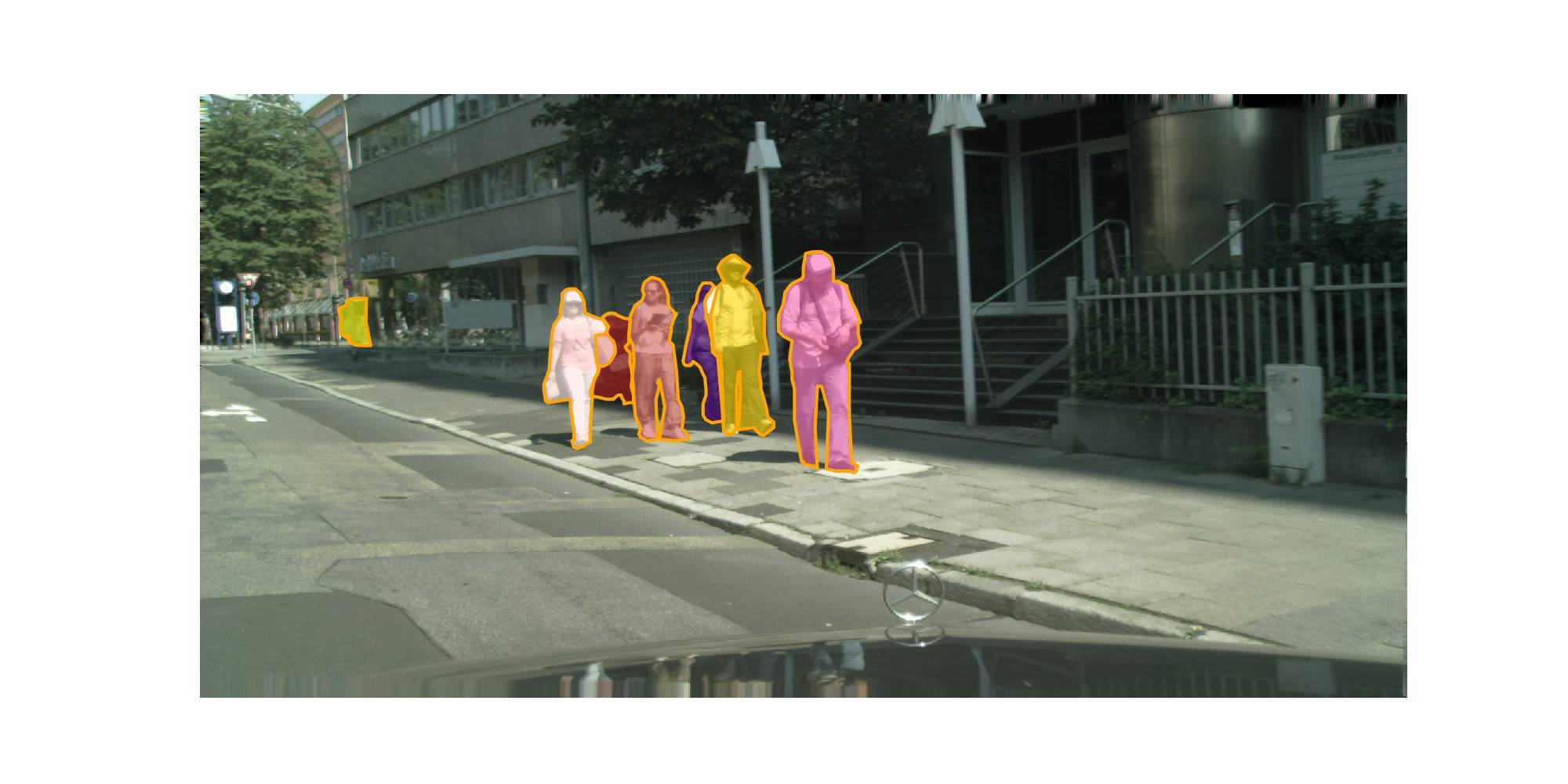}
\includegraphics[width=0.496\linewidth, trim=180 200 150 250,clip]{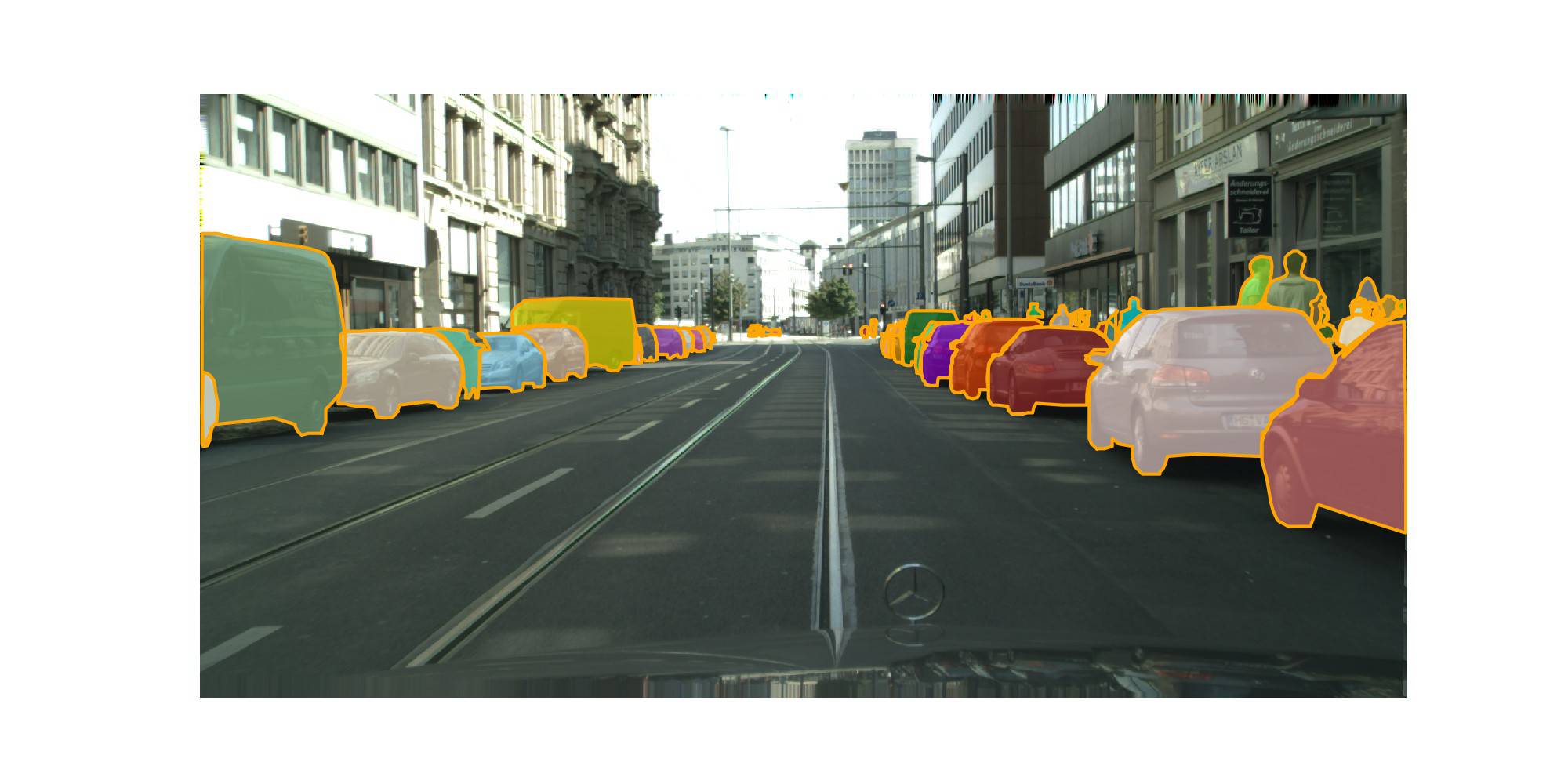}
\vspace{-2mm}
\caption{Qualitative results in automatic mode on Cityscapes. \textbf{Left:}  requiring ground-truth (annotator-provided) bounding boxes, but 0 clicks for the polygons; \textbf{Right:} Full automatic image prediction (FasterRCNN + PolygonRNN++) }
\label{fig:qualitative_cityscapes}
\vspace{-2mm}
\end{figure*}
\begin{figure*}[h!]
\begin{minipage}{0.51\textwidth}
\vspace{-2mm}
\centering
\includegraphics[width=1\linewidth, trim = 40 210 105 0, clip]{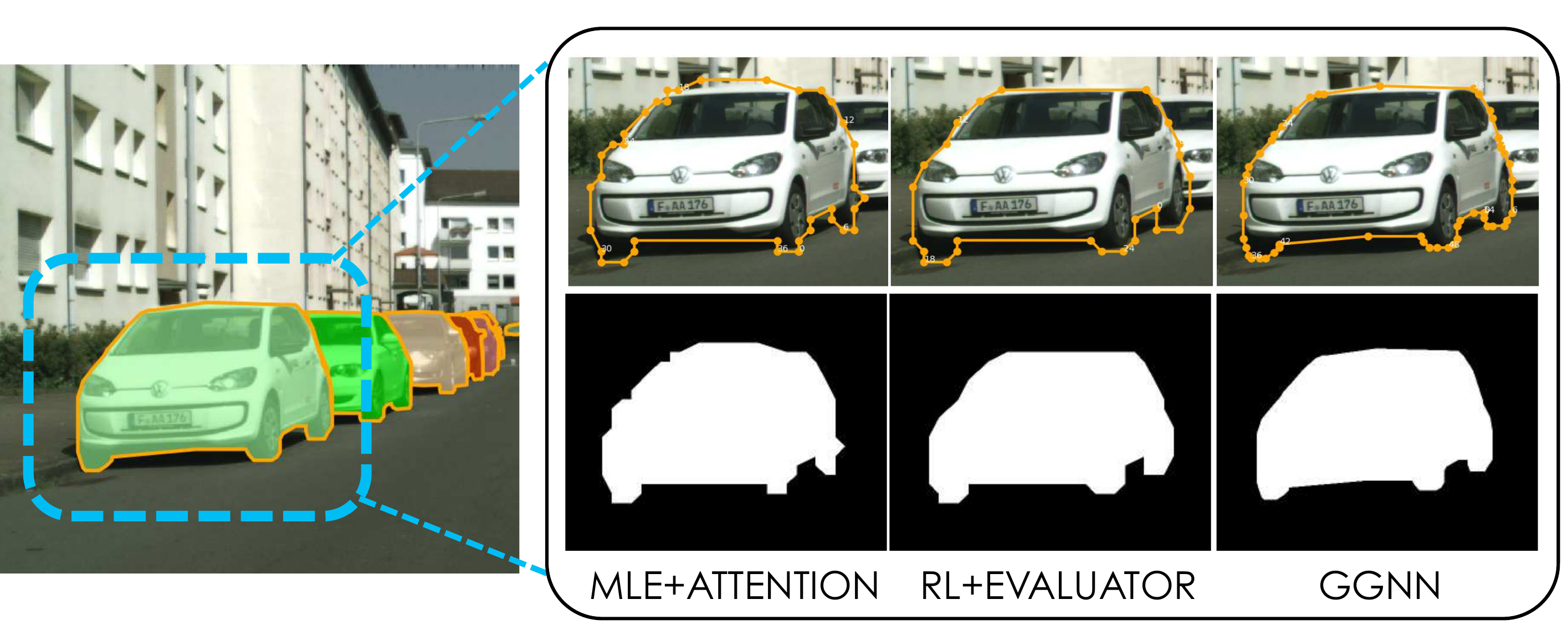}
\vspace{-7mm}
\caption{Results with different components of the model}
\label{fig:qualitative_zoomed}
\end{minipage}\hspace{2mm}
\begin{minipage}{0.475\textwidth}
\centering
\includegraphics[width=1\linewidth, trim= 10 330 300 0, height=8.1em,clip]{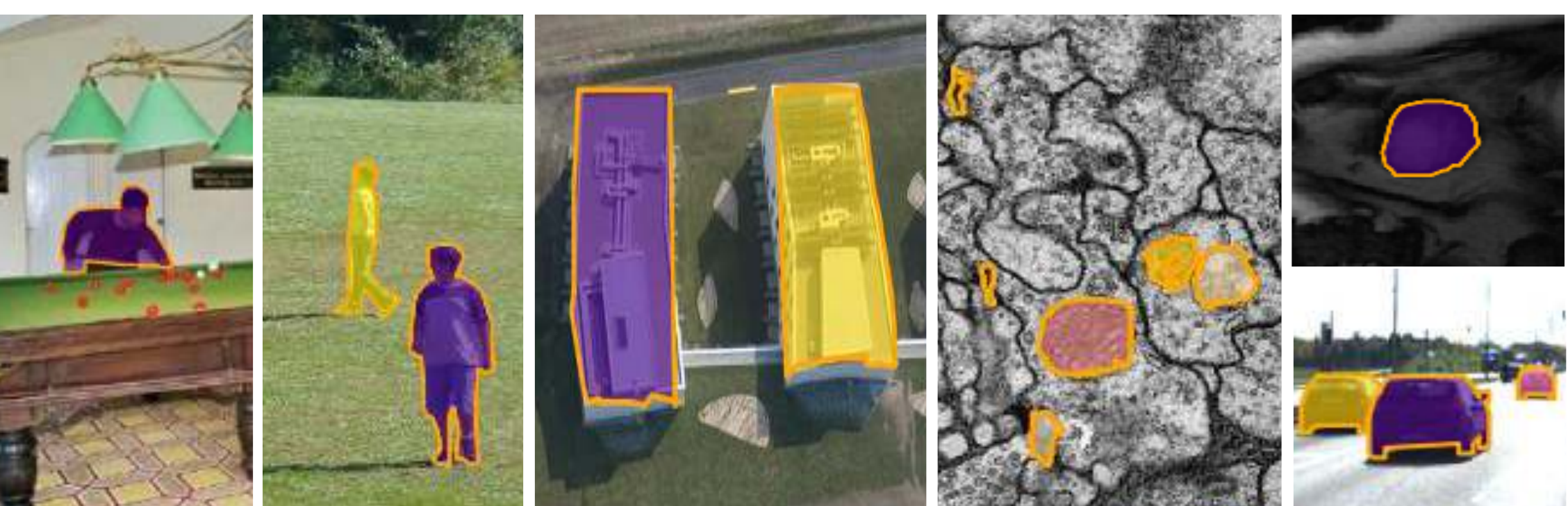}
\vspace{-6mm}
\caption{Qualitative results in automatic mode on different unseen datasets without fine-tuning }
\label{fig:qualitative_several}
\end{minipage}
\vspace{-6mm}
\end{figure*}

\begin{figure}[t!]
\vspace{-0mm}
\includegraphics[width=\linewidth, trim=50 180 0 130, clip]{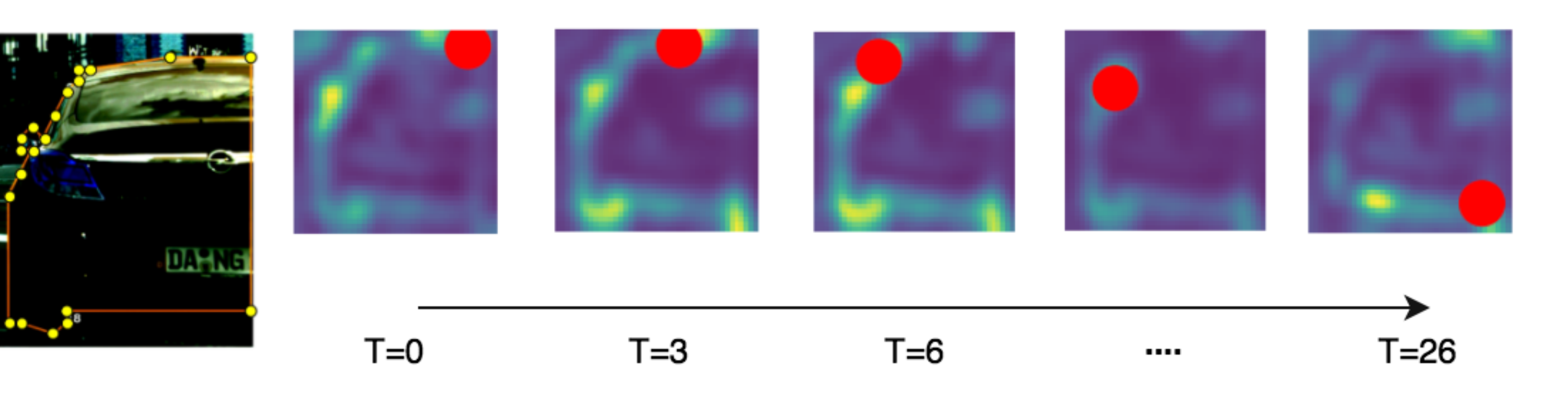}
\vspace{-8mm}
\caption{\small Visualization of attention maps in our model}
\label{fig:attention}
\vspace{-3mm}
\end{figure}

\vspace{-1mm}
\subsection{Cross-Domain Evaluation}
In this section, we analyze the performance of our model on different datasets that capture both shifts in environment (KITTI~\cite{kitti}) and domain (general scenes, aerial, medical). We first use our model trained on Cityscapes {\bf without any fine-tuning} on these datasets.

\vspace{-3mm}
\paragraph*{KITTI~\cite{kitti}:} 
We use Polygon-RNN++ to annotate 741 instances of KITTI~\cite{kitti} provided by~\cite{ChenCVPR14}. 
The results in automatic mode are reported in Table~\ref{tbl:cross_domain_kitti} and the performance with a human in the loop is illustrated in Fig.~\ref{fig:annotator_in_the_loop_kitti}. Our method outperforms all baselines showcasing its robustness to change in environment while being in a similar domain.
With an annotator in the loop, our model requires on average 5 fewer clicks than~\cite{polyrnn} to achieve the same IoU. It achieves human level agreement of $85\%$ as reported by~\cite{ChenCVPR14} by requiring only 2 clicks on average by the (simulated) annotator.
\vspace{-3mm}

\subsubsection{Out-of-Domain Imagery}
\vspace{-2mm}
We consider datasets exhibiting varying levels of domain shift from the Cityscapes dataset in order to evaluate the generalization capabilities of our model.

\vspace{-4mm}
\paragraph*{ADE20K~\cite{ade20k}:} 
The ADE20K dataset is a challenging general scene parsing dataset containing 20,210 images in the training set, 2,000 images in the validation set, and 3,000 images in the testing set. We select the following subset of categories from the validation set: \emph{television receiver}, \emph{bus}, \emph{car}, \emph{oven}, \emph{person} and \emph{bicycle} in our evaluation. We show results for more classes in the Appendix.

\vspace{-4mm}
\paragraph*{Aerial Imagery~\cite{Sun14}:} 
The Aerial Rooftop dataset \cite{Sun14} consists of 65 aerial images of rural scenes containing several building rooftops, a majority of which exhibit fairly complex polygonal geometry. Performance for this dataset is reported for the \textit{test} set.

\vspace{-4mm}
\paragraph*{Medical Imagery~\cite{medical1, medical1b, medical2}:} 
We use two medical segmentation datasets \cite{medical1,medical1b} and \cite{medical2} for our experiments.
The former, used in the Left Ventricle Segmentation Challenge \cite{medical1b}, divides the data of 200 patients equally in the \textit{training} and \textit{validation} sets. We report the performance on a subset of the \textit{validation} set which only includes the outer contours that segment the epicardium.
The latter provides two image stacks (training and testing) each containing 20 sections from serial section Transmission Electron Microscopy (ssTEM) images of the ventral nerve cord. %
We use the mitochondria and synapse segmentations from this data to test our model. Since ground-truth instances for the test set are not publicly available, we evaluate on the training set. %

\vspace{-4mm}
\paragraph*{Quantitative Results:} 
For out-of-domain datasets, we introduce a baseline (named Ellipse) which fits an ellipse into the GT bounding box which is motivated by the observation that many instances in~\cite{medical1b} are ellipses. We show results with perfect and expanded bounding boxes (expansion similar to our model) for Square Box and Ellipse. DeepMask and SharpMask were evaluated with perfect bounding boxes with the threshold suggested by the authors. Table~\ref{tbl:performance_cross_domain_all}, demonstrates high generalization capabilities of our model.

\vspace{-4mm}
\paragraph*{Online Fine-tuning:}
In these experiments, our simulated annotator has parameters $T=1$ and $T_2=0.8$. Fig.~\ref{fig:active} reports the percentage of clicks saved with respect to GT polygons for our Cityscapes model and the online fine-tuned models. We see that our adaptive approach overcomes stark domain shifts with as few as one chunk of data (40 images for Sunnybrook, 3 for ssTEM, 200 for ADE and 20 for Aerial) showcasing strong generalization. Overall, we show at least 65\% overall reduction in the number of clicks across all datasets, with the numbers almost at 100\% for the Sunnybrook Cardiac MR dataset. We believe these results pave the way towards a real annotation tool that can learn along with the annotator and significantly reduce human effort.

\subsection{Qualitative Results}
Fig.~\ref{fig:qualitative_cityscapes} shows example predictions obtained in automatic mode on Cityscapes.
We illustrate the improvements from specific parts of the model in Fig.~\ref{fig:qualitative_zoomed}. We see how using RL and the evaluator network leads to crisper predictions, while the GGNN upscales, adds points and builds a polygon resembling human annotation. Fig.~\ref{fig:qualitative_several} showcases automatic predictions from PolygonRNN++ on the out-of-domain datasets. We remind the reader that this labeling is obtained by exploiting GT bounding boxes, and no fine-tuning.

\subsection{Interaction with Human Annotators} 
\label{ss:human_annotator}
To corroborate our findings with a simulated annotator, we also conducted a small scale experiment with real human annotators in the loop. To this end, we implemented a very simple annotation tool that runs our model at the backend. 
We obtained 54 \emph{car} instances from Cityscapes used by~\cite{polyrnn} for their human experiment. We asked two human subjects to annotate these interactively using our model, and two to annotate manually. While we explain how the tool works, we do not train the annotators to use our tool. All our annotators were in-house. 

Timing begins when an annotator first clicks on an object, and stops when the "submit" button is clicked. While using our model, the annotator needs to draw a bounding box around the object, which we include in our reported timing. Note that we display the object to the annotator by cropping an image inside an enlarged box. Thus our annotators are fast in drawing the boxes, taking around 2 seconds on average. In the real scenario, the annotator would be annotating the full image, thus more time would be spent on placing boxes around the objects -- however, this step is typically common to fully manual annotation tools as well.

The results are presented in Table~\ref{tbl:human_exp}. We observe that when using our model, annotators are 3x faster, with only slightly lower IoU agreement with the ground truth. Note that we used a basic version of the tool, with scope for improvement in various engineering aspects. The authors of~\cite{polyrnn} reported that on these examples, human subjects needed on average 42.2 seconds per instance using GrabCut~\cite{Rother2004SIGGRAPH}, while also achieving a lower IoU (70.7).

We also use our model on cross-domain annotation. In particular, we use the ADE20k dataset and our model trained on Cityscapes (no fine-tuning). We randomly chose a total of 40 instances of \emph{car}, \emph{person}, \emph{sofa} and \emph{dog}. Here \emph{car} and \emph{person} are two classes seen in Cityscapes (i.e., \emph{person} $\sim$ \emph{pedestrian} in Cityscapes), and \emph{sofa} and \emph{dog} are {\bf unseen} categories. 
From results in Table~\ref{tbl:human_exp}, we observe that the humans were still faster when using our tool, but less so, as expected. Agreement in terms of IoU is also lower here, indicating potential biases by annotators to accept less-than perfect predictions when more corrections are needed. We are developing a more complete annotation tool, and plan to investigate such phenomena on a larger scale.

\begin{table}[t!]
\vspace{-0mm}
\begin{center}
\footnotesize
\addtolength{\tabcolsep}{-2pt}
\begin{tabular}{|l|c|c||c|c|}
	\hline
	& \multicolumn{2} {c||}{Cityscapes} & \multicolumn{2} {c|}{ADE}\\
	\hline
	& Time (s) & IoU (\%) & Time (s) & IoU (\%)\\
	\hline
	manual & 39.7 & 76.2 & 29.2 & 80.63\\
	with PolyRNN++ & 14.7 & 75.4 & 19.3 & 75.9 \\
	\hline
\end{tabular}
\vspace{-2mm}
   \caption{\footnotesize{\bf Real Human Experiment}: {\bf In-domain} on 50 randomly chosen Cityscapes car instances (left) and {\bf Out-of-domain} on 40 randomly chosen ADE20K instances (right). No fine-tuning was used in ADE experiment.}
\label{tbl:human_exp}
\vspace{-4mm}
\end{center}
\end{table}

\vspace{-3.5mm}
\paragraph*{Limitations: } 
Our model predicts only one polygon per box and typically annotates the more central object. 
If one object
breaks the other, our approach tends to predict the occluded object as a single polygon.
As a result, current failures cases are mostly around big multi-component objects.
Note also that we do not handle holes which do not appear in any of our tested datasets. 
In interactive mode, we would greatly benefit by allowing the human to add/remove points.
\vspace{-4.5mm}
\section{Conclusion}
\label{sec:conc}
In this paper, we proposed Polygon-RNN++, a model for object instance segmentation that can be used to interactively annotate segmentation datasets. The model builds on top of Polygon-RNN~\cite{polyrnn}, but introduces several important improvements that significantly outperform the previous approach in both, automatic and interactive modes. We further show robustness of our model to noisy annotators, and show how it generalizes to novel domains. We also show that with a simple online fine-tuning scheme, our model can be used to effectively adapt to novel, out-of-domain datasets.
\vspace{-0mm}

\section*{Acknowledgement}
We gratefully acknowledge support from NVIDIA for their donation of several GPUs used for this research. We also thank Kaustav Kundu for his help and advice, and Relu Patrascu for infrastructure support.

{\small
\bibliographystyle{ieee}
\bibliography{egbib}
}

\vspace{4mm}
\section{Appendix}

\begin{figure*}[t!]
\begin{minipage}[t]{0.332\linewidth}
\centering
\includegraphics[width=\linewidth,height=3cm,clip,trim=35 0 65 30,clip]{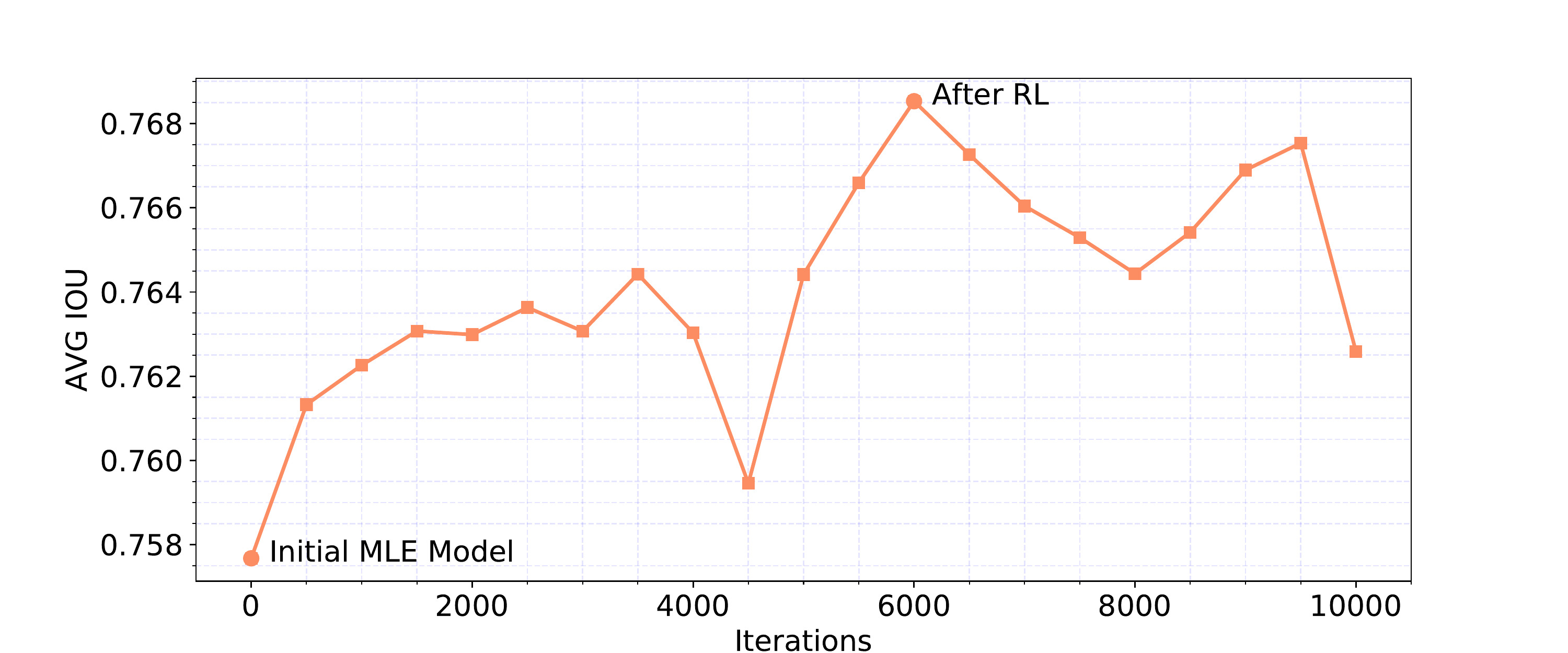}\\[0.5mm]
\subcaption{ Average IOU Inference Greedy Pass}
\end{minipage}
\begin{minipage}[t]{0.332\linewidth}
\centering
\includegraphics[width=\linewidth,height=3cm,trim=20 0 25 30,clip]{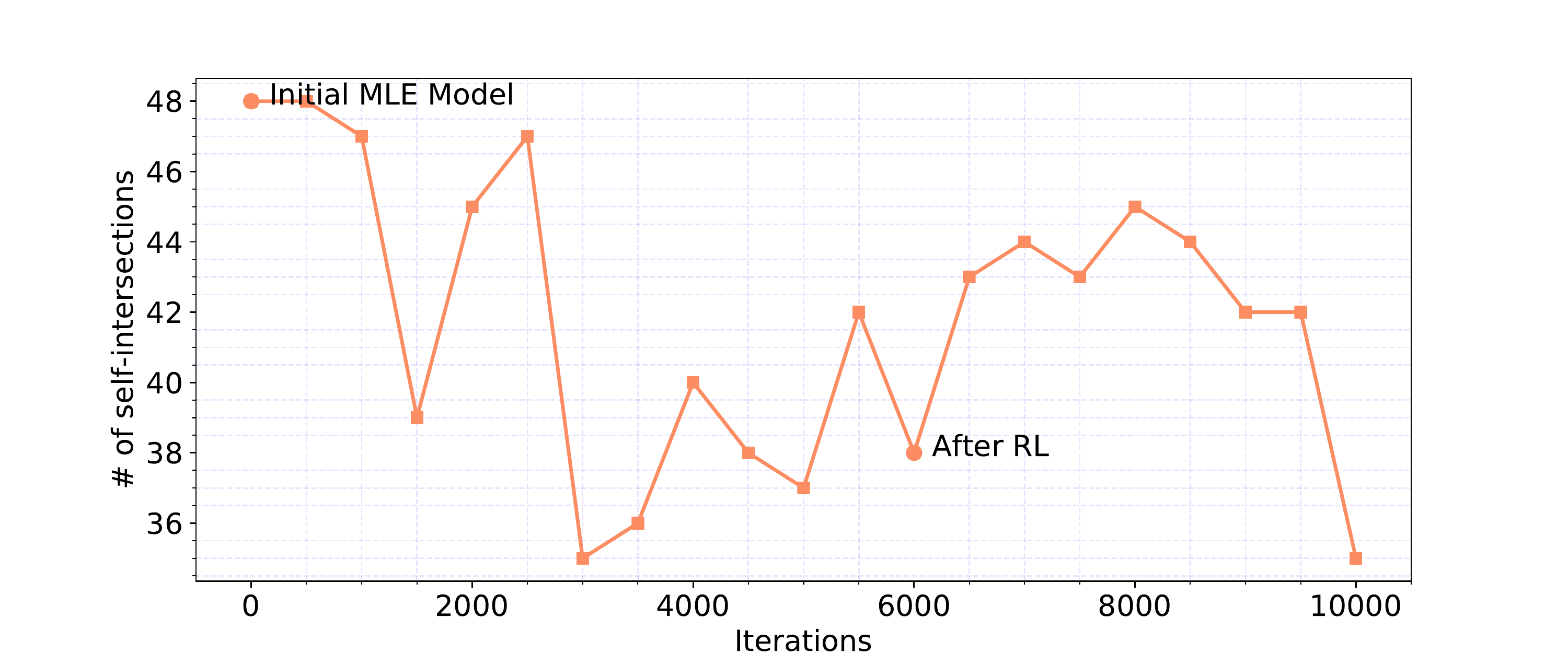}\\[0.5mm]
\subcaption{Number of Self-Intersections}
\end{minipage}
\begin{minipage}[t]{0.332\linewidth}
\centering
\includegraphics[width=\linewidth,height=3cm,trim=50 0 75 30,clip]{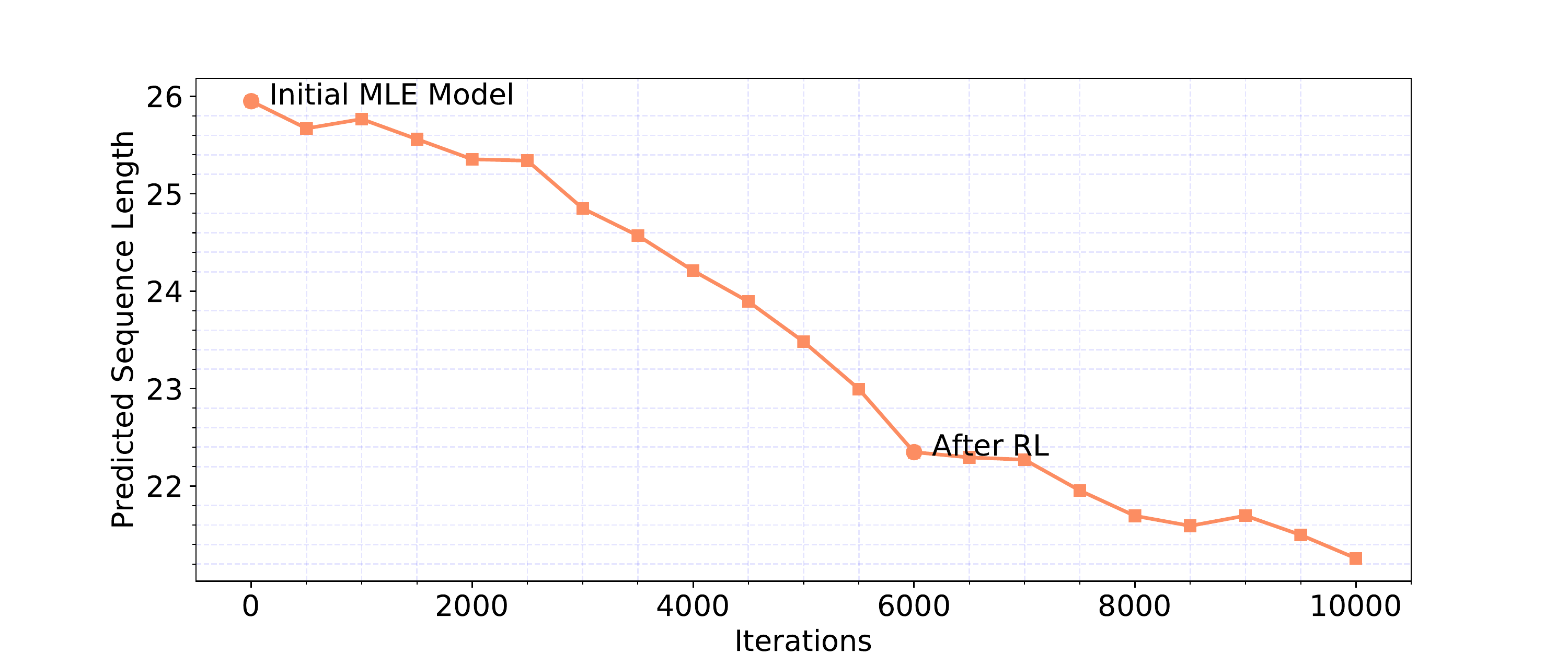}\\[0.5mm]
\subcaption{Polygons AVG Length}
\end{minipage}
\caption{Performance during RL training on our \textit{validation} set on the Cityscapes dataset.}
\label{fig:RL}
\end{figure*}

\begin{figure*}[h]
\begin{minipage}[t]{0.5\linewidth}
\centering
\includegraphics[width=1\linewidth,trim=30 00 70 32,clip]{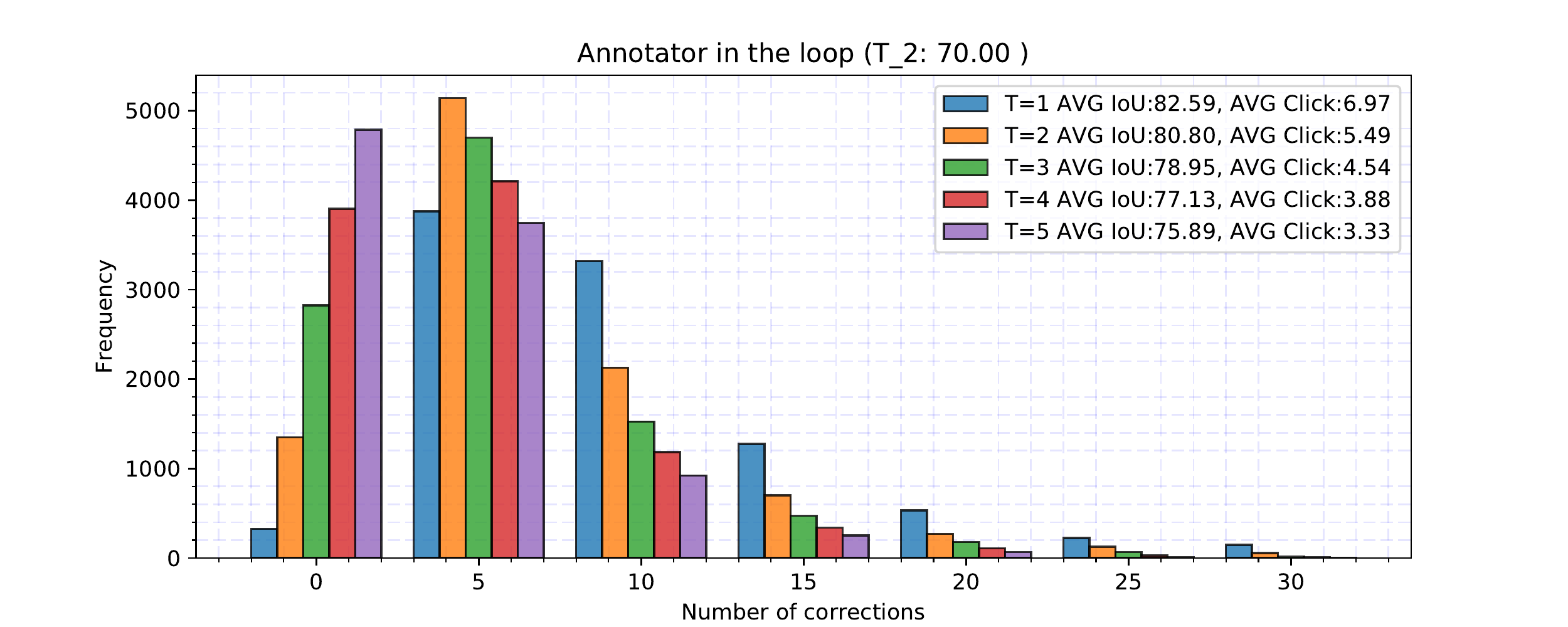}\\[0.5mm]
\subcaption{ $T_2=0.7$}
\end{minipage}
\begin{minipage}[t]{0.5\linewidth}
\centering
\includegraphics[width=1\linewidth,trim=30 00 70 32,clip]{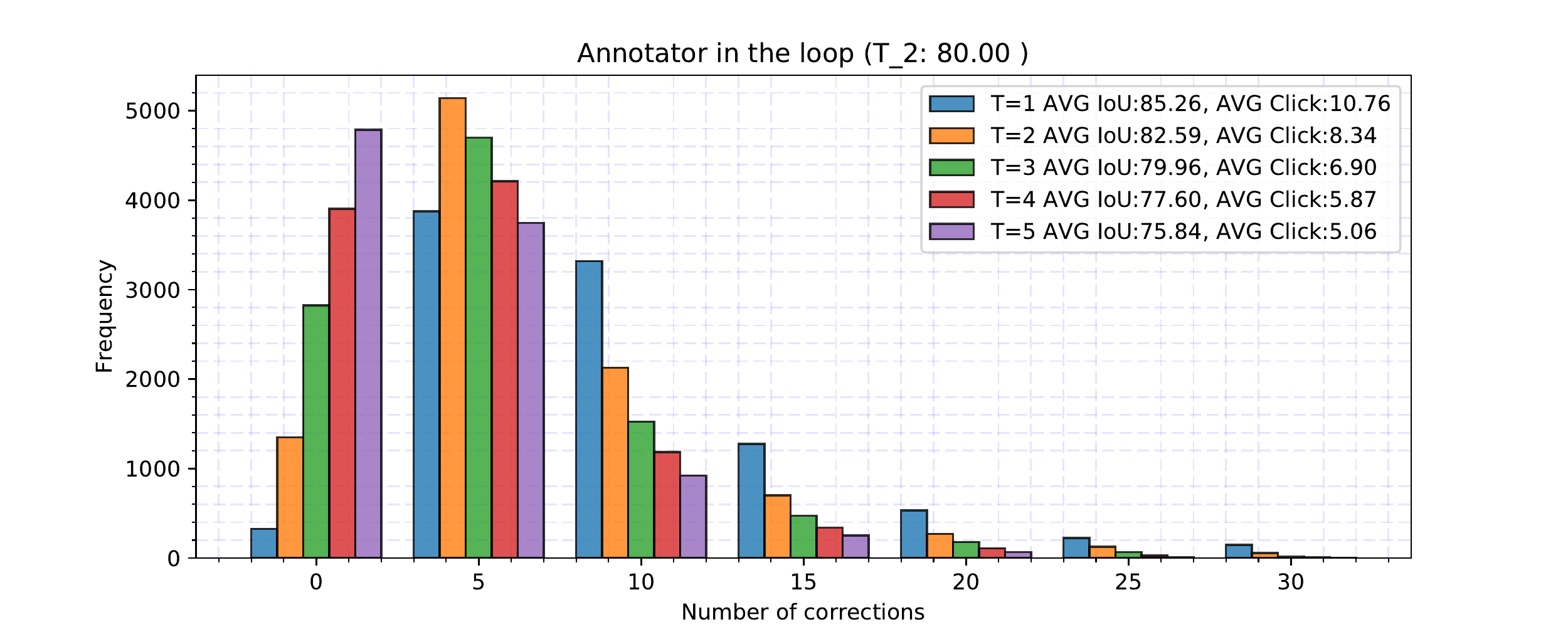}\\[0.5mm]
\subcaption{ $T_2=0.8$}
\end{minipage}

\vspace{3mm}
\begin{minipage}{0.5\linewidth}
\centering
\includegraphics[width=1\linewidth,trim=30 00 70 32,clip]{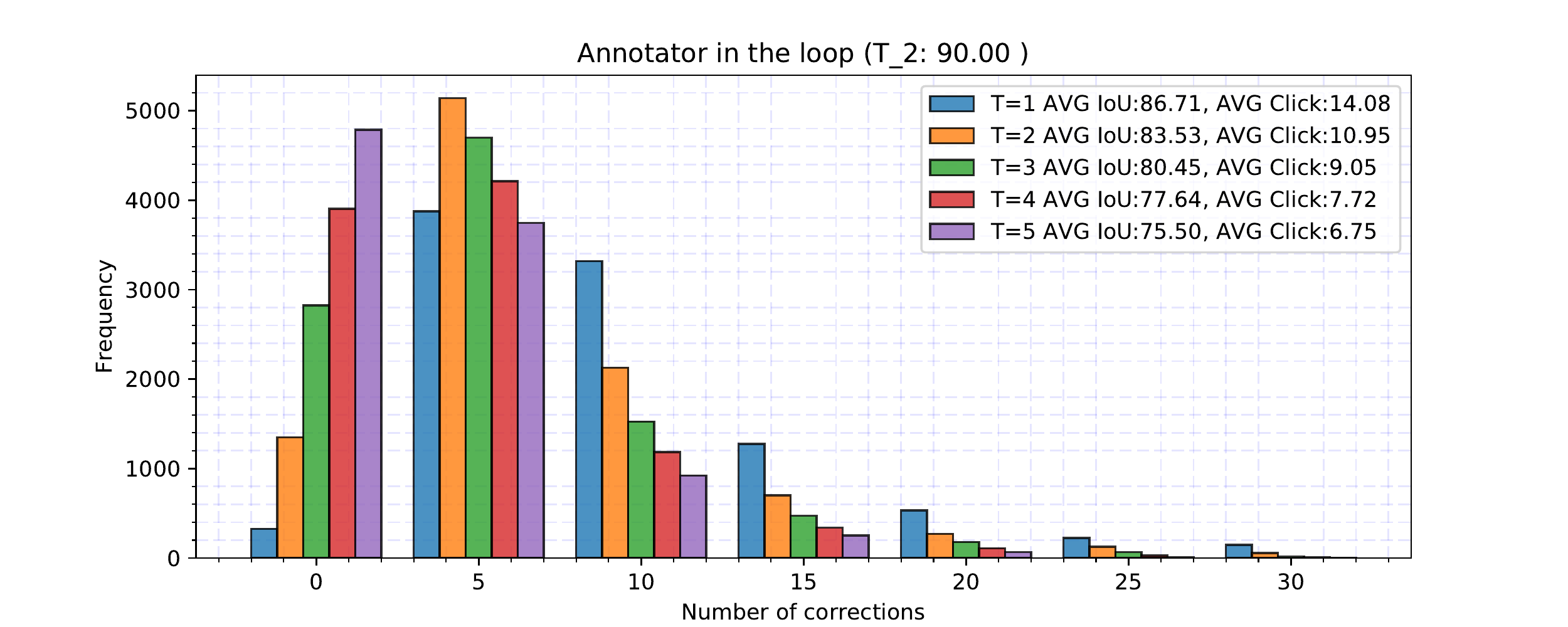}\\[0.5mm]
\subcaption{ $T_2=0.9$}
\end{minipage}
\begin{minipage}{0.5\linewidth}
\centering
\includegraphics[width=1\linewidth,trim=30 00 70 32,clip]{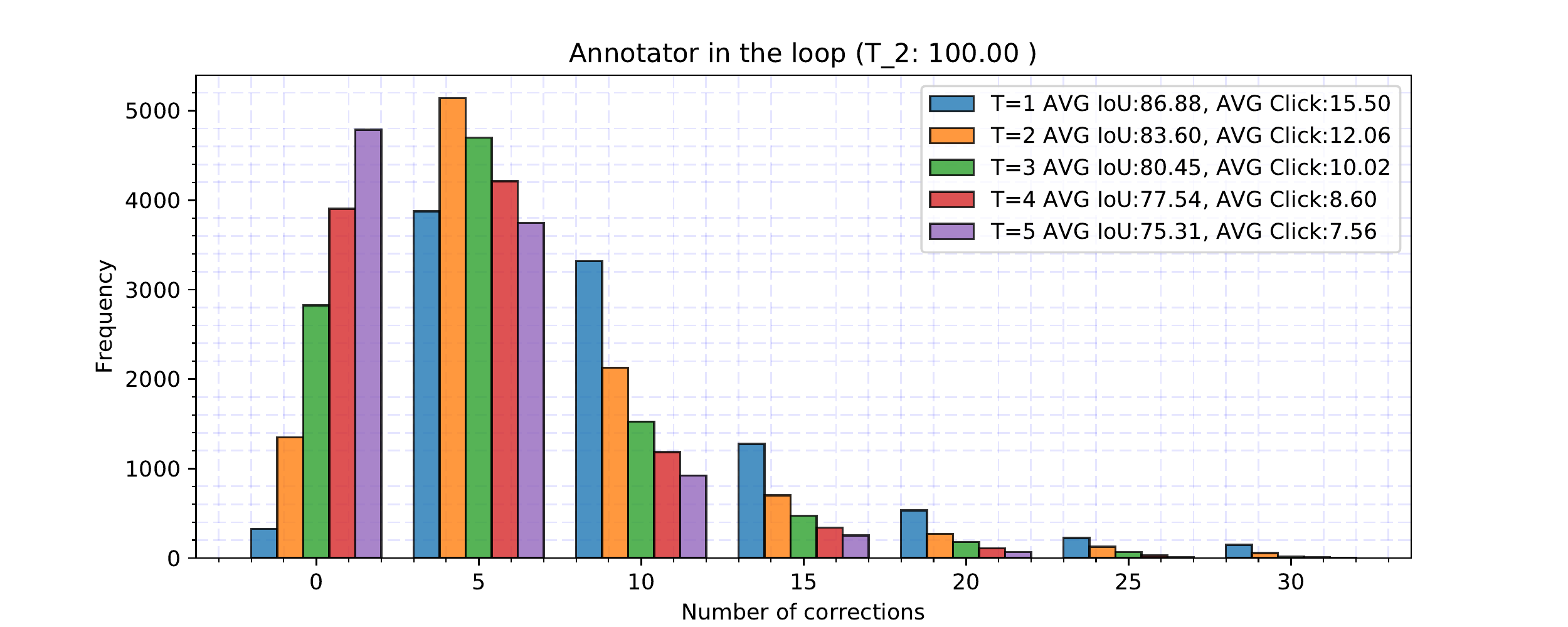}\\[0.5mm]
\subcaption{ $T_2=1.0$}
\end{minipage}
\caption{Interactive Mode on Cityscapes for different values of $T_2$}
\label{fig:hist}
\end{figure*}

In the appendix, we provide additional qualitative and quantitative results for our Polygon-RNN++ model.
In particular, we show an additional ablation study  %
that emphasizes the importance of the evaluator network. 
We also illustrate the performance of our model across the course of training during the RL training phase. We provide further details on automatic full-image object instance segmentation. 
Finally, we provide several qualitative examples for both the automatic and interactive modes. 
\vspace{-3mm}

\paragraph{Training with RL.}
In Fig.~\ref{fig:RL}, we plot the performance in terms of mean IoU during RL training on Cityscapes \cite{cityscapes}. Additionally, we show the average length of polygons. Note that nothing in our model prevents predicting polygons that include self-intersection. We also investigate this issue in this plot.  In comparison with the MLE model, we can see that we obtain an increase in terms of mean IoU. Note also that by directly optimizing IoU, the average length of the predicted polygons and the number of polygons with at least one self-intersection decrease.
\vspace{-3mm}

\paragraph{Evaluator Network.} In Table~\ref{tbl_iou_rl_performance_with_multicomp} we compare different decoding strategies, and showcase the importance of the evaluator network. We separate two cases: handling multiple first vertex candidates, as well as multiple sequences (polygons) that follow from each first vertex candidate. We see that while beam search for both the first vertex and the full sequence outperforms greedy decoding, the improvement is minor ($0.5\%$, second row).  Following $K$ first vertices using greedy decoding for the sequence and using the evaluator network to choose between the $K$ polygons results in $2.2\%$ over beam search (third row). Using beam search for the sequence (and evaluator network to choose between the $K$ polygons in the end) further increases performance ($0.3\%$, fourth row). In the last row, we use beam search until the last predicted polygon vertex, and use the evaluator network to also choose between the $K$ last vertex candidates (for each first vertex candidate). This gets us another $0.2\%$ (last row).
\vspace{-3mm}

\begin{table*}[h]
{\small
\begin{center}
\begin{tabular}{|l|c|c|c|c|c|c|c|c|c|c|}
\hline
First Vertex & Sequence & Bicycle & Bus & Person & Train & Truck & Motorcycle & Car & Rider & Mean \\
\hline\hline
  Greedy & Greedy   &  {57.57}  &  75.74   & {68.78}   &   {59.40}  &  75.97  &  {58.19}   &  {75.88}   & {65.47} & 67.13 \\
  BeamSearch (BS) & BS   &  {58.50}  &  74.43   & {68.68}   &   {61.73}  &  75.90  &  {58.44}   &  {75.51}   & {66.73} & 67.49 \\
\hline  
 Eval. Net & Greedy  & {61.62}  &  {79.31}  &  {70.37}   & {62.17}   &   {77.45}  &   {60.71}  &  {77.80}   & {68.30}   &  {69.72}  \\
 Eval. Net & BS &  {62.04}  &  {79.19}  &   \bf{70.87}   & {62.61}   &    {77.81}  &   \bf{62.00}  &  {77.87}   &   {68.25}   & {70.08}  \\
 Eval. Net & BS- Eval. Net &  \bf{62.34}  & \bf{79.63}   &  {70.80}   & \bf{62.82}   &  \bf{77.92}    &  {61.69}  & \bf{78.01}  &  \bf{68.46}   &   \bf{70.21} \\
\hline
\end{tabular}
\end{center}
}
\vspace{-5mm}
\caption{The role of the evaluation network. Here \emph{Greedy} denotes greedy decoding of the polygon, and \emph{BS} indicates beam-search. We use $K=5$.  Performance (IoU in \%) in automatic mode for all Cityscapes classes.}
\label{tbl_iou_rl_performance_with_multicomp}
\end{table*}

\paragraph{Output Resolution and  Sensitivity to T of GGNN.} Note that our GGNN formulation is efficient and can handle large output sizes. We experimented with two output resolutions, $112\times 112$, and $224\times 224$. The result for the $224\times 224$ was only $0.02\%$ better than that of $112\times 112$, but required longer training times. Thus, our choice in the paper is $112\times 112$. \\
In Table~\ref{tbl:sensitivityT} we report results for different $T$ (number of propagation steps), showing stable results across several options.

\begin{table}[t!]
\begin{center}
\small
\begin{tabular}{|l|c|c|c|}
	\hline
	$T_{GGNN}$ (\# of propagation steps) & 3 & 5 & 7 \\
	\hline
	AVG IoU (\%) & 71.37 & 71.38 & 71.46 \\
	\hline
\end{tabular}
\vspace{-2mm}
\caption{Performance (automatic mode) for different number of propagation steps ($T_{GGNN}$) in GGNN. Experiment done on Cityscapes. We report the performance averaged across all categories. }
\label{tbl:sensitivityT}
\end{center}
\vspace{-4mm}
\end{table}

\begin{figure*}[ht]
	\centering
	\begin{tabular}{c c}
	\bf{PolygonRNN++ (with GT boxes)} & \bf{Human Annotator}\\[1mm]
	\includegraphics[width=0.496\linewidth,trim=180 200 150 190,clip]{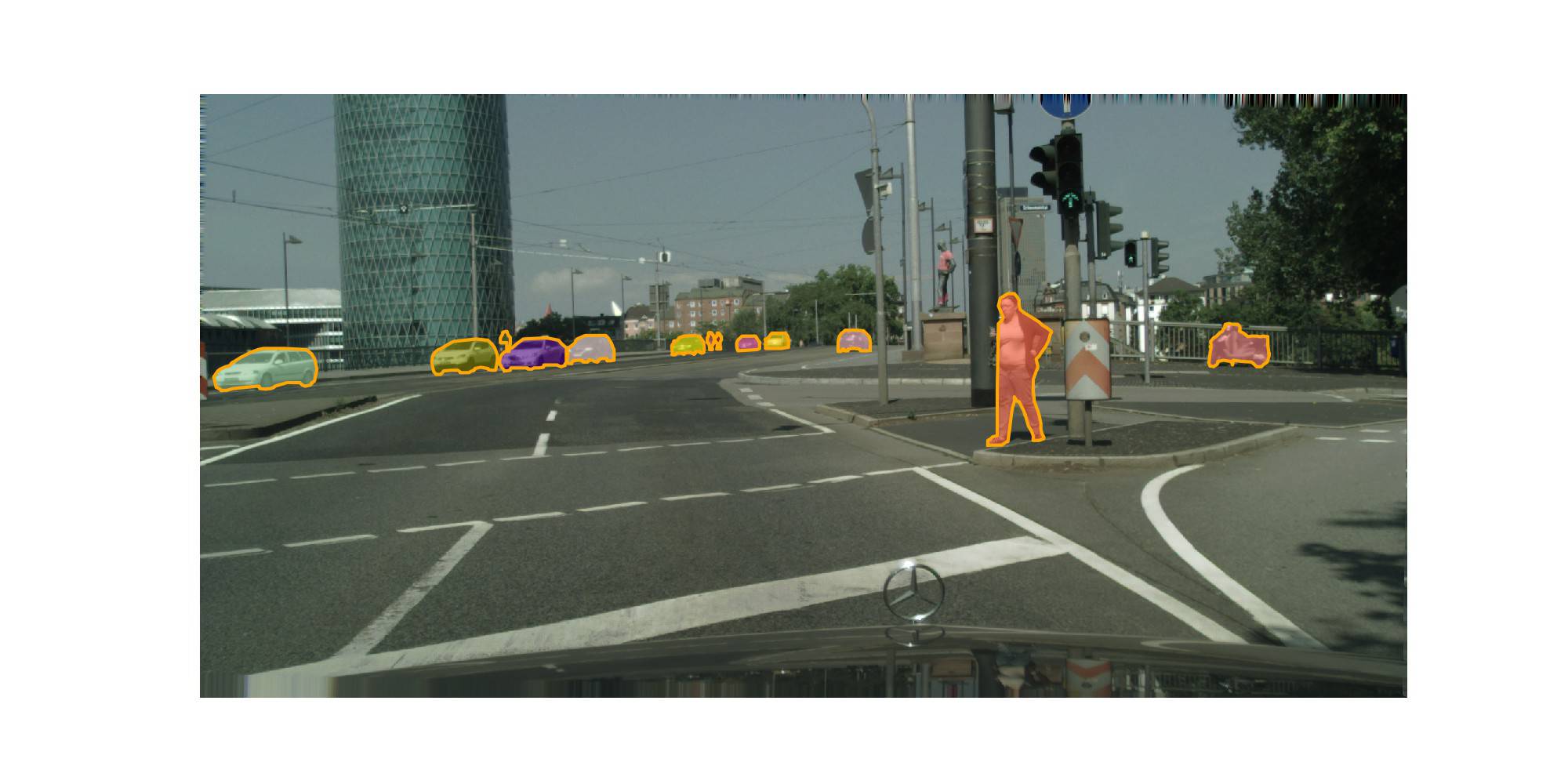} & \includegraphics[width=0.496\linewidth,trim=180 200 150 190,clip]{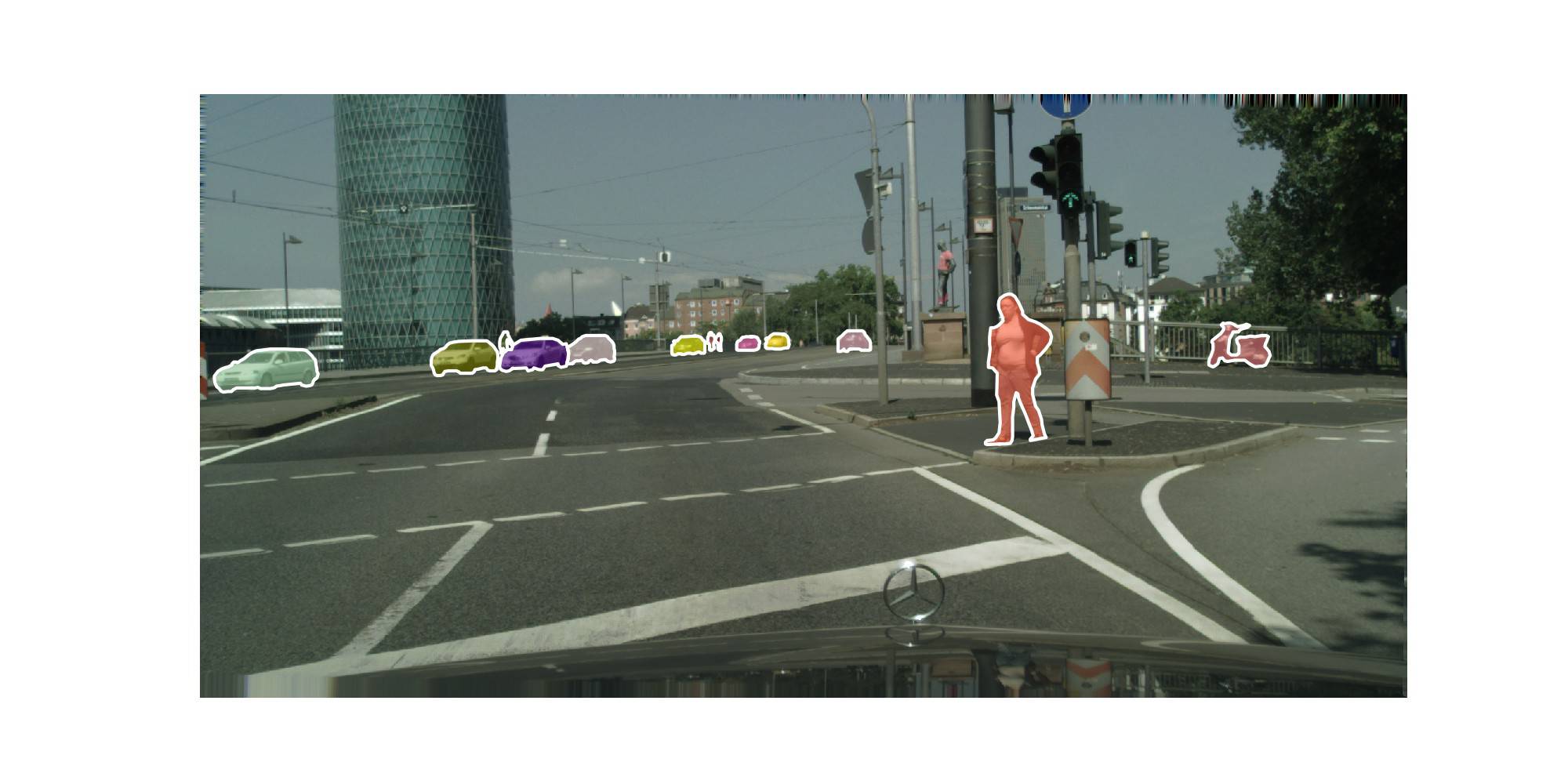} \\[-0.3mm] 
	\bf{0 clicks} & \bf{648 clicks}\\[2mm]

	\includegraphics[width=0.496\linewidth,trim=180 280 150 100,clip]{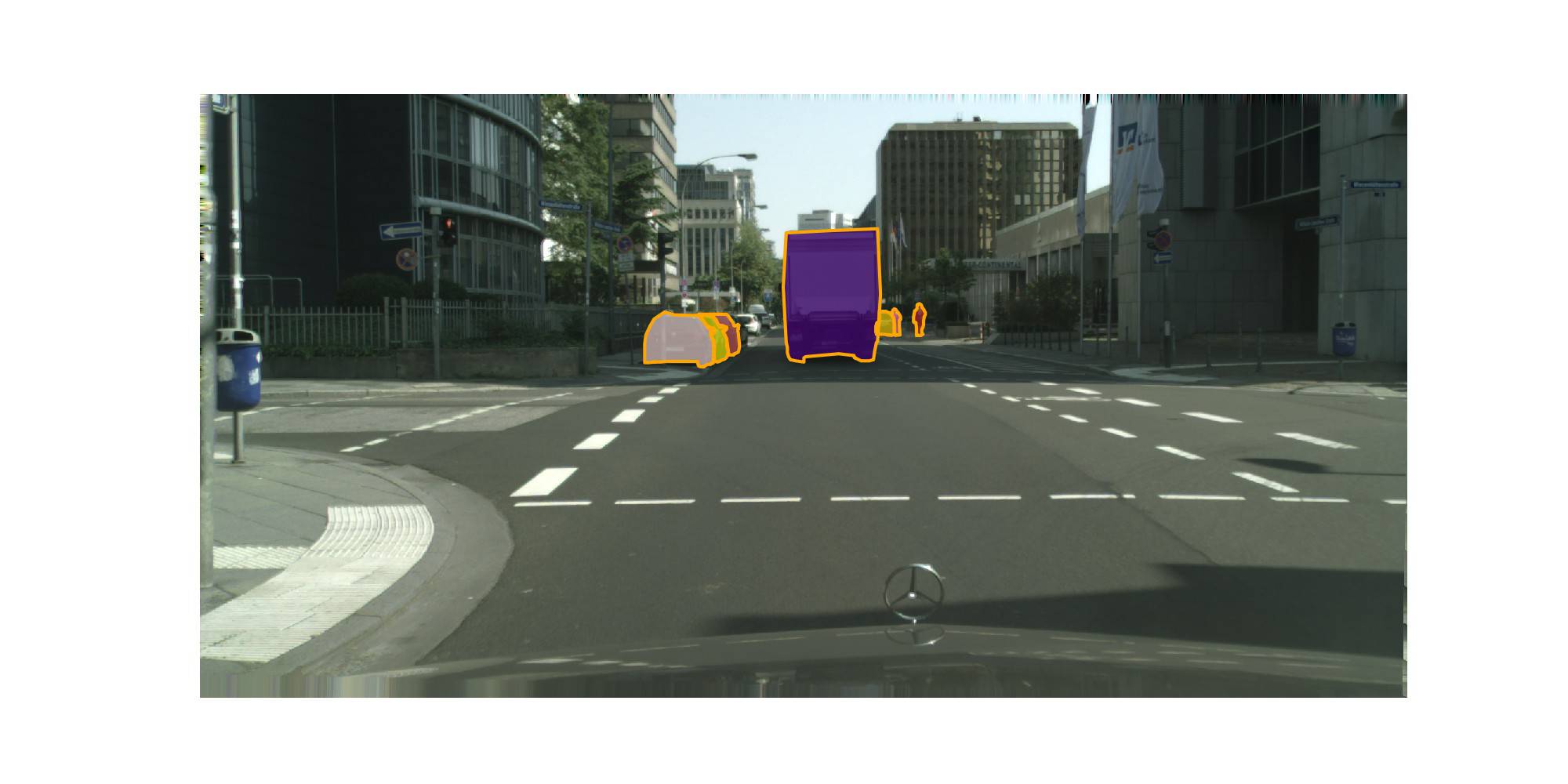} & \includegraphics[width=0.496\linewidth,trim=180 280 150 100,clip]{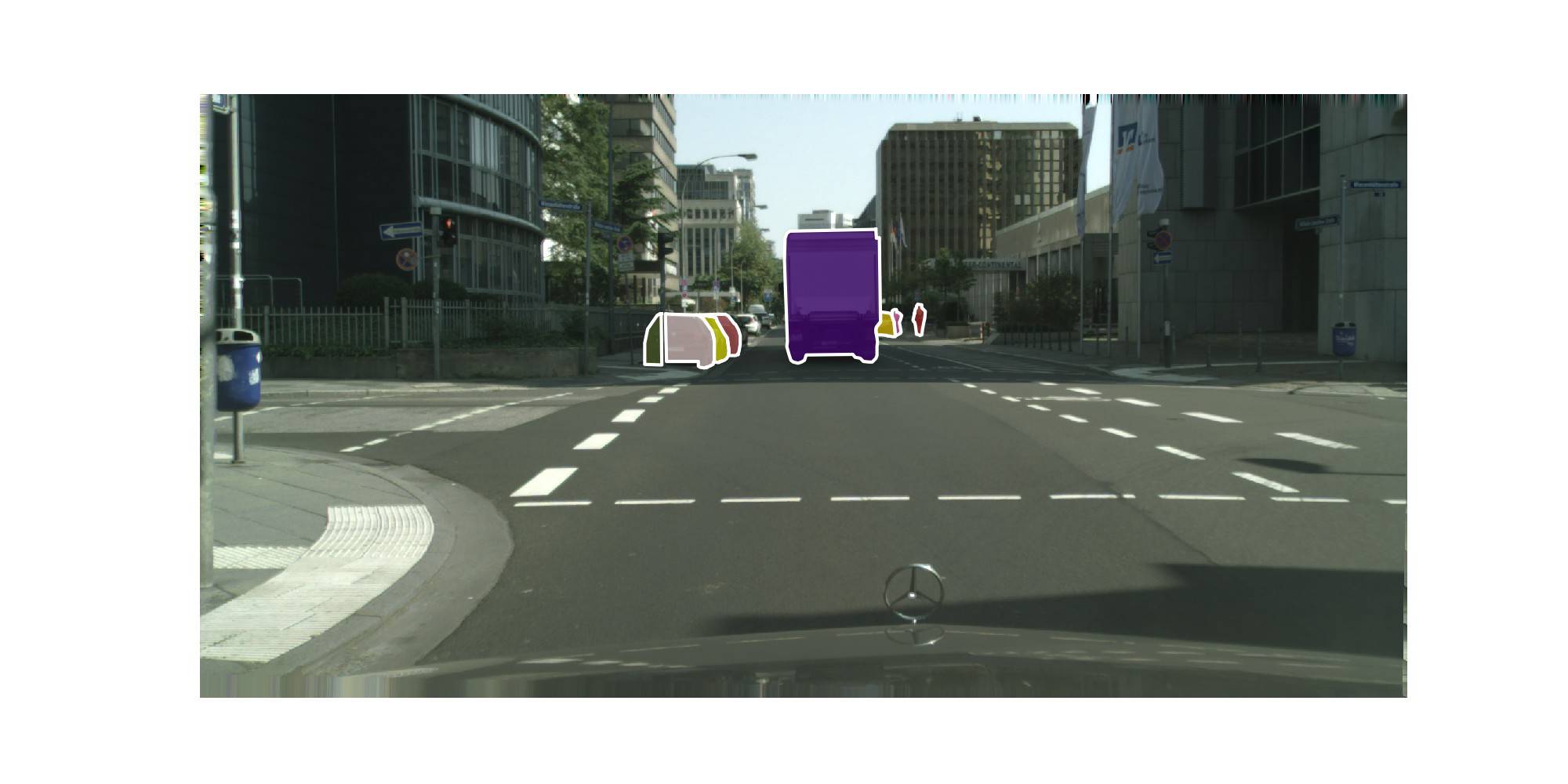}\\[-0.3mm] 
	\bf{0 clicks} & \bf{161 clicks}\\[2mm]

	\includegraphics[width=0.496\linewidth,trim=180 130 150 240,clip]{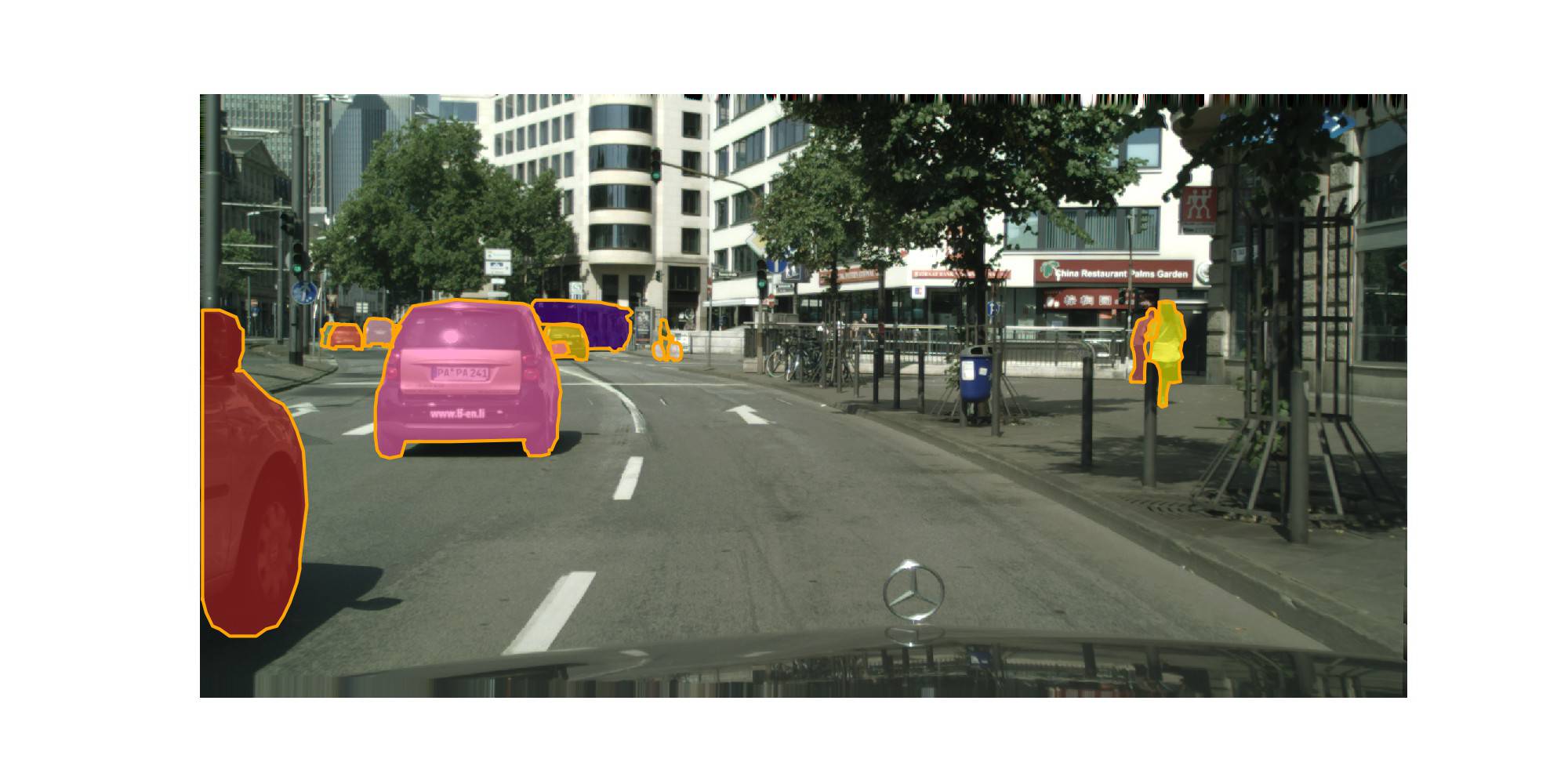} & \includegraphics[width=0.496\linewidth,trim=180 130 150 240,clip]{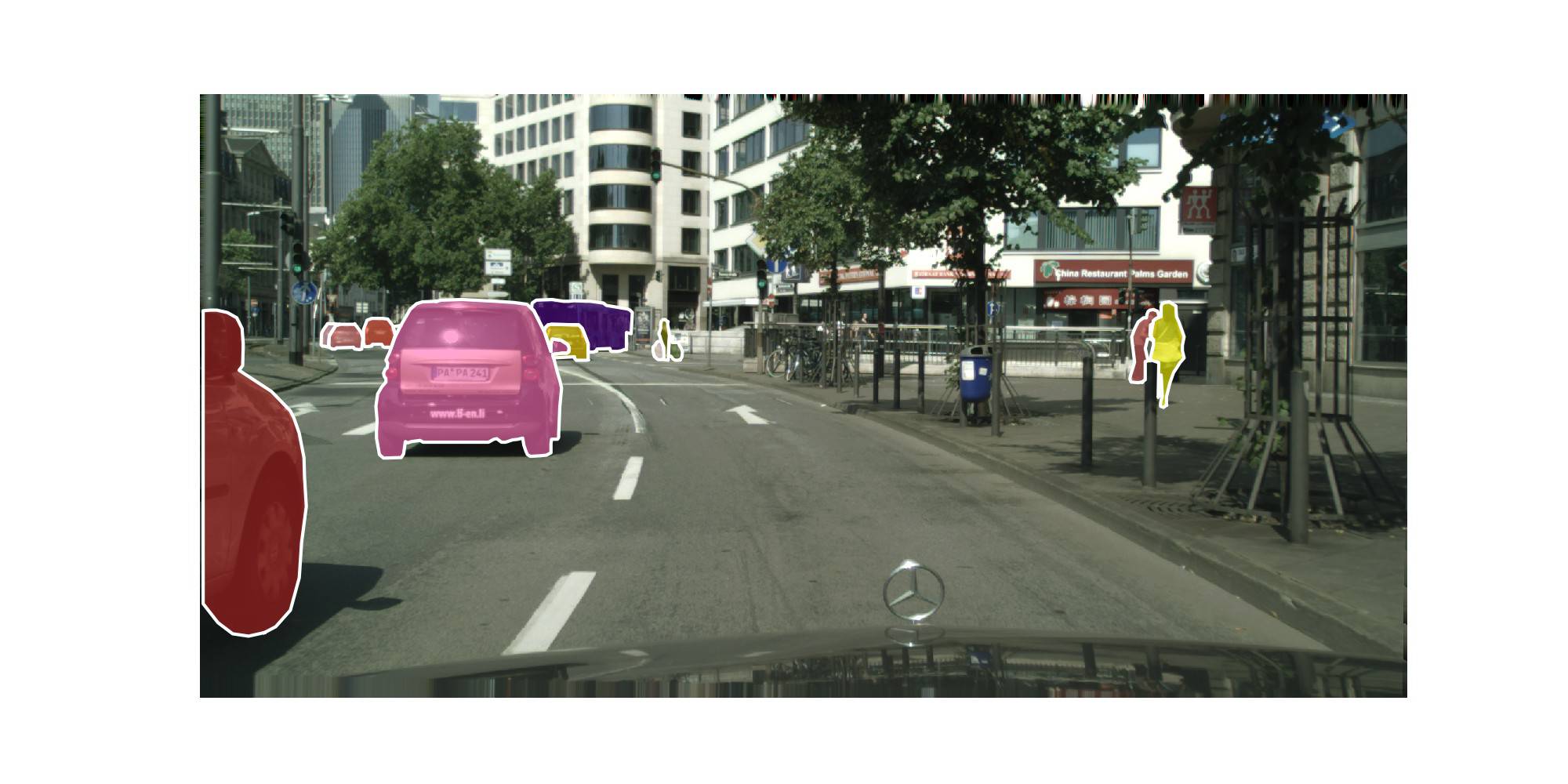}\\[-0.3mm] 
	\bf{0 clicks} & \bf{362 clicks}\\[2mm]
	
	\includegraphics[width=0.496\linewidth,trim=180 280 150 100,clip]{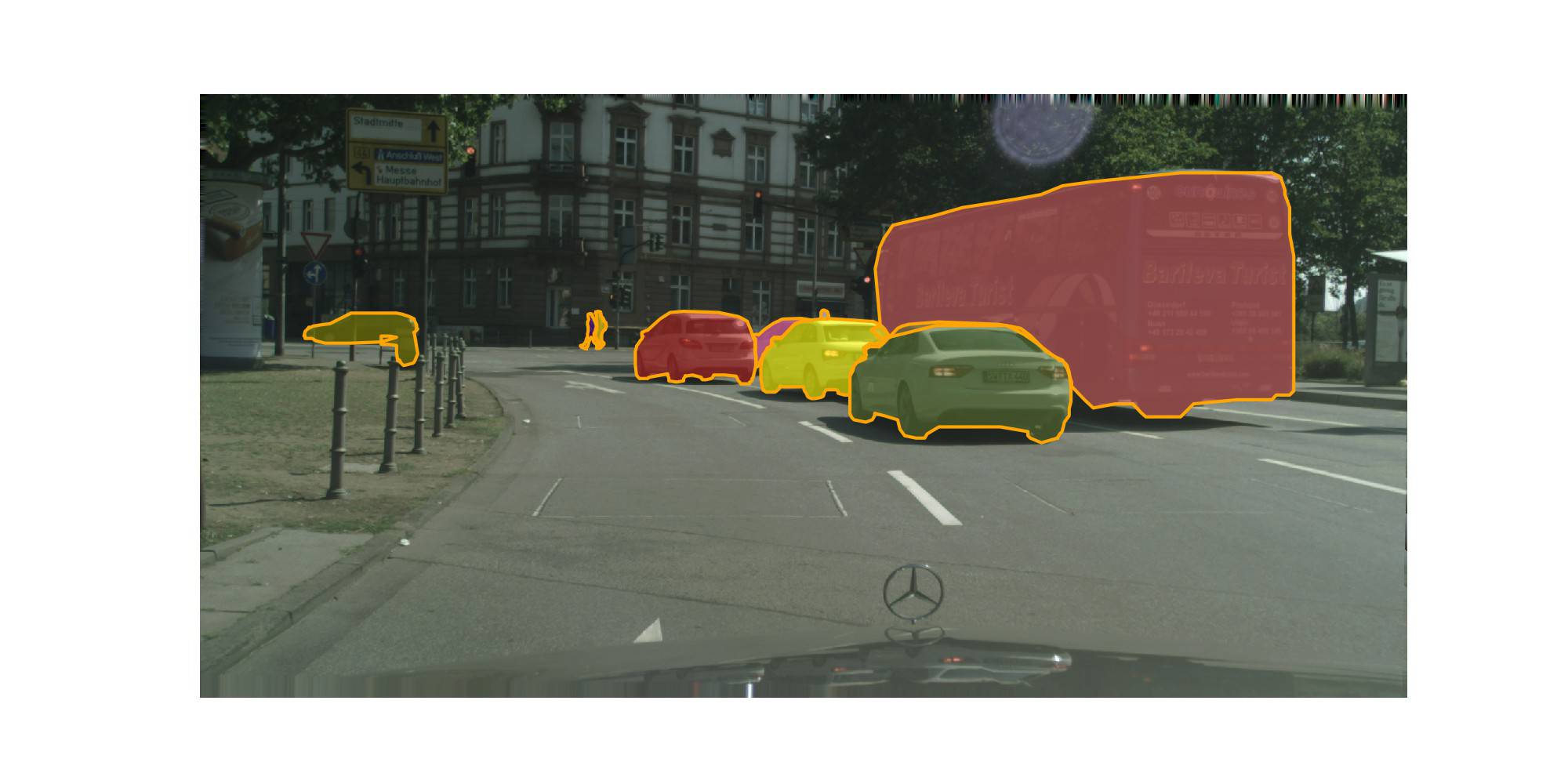} & \includegraphics[width=0.496\linewidth,trim=180 280 150 100,clip]{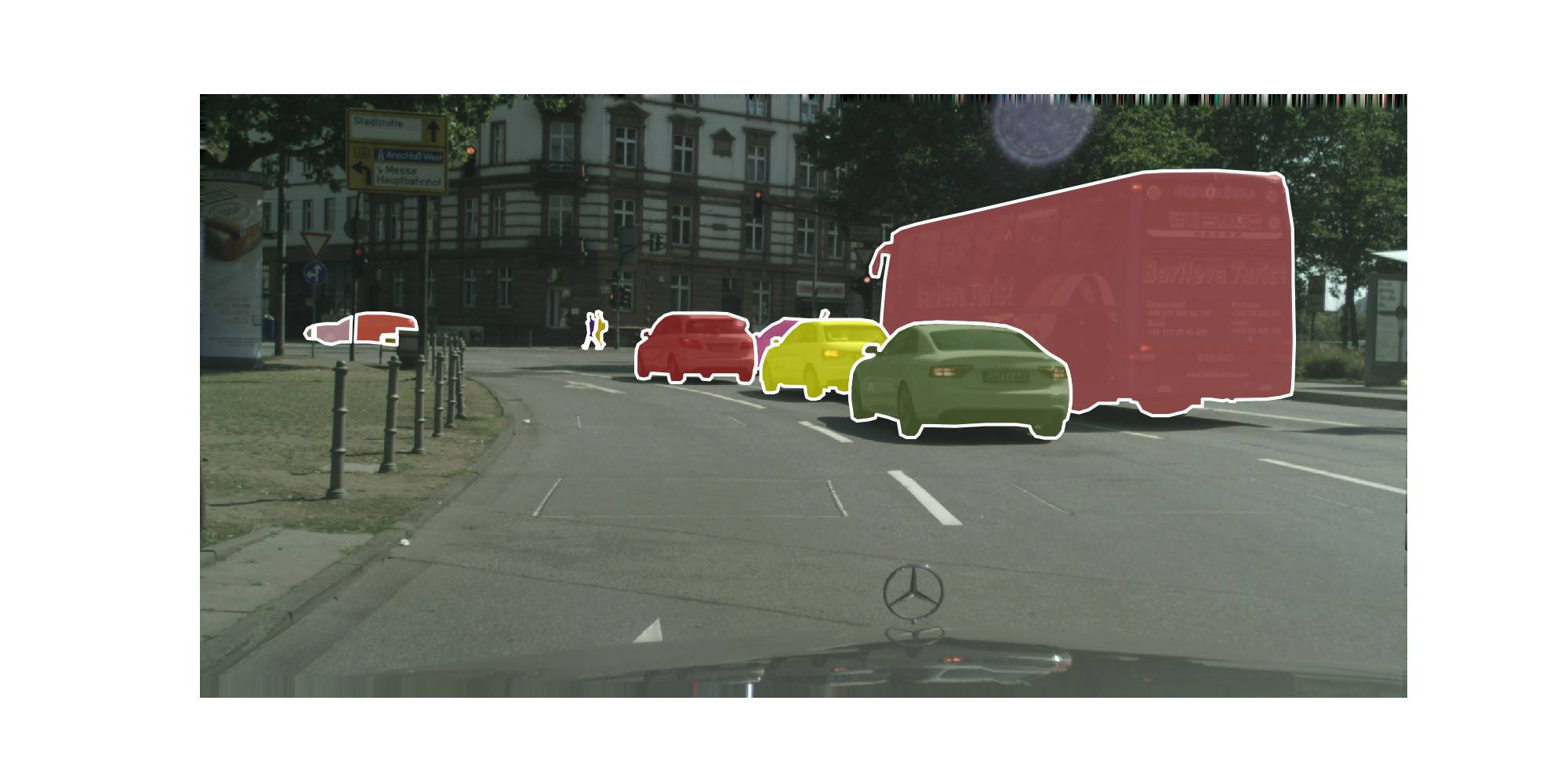}\\[-0.3mm] 
	\bf{0 clicks} & \bf{417 clicks}\\[2mm]
	
	\includegraphics[width=0.496\linewidth,trim=180 170 150 220,clip]{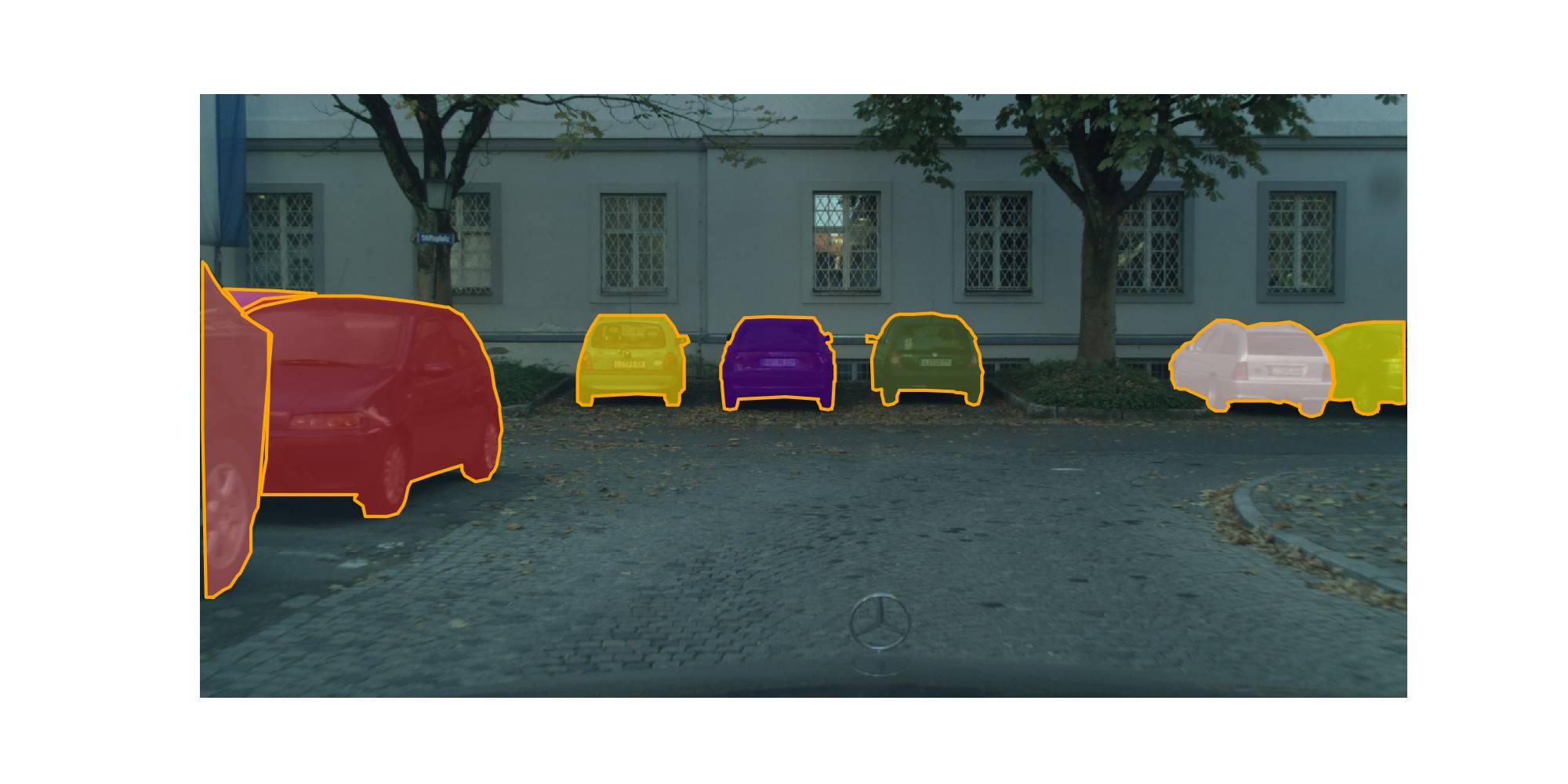} & \includegraphics[width=0.496\linewidth,trim=180 170 150 220,clip]{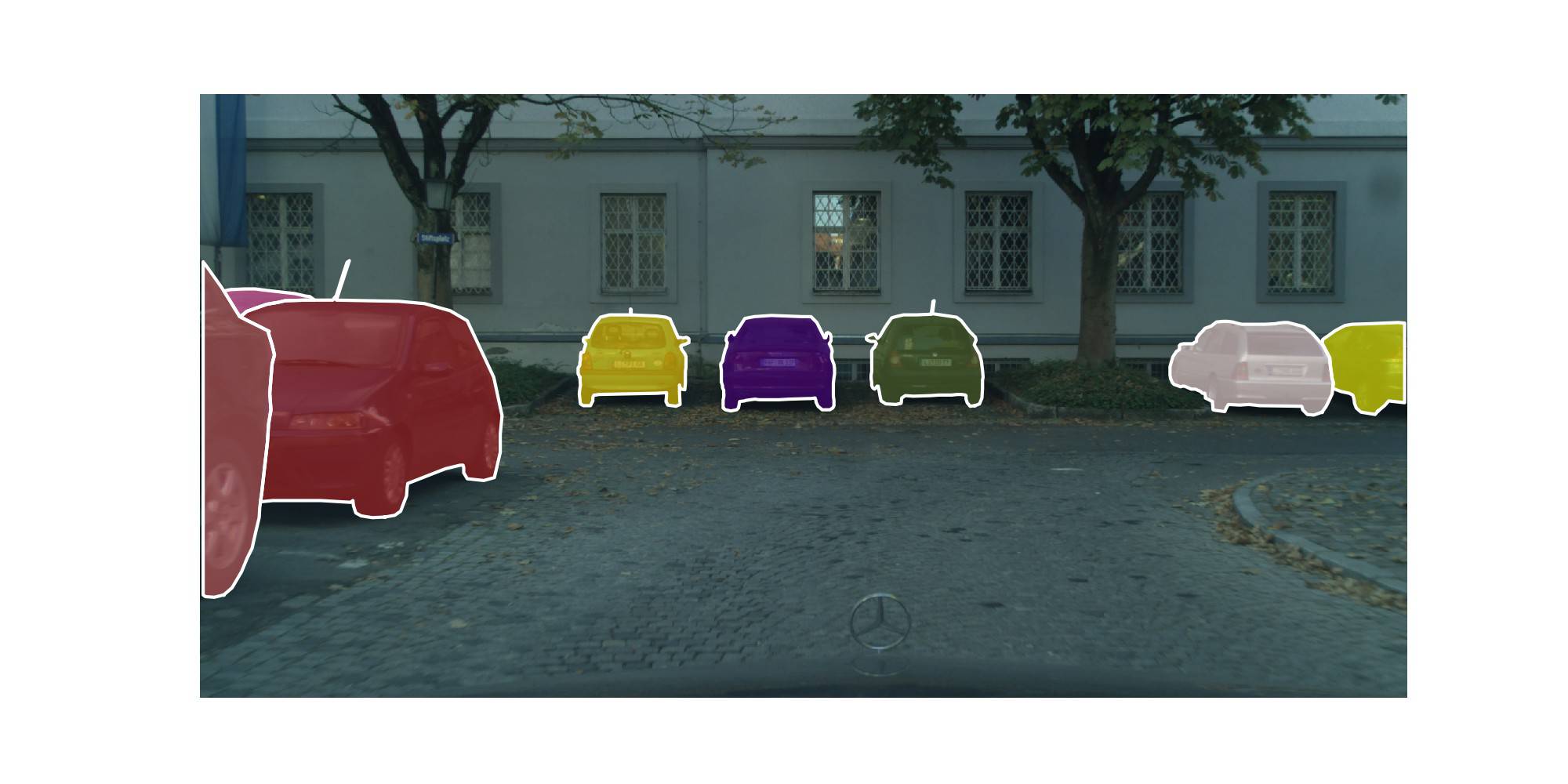}\\[-0.3mm] 
	\bf{0 clicks} & \bf{460 clicks}\\[2mm]

	\includegraphics[width=0.496\linewidth,trim=180 140 150 250,clip]{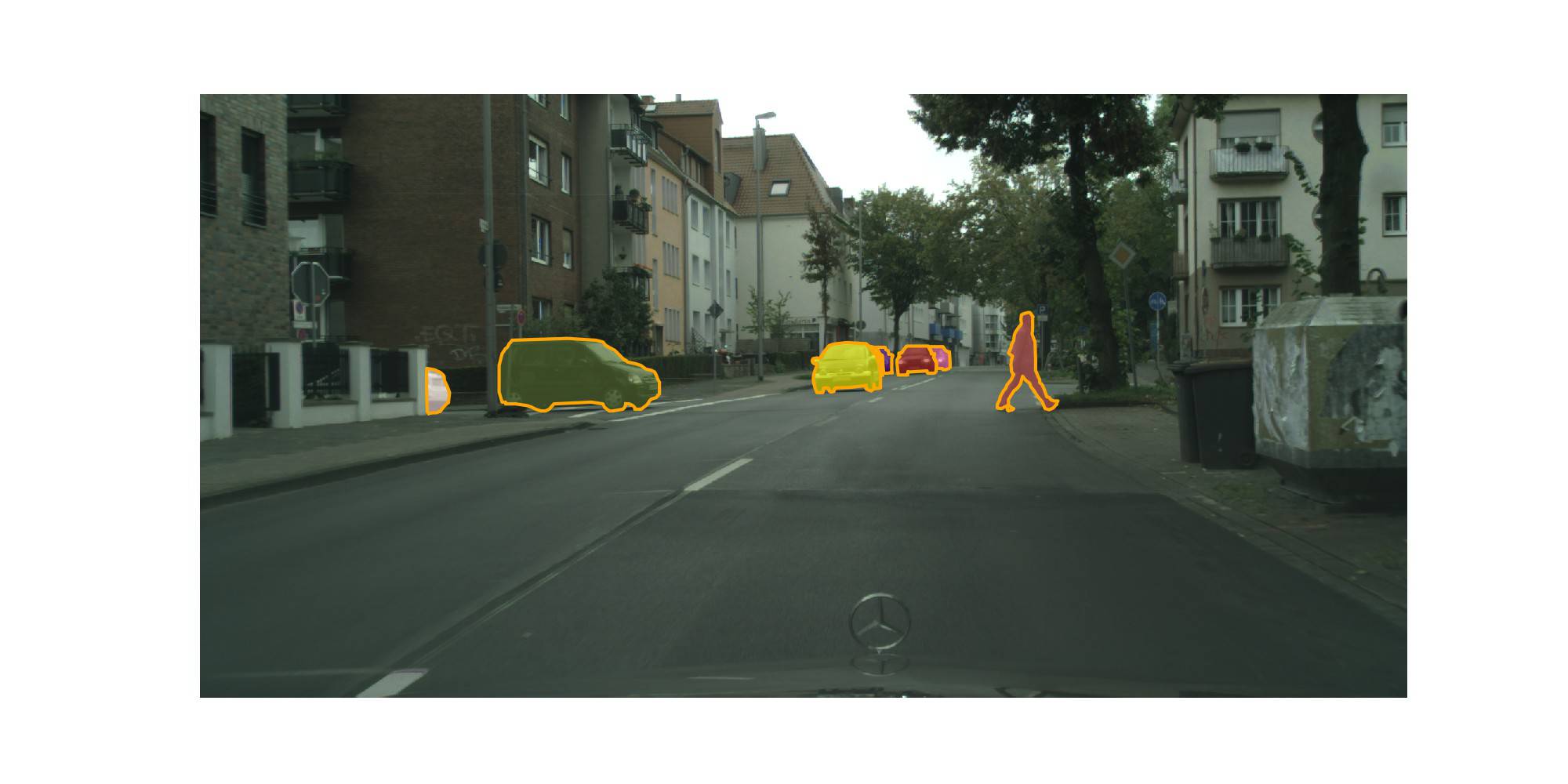} & \includegraphics[width=0.496\linewidth,trim=180 140 150 250,clip]{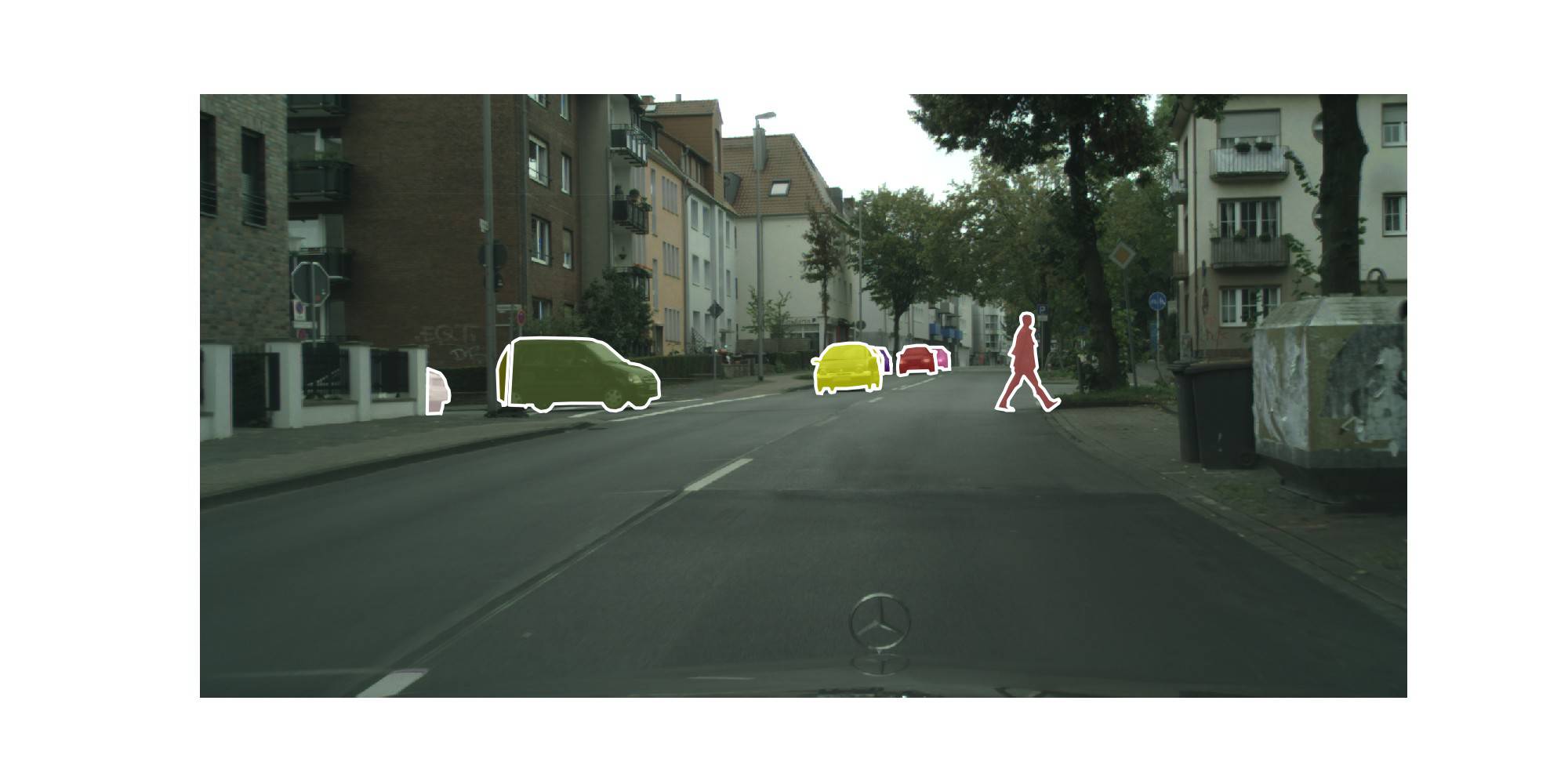}\\[-0.3mm] 
	\bf{0 clicks} & \bf{280 clicks}\\

\end{tabular}
\caption{{\bf Automatic mode on Cityscapes dataset}: Qualitative comparison between a human annotator vs PolygonRNN++ in automatic mode on Cityscapes. This model exploits GGNN to output a polygon at a higher resolution. Note that our model relies on bounding boxes.}
\label{fig:human_auto_cityscapes1}
\end{figure*}

\begin{figure*}[ht]
	\begin{tabular}{c c}
	\bf{PolygonRNN++ (with GT boxes)} & \bf{Human Annotator}\\[1mm]
	\includegraphics[width=0.496\linewidth,trim=180 200 150 190,clip]{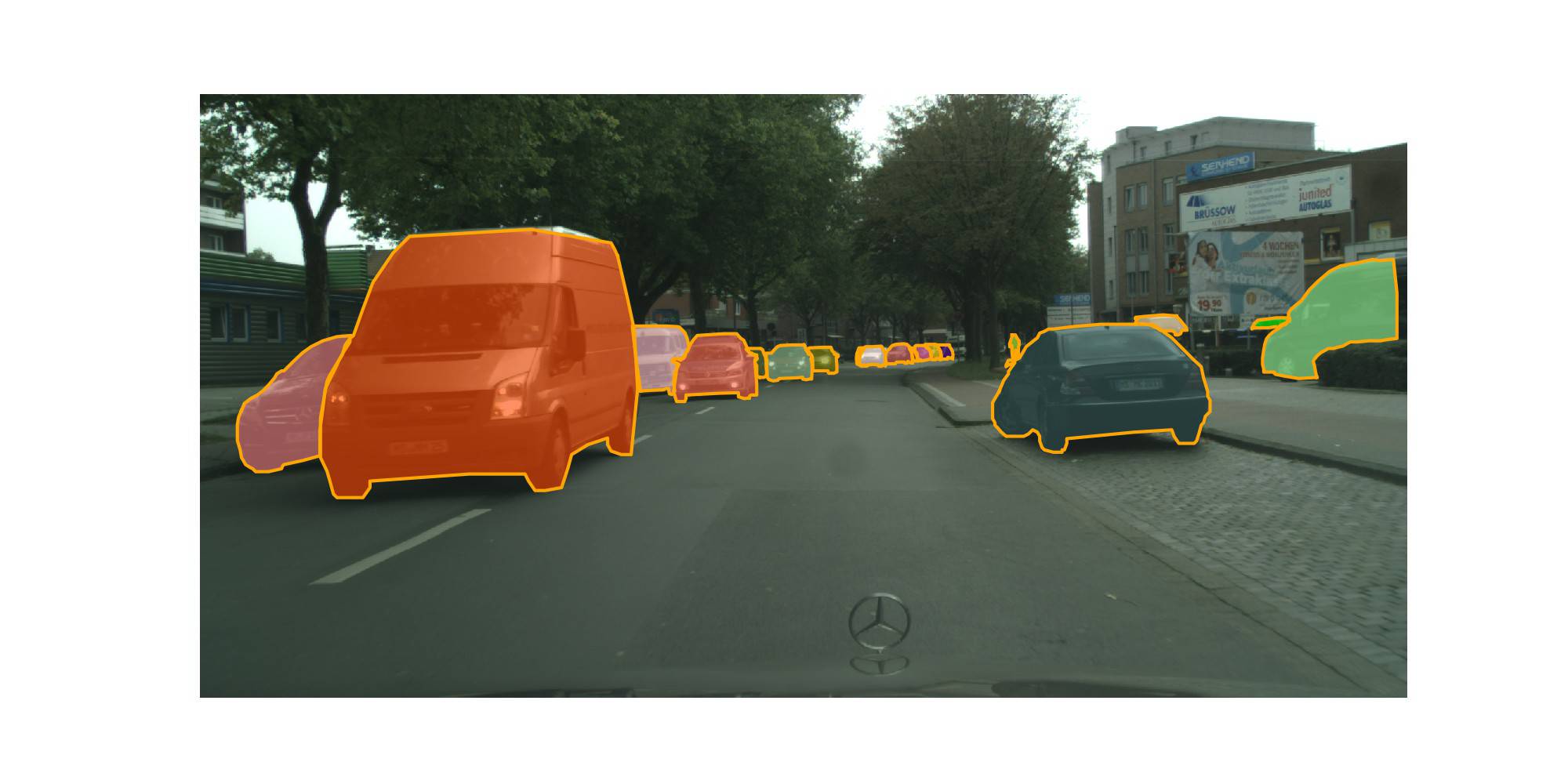} & \includegraphics[width=0.496\linewidth,trim=180 200 150 190,clip]{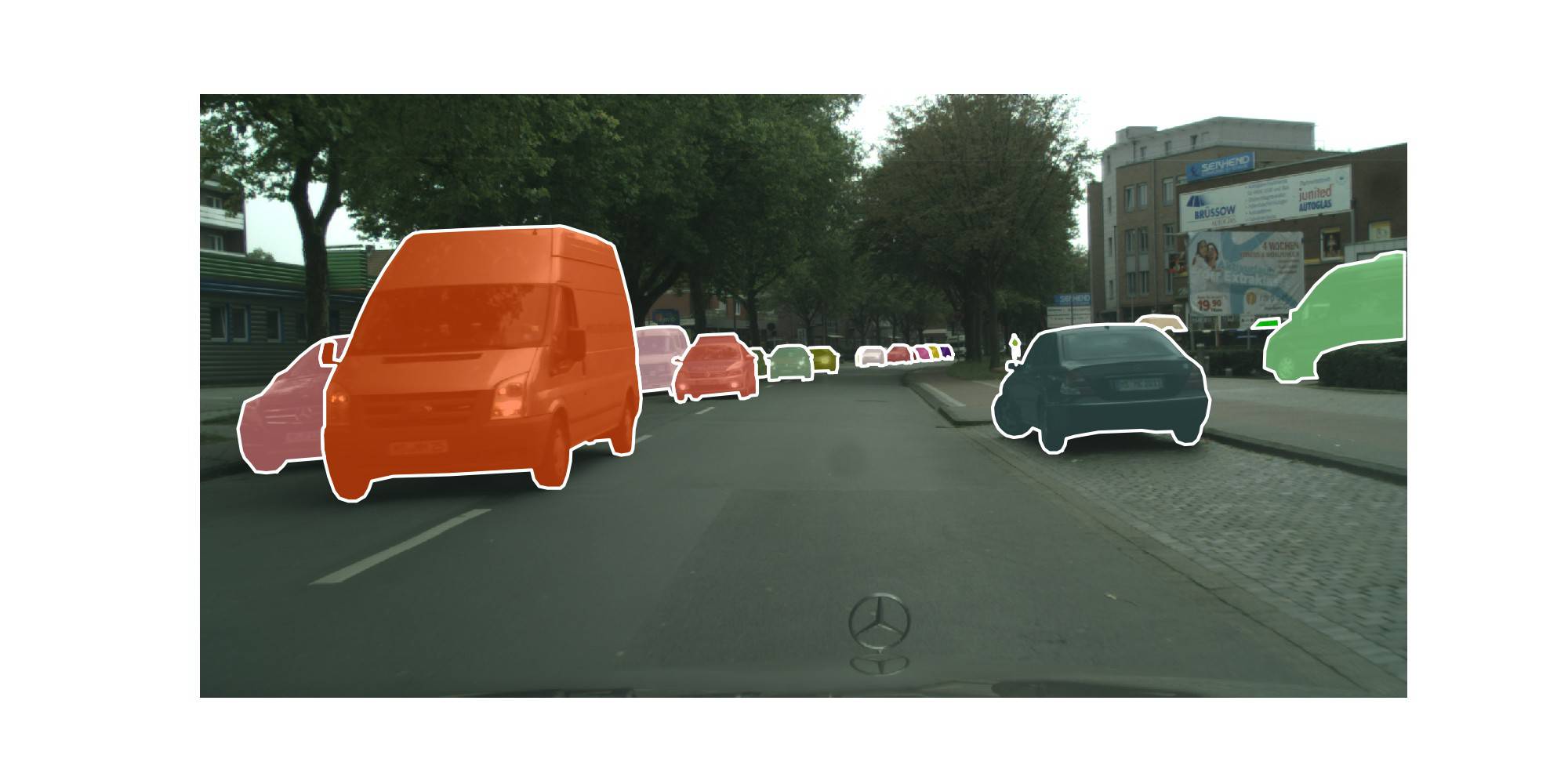}\\[-0.3mm] 
	\bf{0 clicks} & \bf{576 clicks}\\[2mm]

	\includegraphics[width=0.496\linewidth,trim=180 200 150 190,clip]{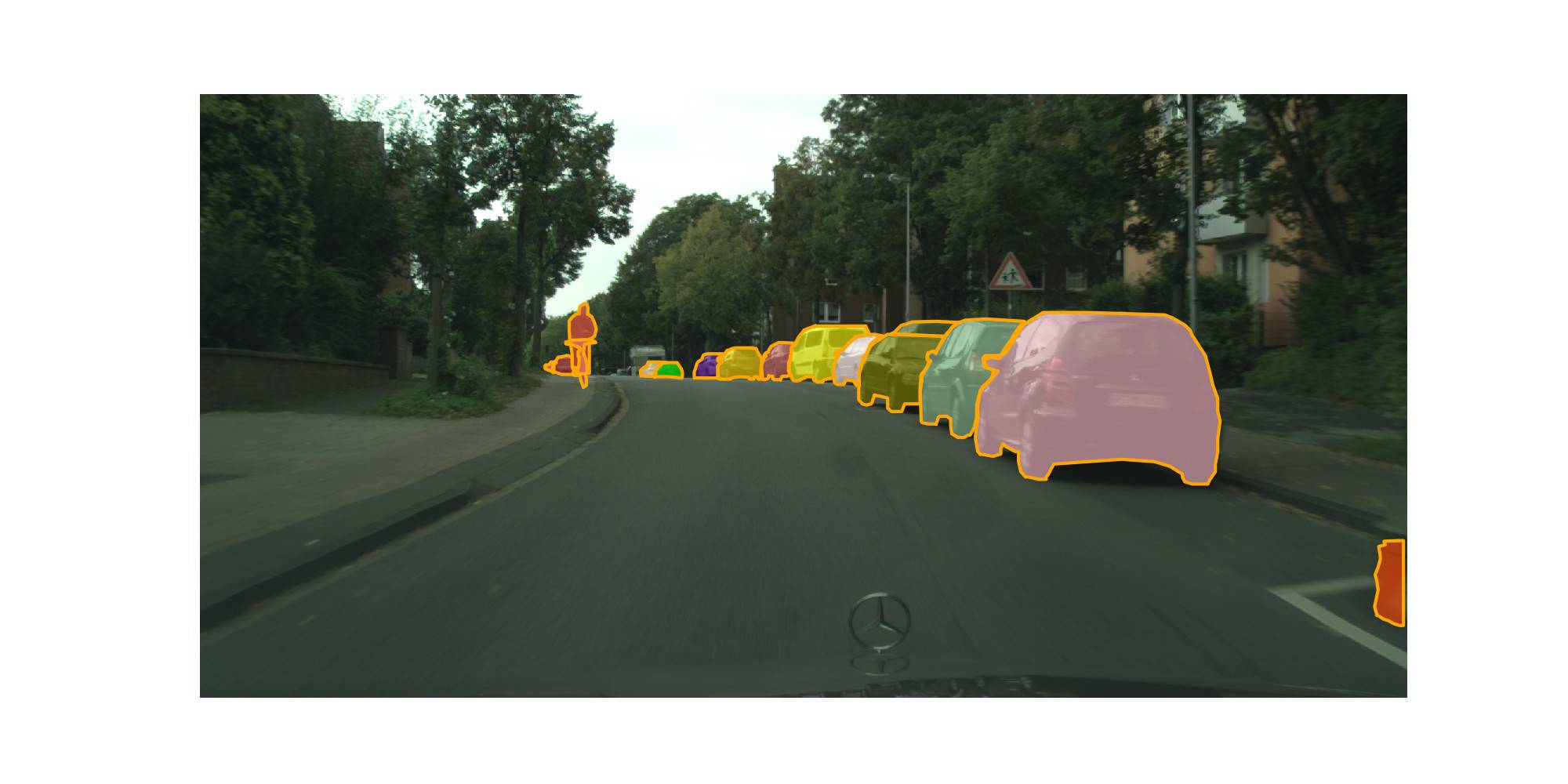} & \includegraphics[width=0.496\linewidth,trim=180 200 150 180,clip]{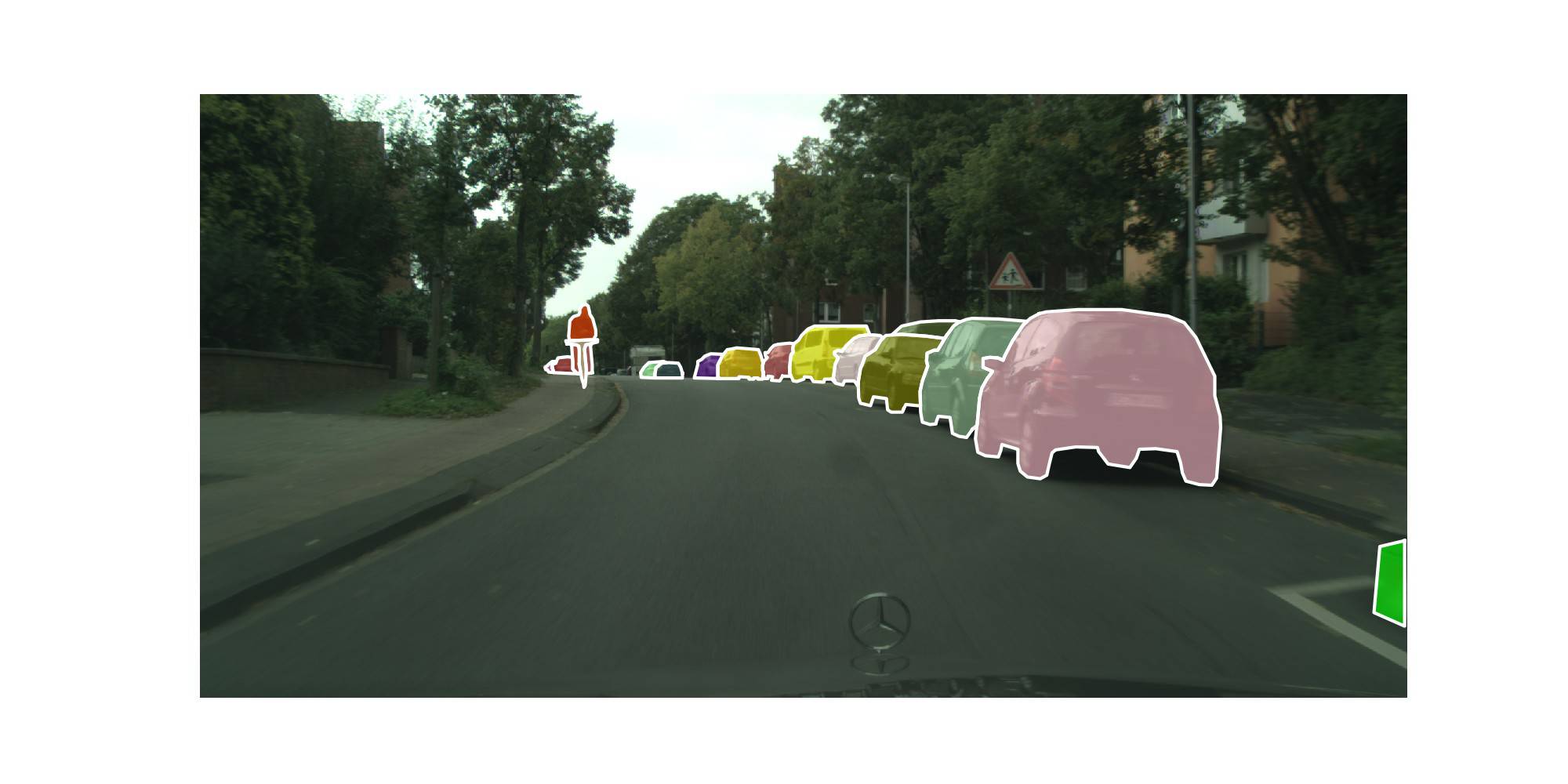}\\[-0.3mm] 
	\bf{0 clicks} & \bf{389 clicks}\\[2mm]

	\includegraphics[width=0.496\linewidth,trim=180 200 150 190,clip]{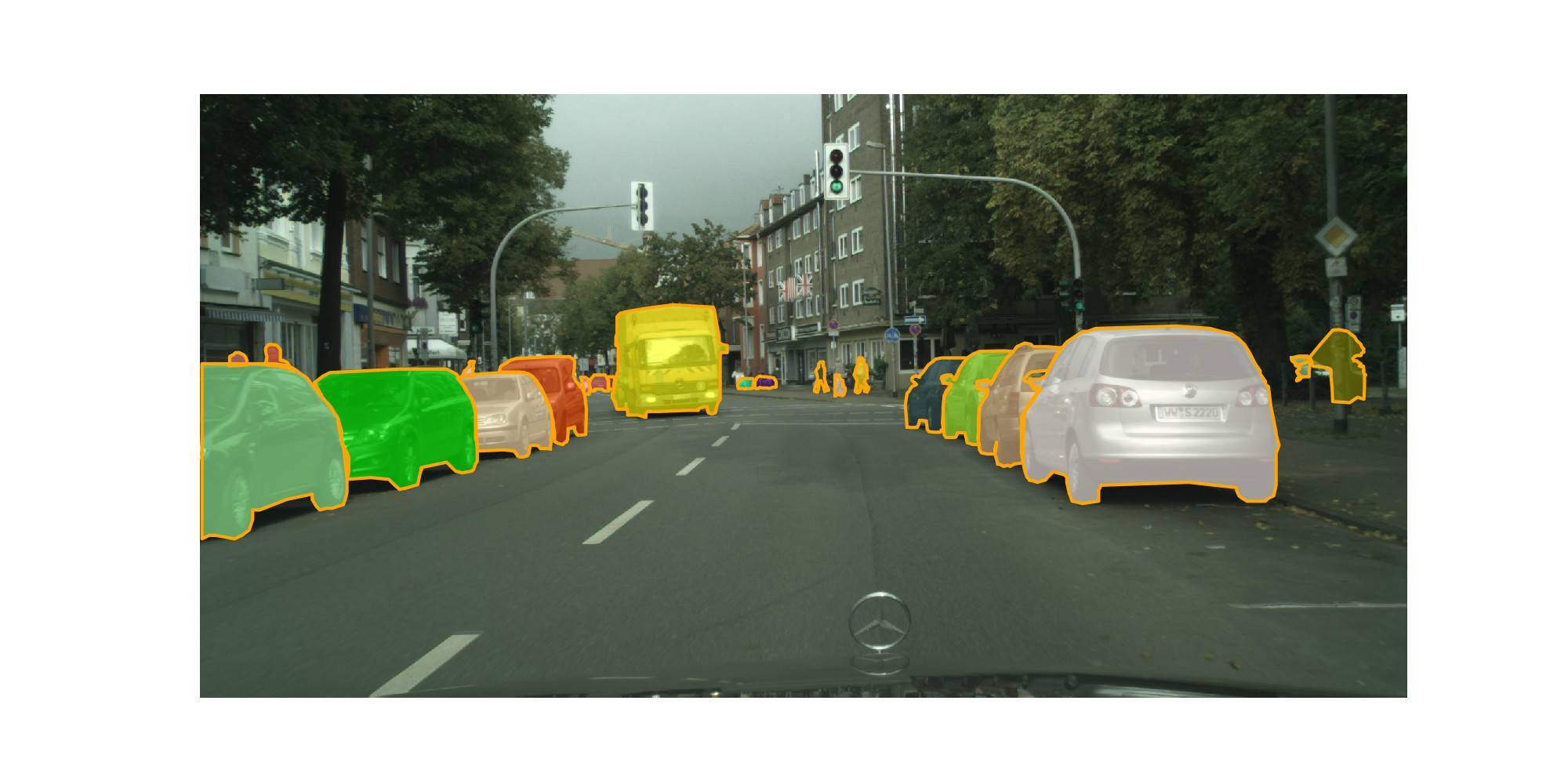} & \includegraphics[width=0.496\linewidth,trim=180 200 150 190,clip]{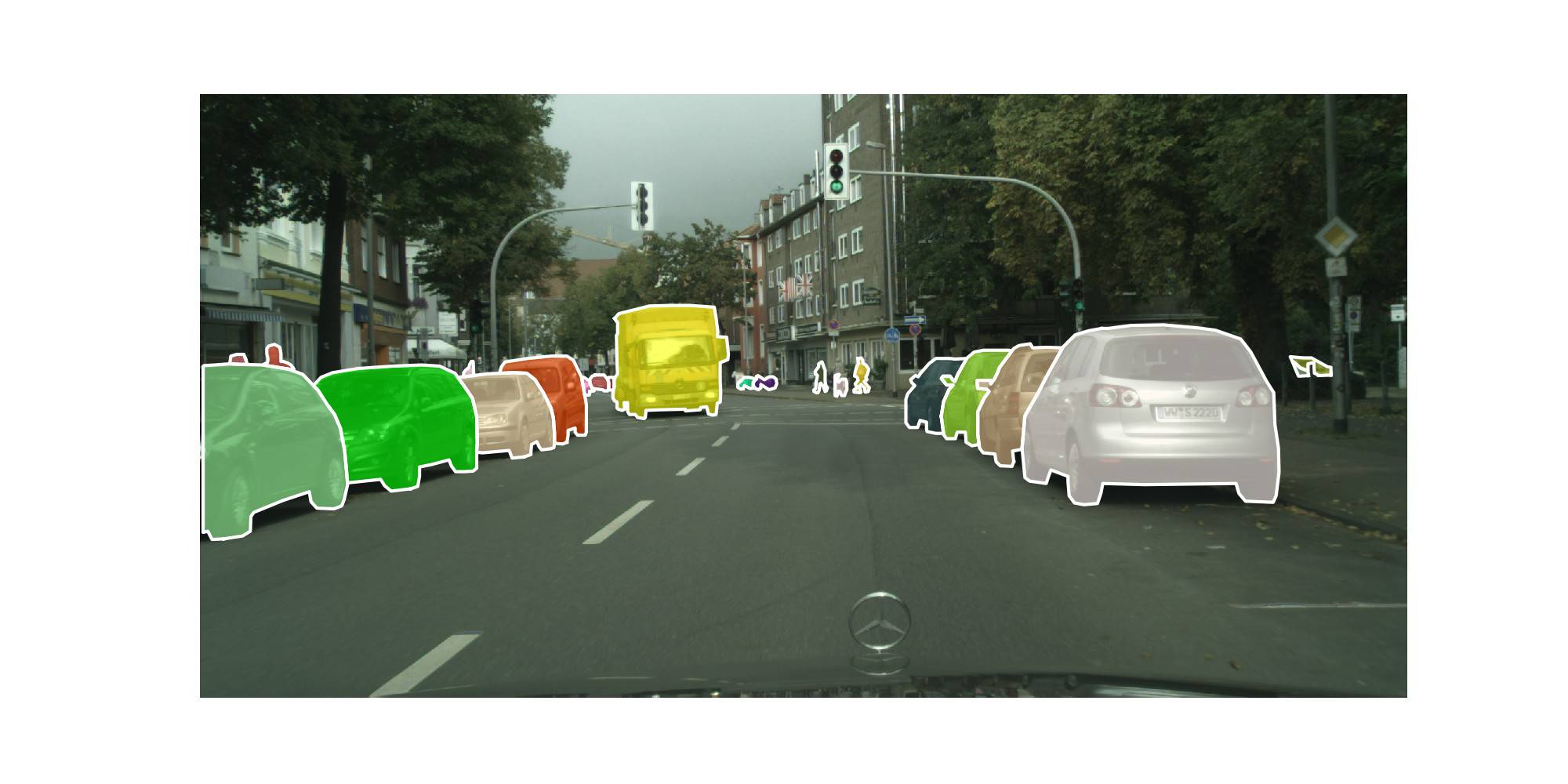}\\[-0.3mm] 
	\bf{0 clicks} & \bf{539 clicks}\\[2mm]

	\includegraphics[width=0.496\linewidth,trim=180 200 150 180,clip]{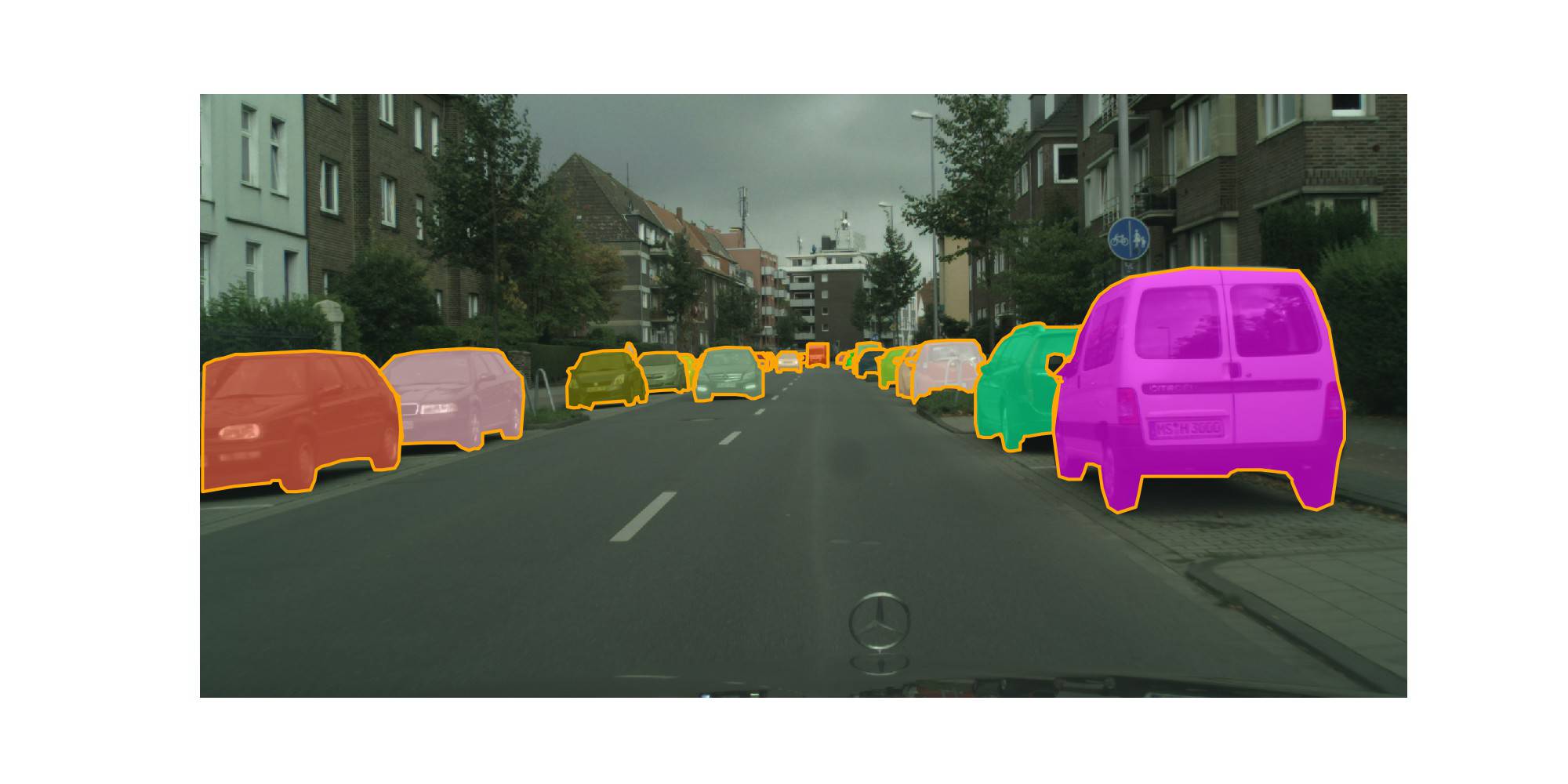} & \includegraphics[width=0.496\linewidth,trim=180 200 150 180,clip]{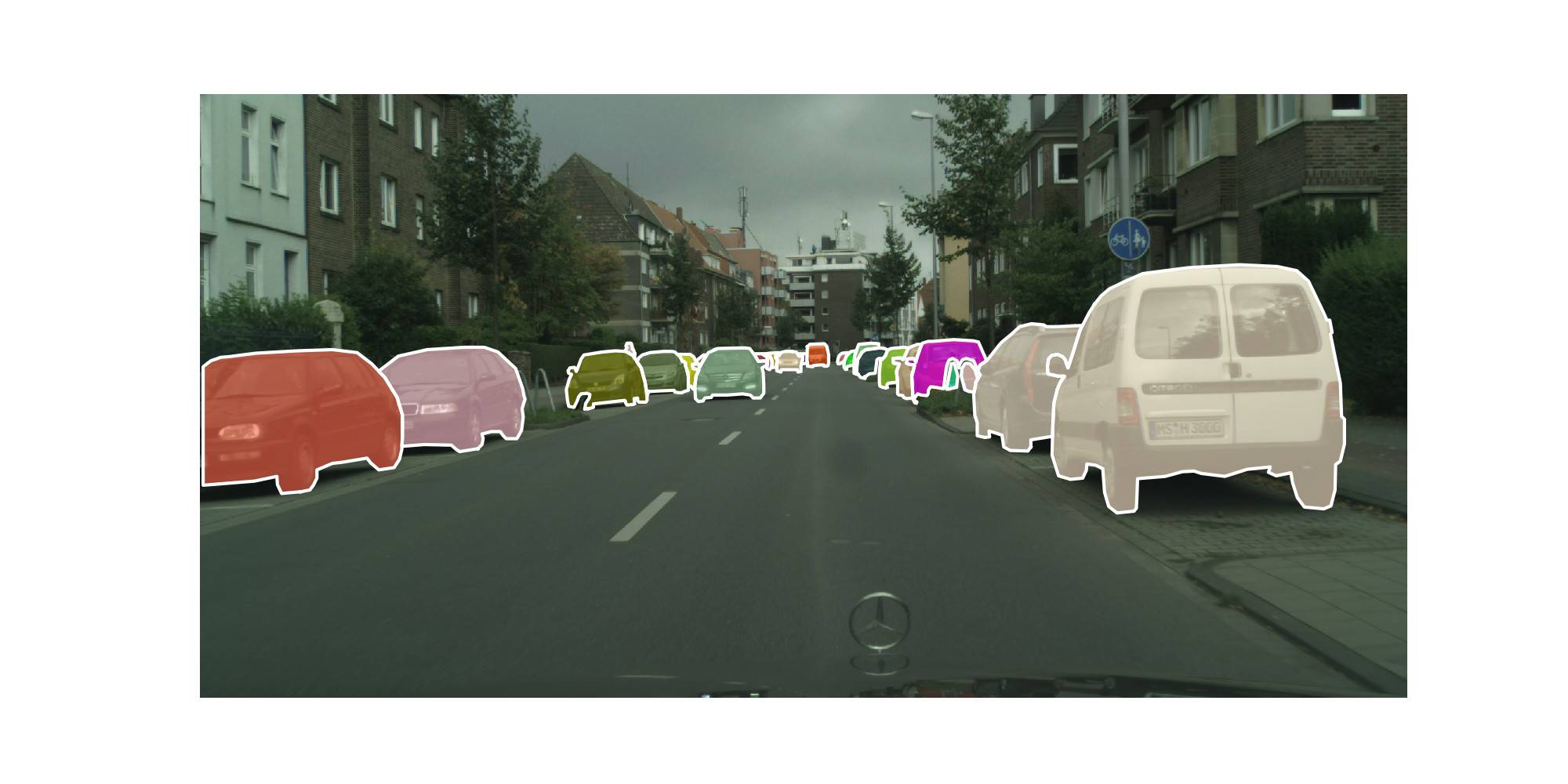} \\[-0.3mm] 
	\bf{0 clicks} & \bf{636 clicks}\\[2mm]

	\includegraphics[width=0.496\linewidth,trim=180 200 150 180,clip]{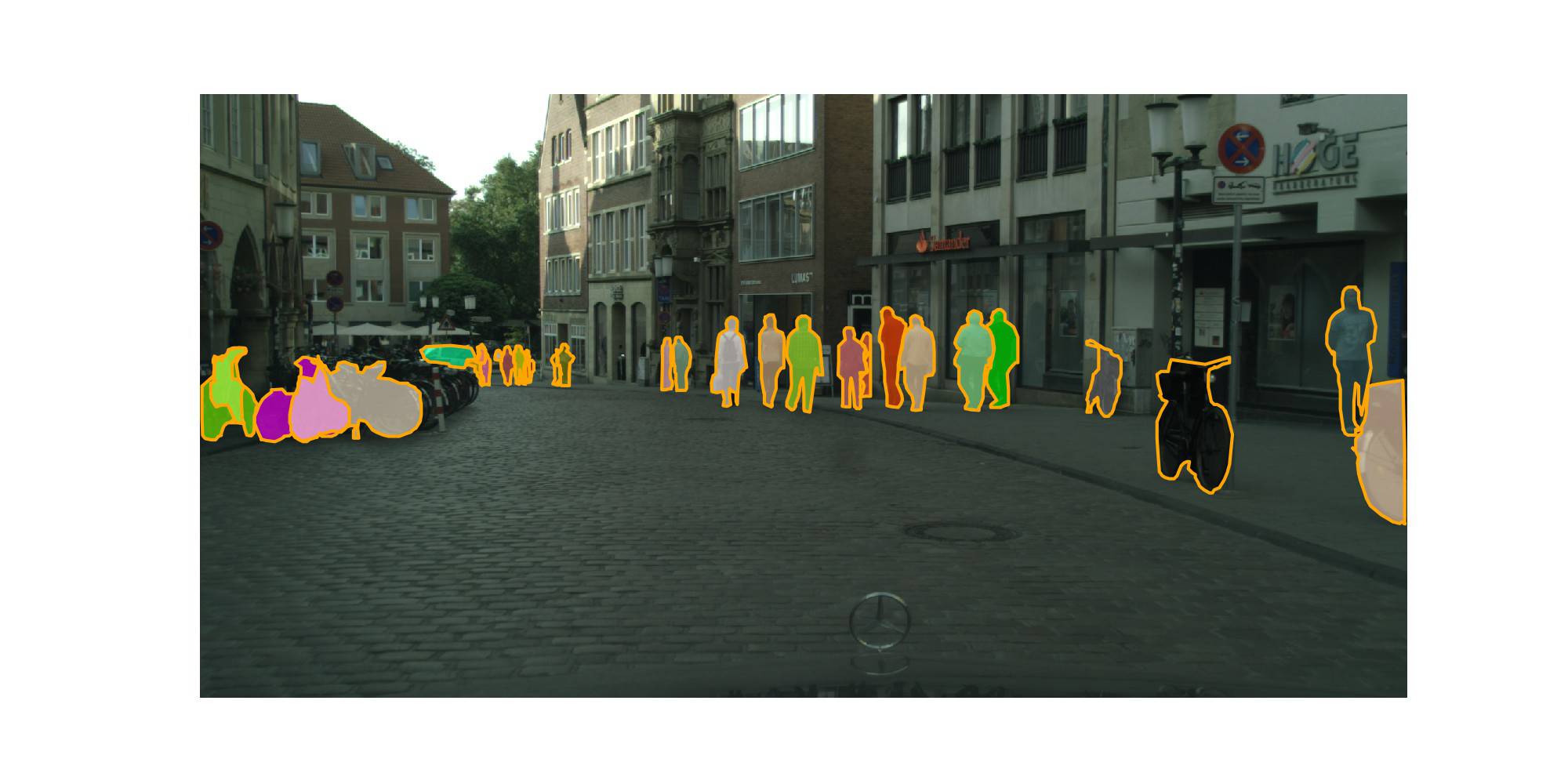} & \includegraphics[width=0.496\linewidth,trim=180 200 150 180,clip]{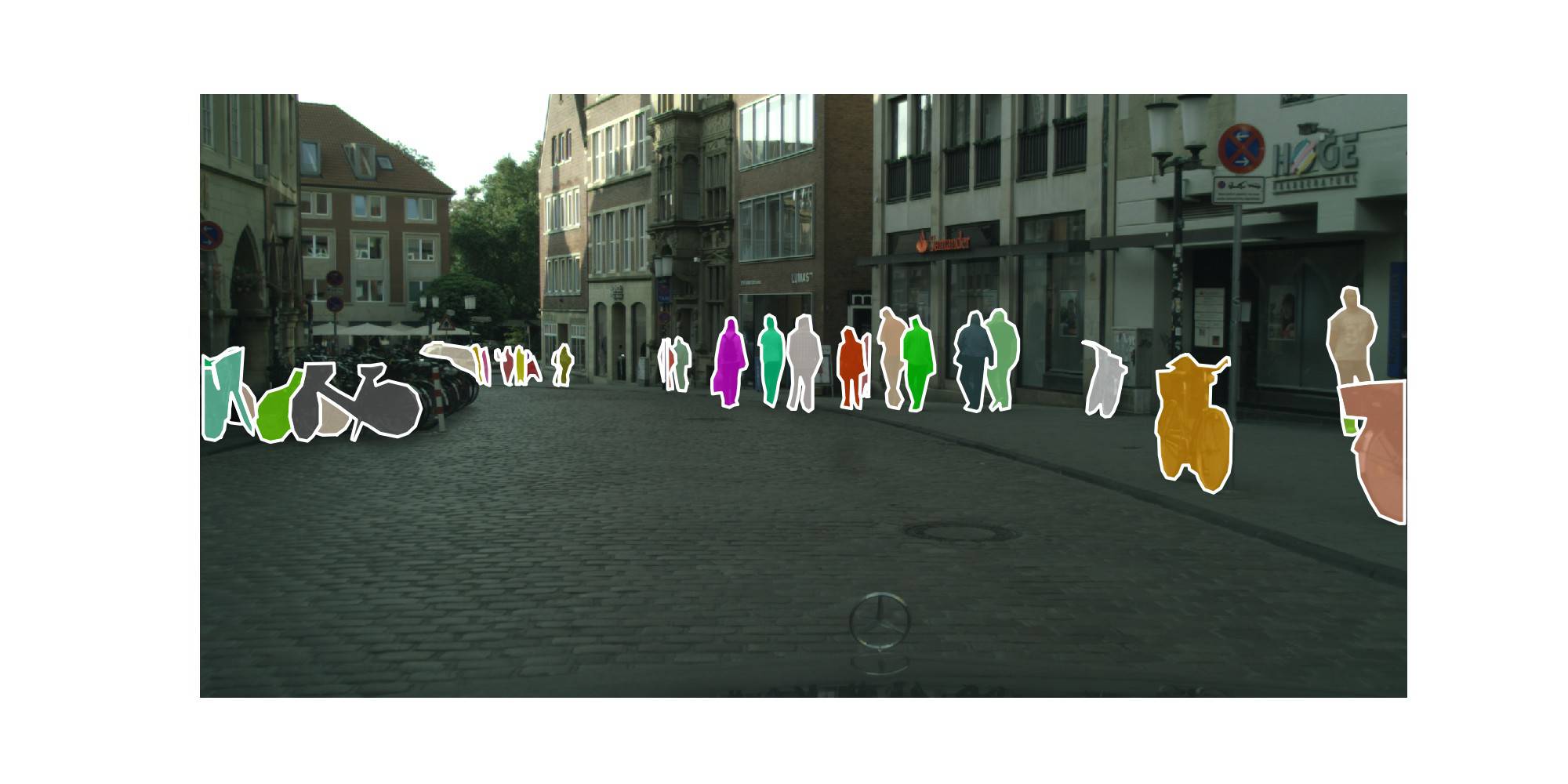}\\[-0.3mm] 
	\bf{0 clicks} & \bf{716 clicks}\\[2mm]

	\includegraphics[width=0.496\linewidth,trim=180 200 150 180,clip]{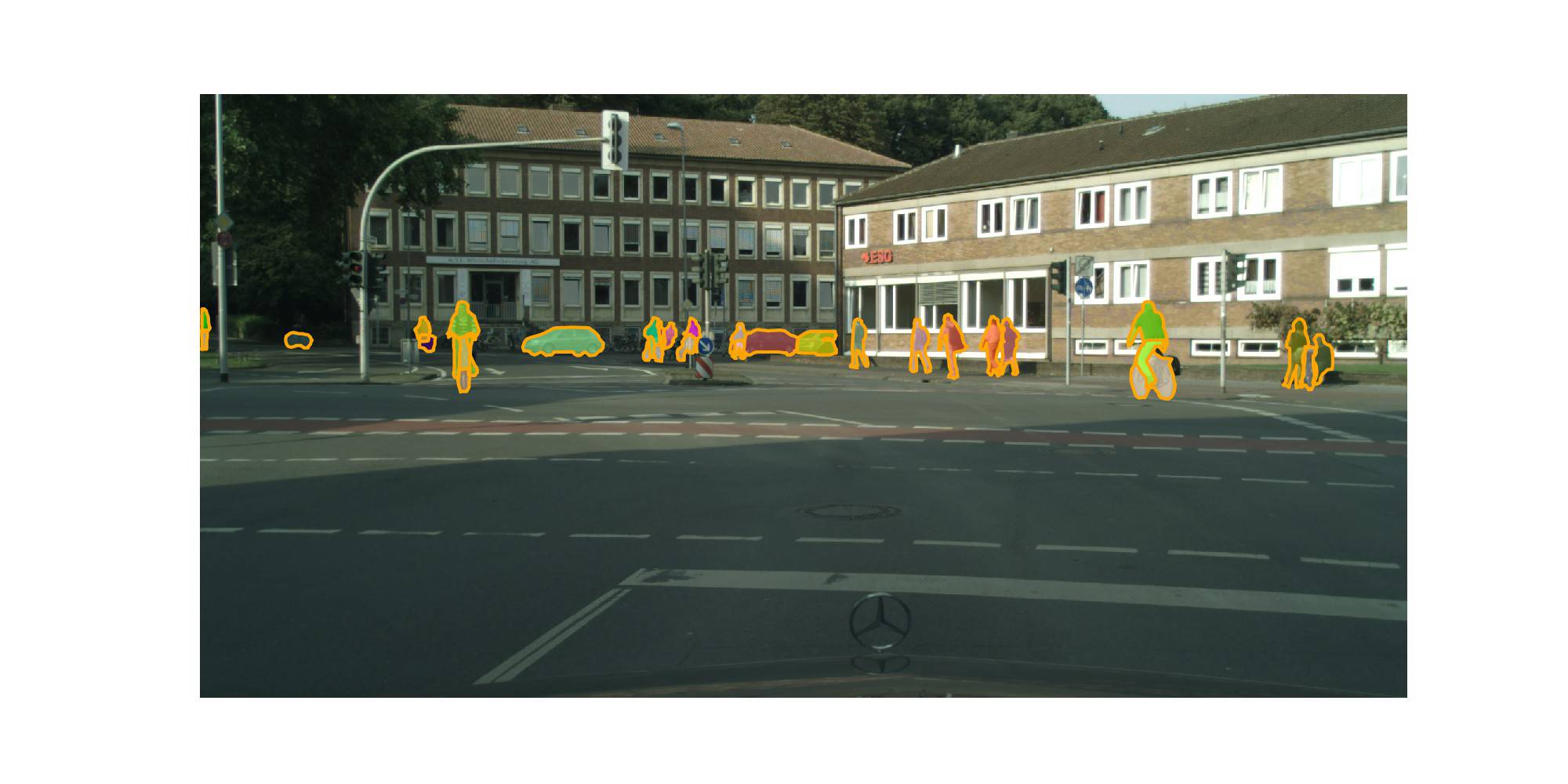} & \includegraphics[width=0.496\linewidth,trim=180 200 150 180,clip]{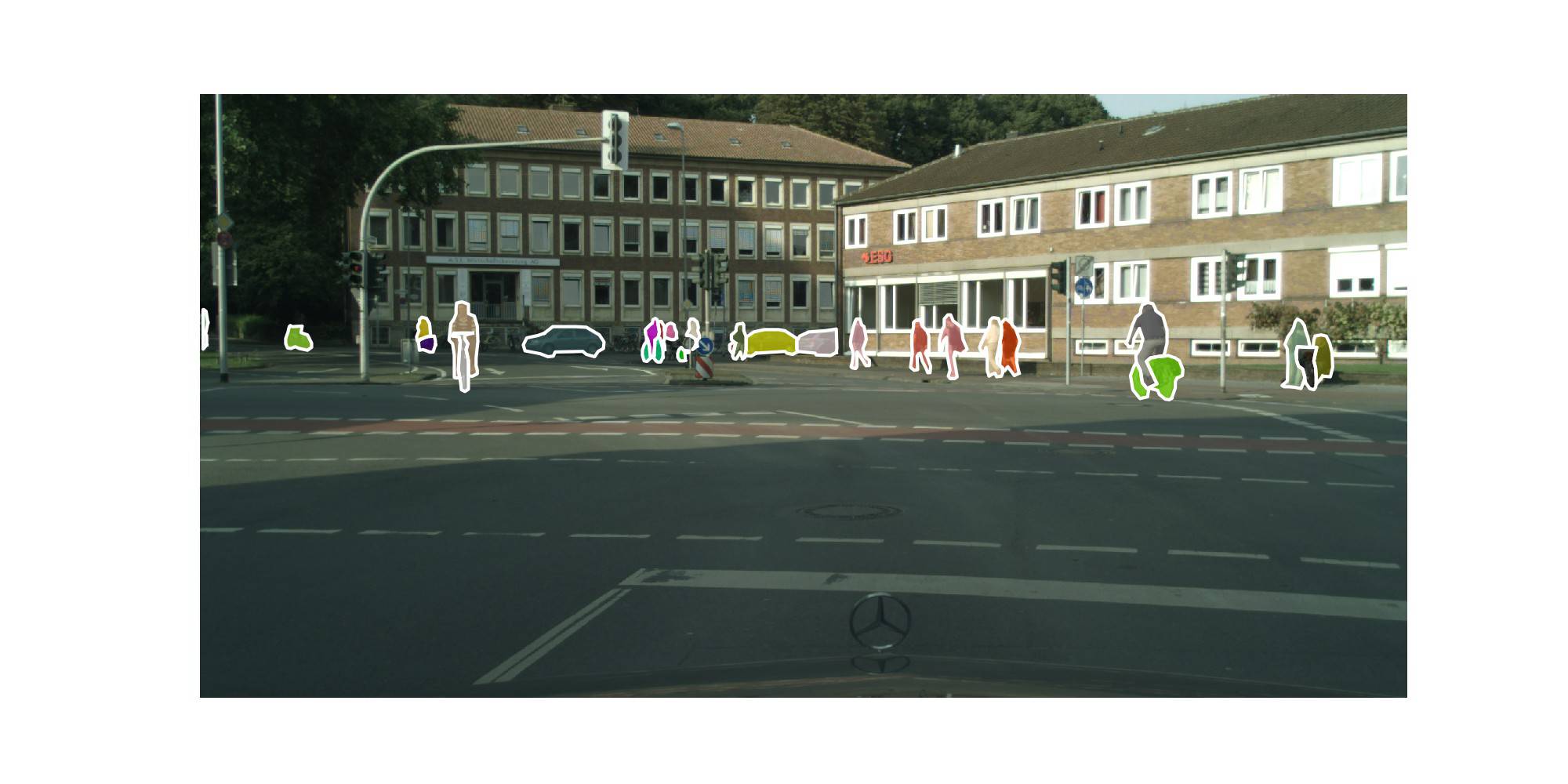}\\
	\bf{0 clicks} & \bf{573 clicks}\\

\end{tabular}
\caption{{\bf Automatic mode on Cityscapes dataset}: Qualitative comparison between a human annotator vs PolygonRNN++ in automatic mode on Cityscapes. This model exploits GGNN to output a polygon at a higher resolution. Note that our model relies on ground-truth bounding boxes.}
\label{fig:human_auto_cityscapes2}
\end{figure*}

\paragraph{Automatic Mode in Cityscapes.} In Figure~\ref{fig:human_auto_cityscapes1} and \ref{fig:human_auto_cityscapes2} we provide a qualitative comparison between our model in automatic mode and the ground-truth polygons.
The first column illustrates the predicted full image while the second shows the GT polygons. 
We remind the reader that here, our model exploits ground-truth bounding boxes.

\vspace{-3mm}

\paragraph{Full-image Instance-Level Segmentation on Cityscapes.}
We evaluate our model on the task of instance segmentation. In our scenario, this can also be seen as an automatic full image annotation task.
Since PolygonRNN++ requires bounding boxes, we use FasterRCNN~\cite{fasterrcnn} for object detection on the whole image.
In particular, we train the best FasterRCNN model of~\cite{Speed_Accuracy} (pre-trained on MS-COCO) on the Cityscapes dataset (fine annotations only). 
The predicted boxes are then fed to our model to produce polygonal instance segmentations. Evaluating Polygon-RNN++ with FasterRCNN on the {\bf Cityscapes test set} achieves $22.8 \%$  AP and $42.6\%$ $AP_{50}$. 

A major limitation of our model in this scenario is that it only predicts one polygon per bounding-box.
As a result, multi-component object-masks, coming from occluding objects, are heavily penalized by the evaluation procedure.

In order to tackle this issue, we further use semantic information. Specifically, we use the predicted semantic segmentation results from~\cite{pspnet}. We then perform a logical ``and" operation between the  predicted class-semantic map and our instance polygonal prediction. 
Following this scheme, we achieve $25.49\%$  AP and $45.47\%$ $AP_{50}$ on the {\bf test set}. 

\begin{figure*}[ht]
	\begin{tabular}{c c}
	\includegraphics[width=0.496\linewidth,trim=180 200 150 190,clip]{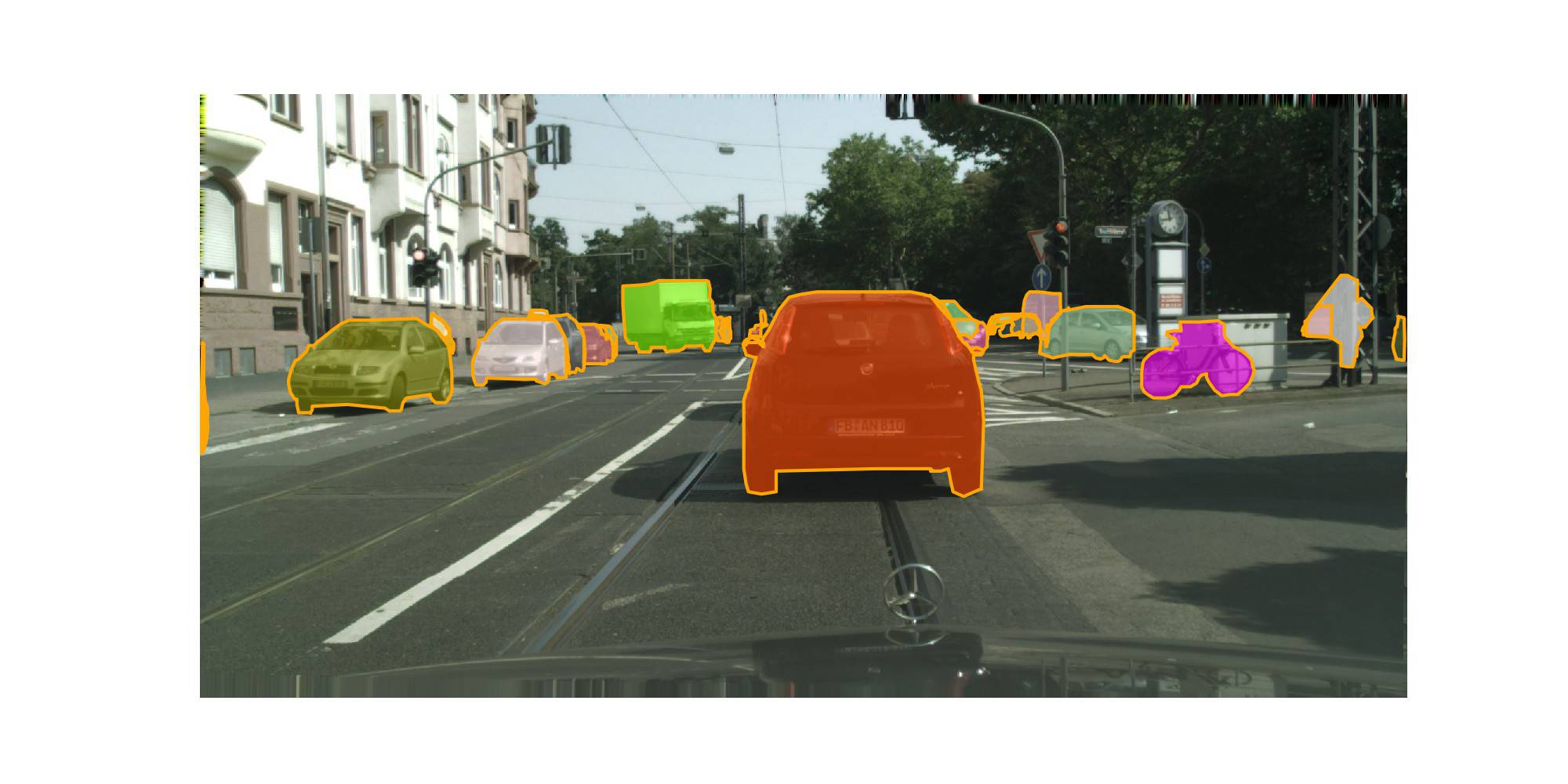}& %
	\includegraphics[width=0.496\linewidth,trim=180 200 150 190,clip]{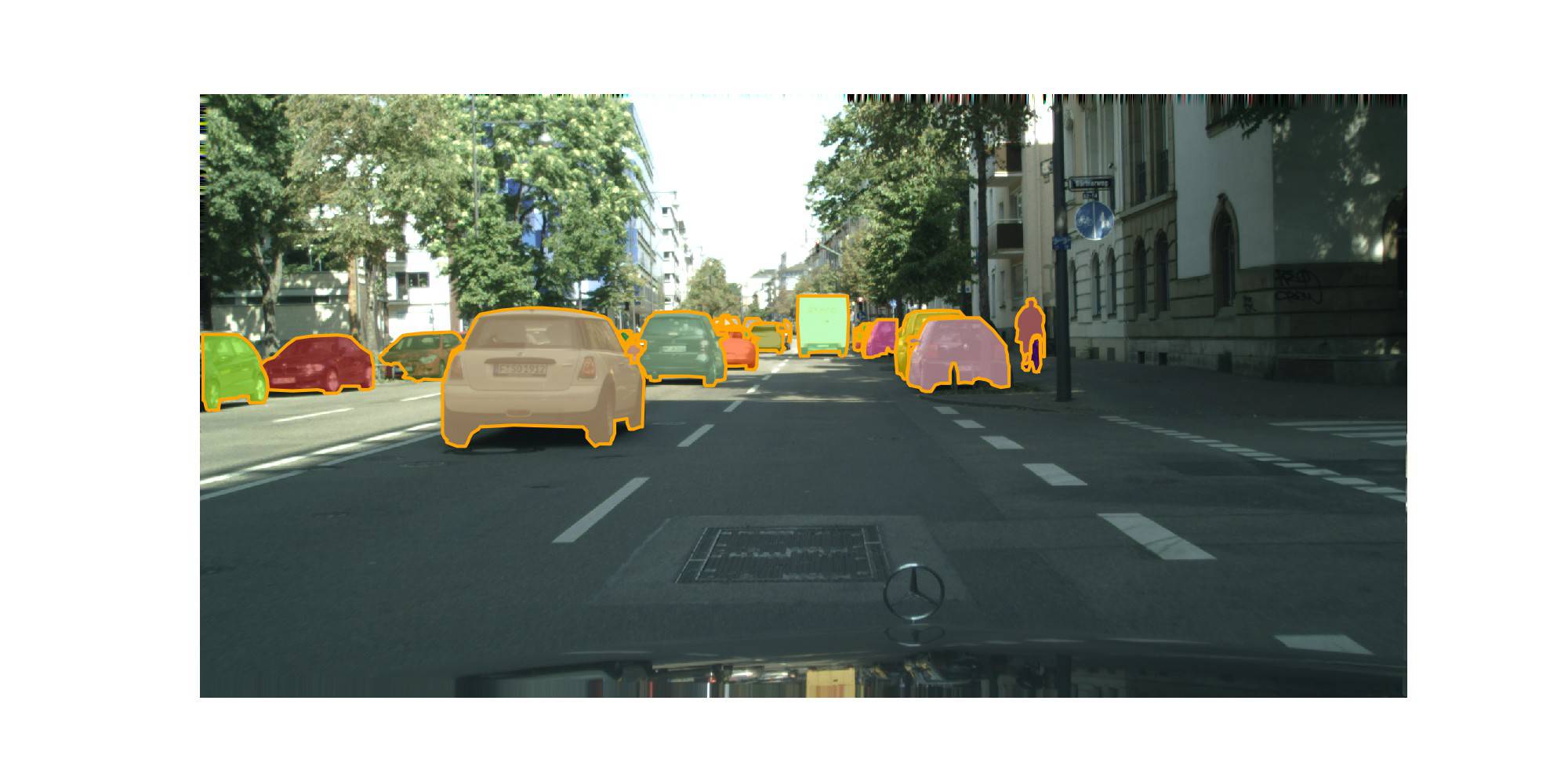}\\ 
	\includegraphics[width=0.496\linewidth,trim=180 200 150 190,clip]{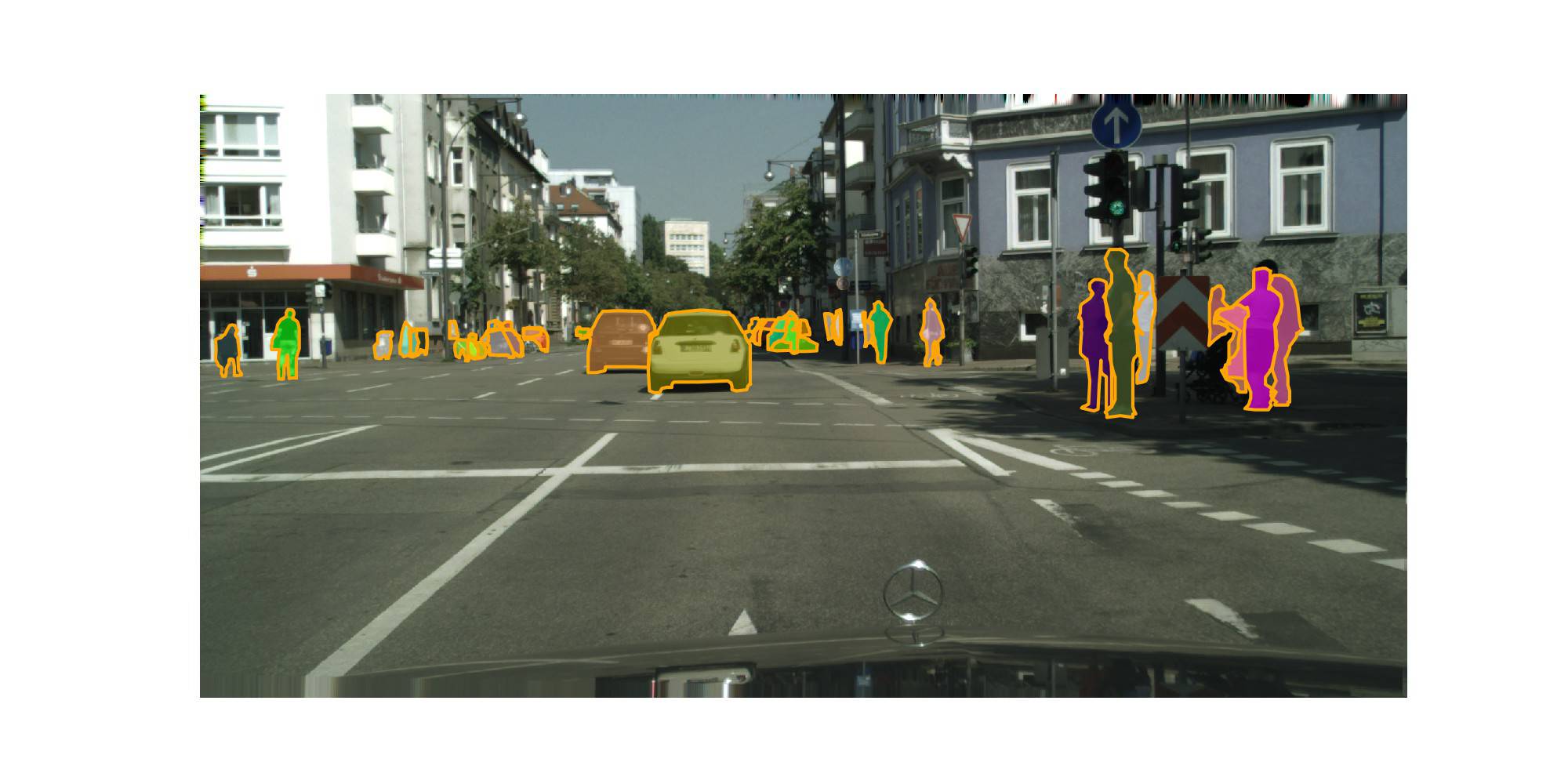} & 
	\includegraphics[width=0.496\linewidth,trim=180 200 150 180,clip]{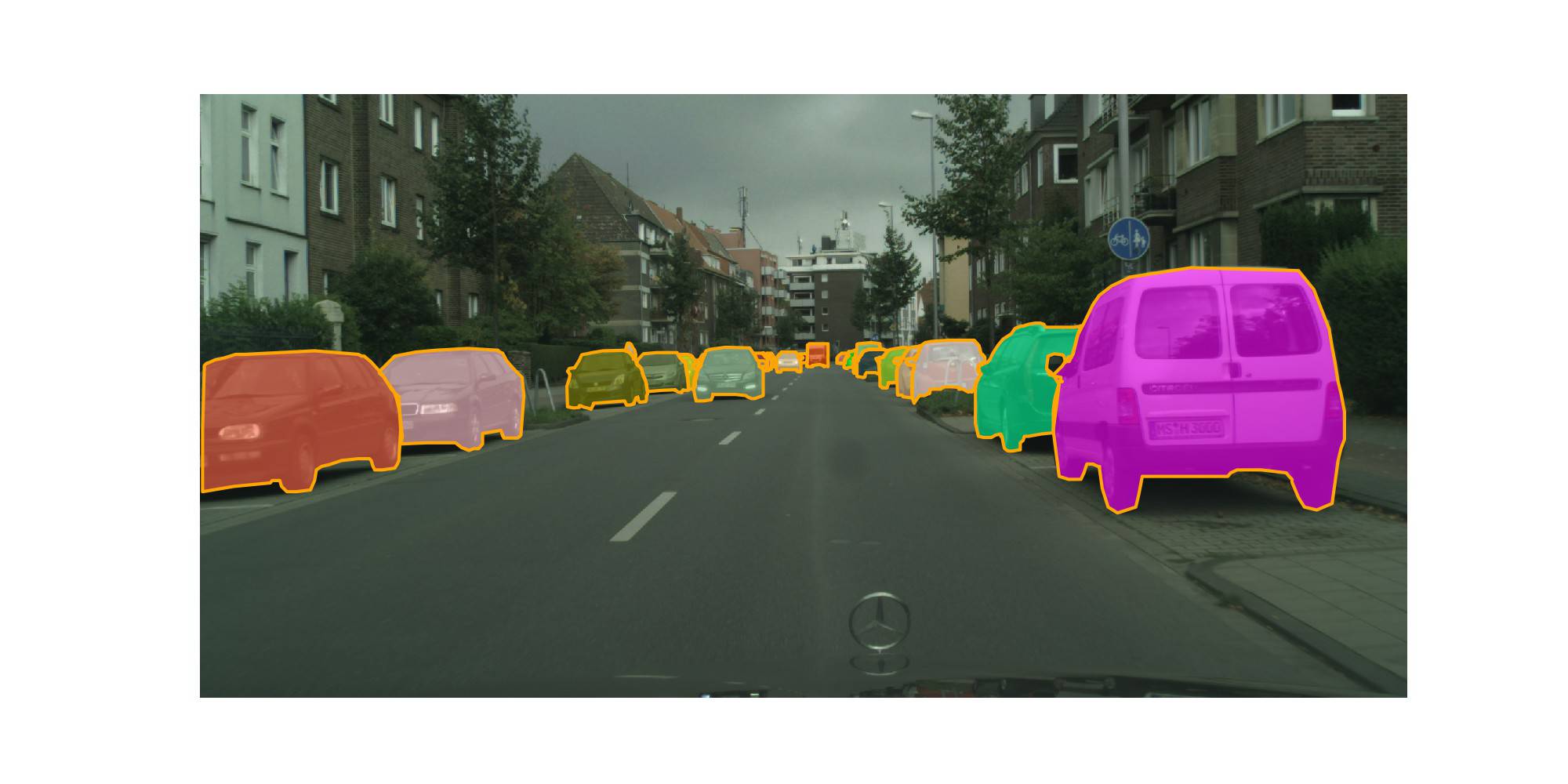}\\ 
	\includegraphics[width=0.496\linewidth,trim=180 200 150 180,clip]{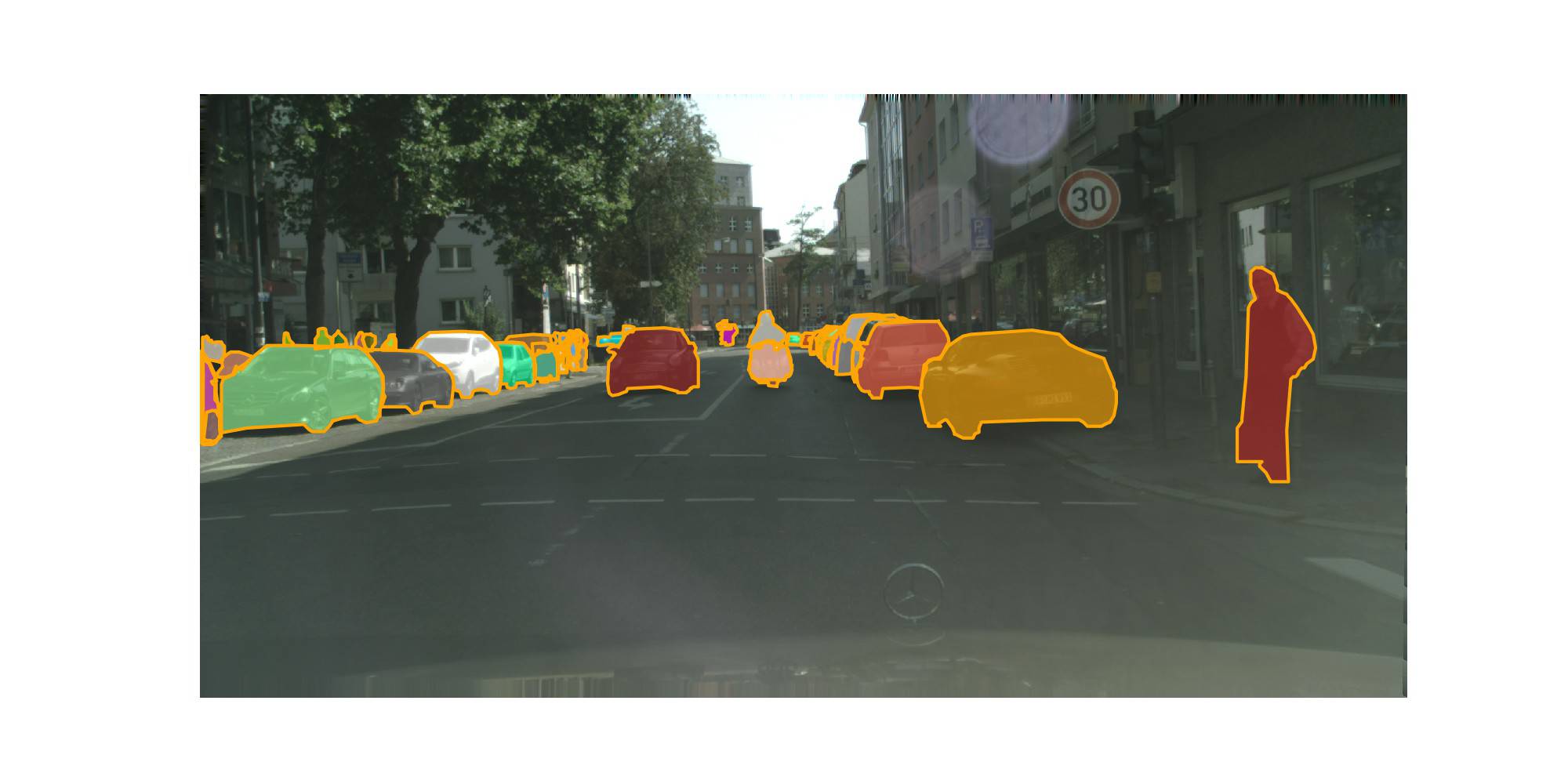} & 
	\includegraphics[width=0.496\linewidth,trim=180 200 150 180,clip]{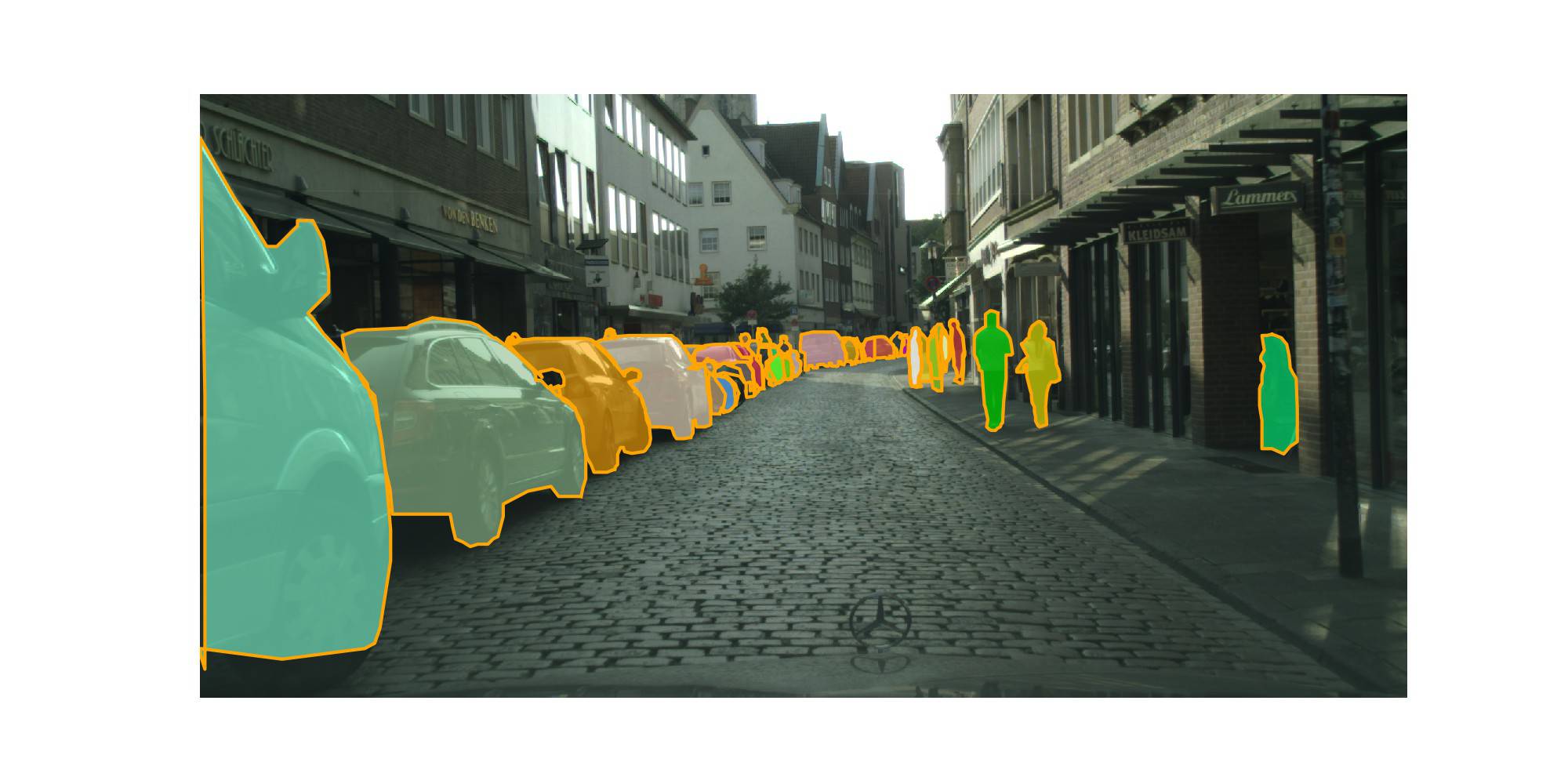}\\ 
	\includegraphics[width=0.496\linewidth,trim=180 200 150 180,clip]{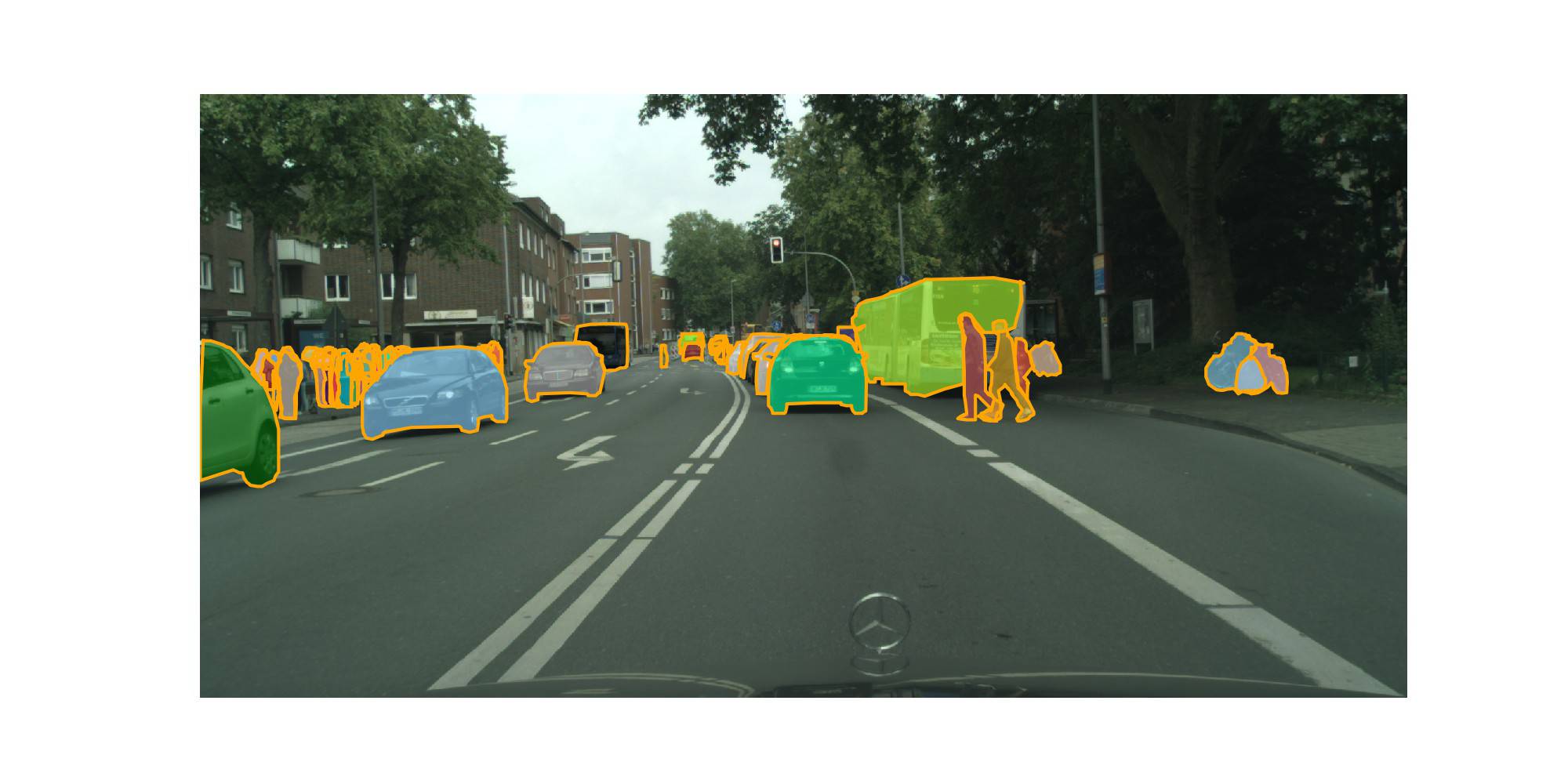} & \includegraphics[width=0.496\linewidth,trim=180 200 150 180,clip]{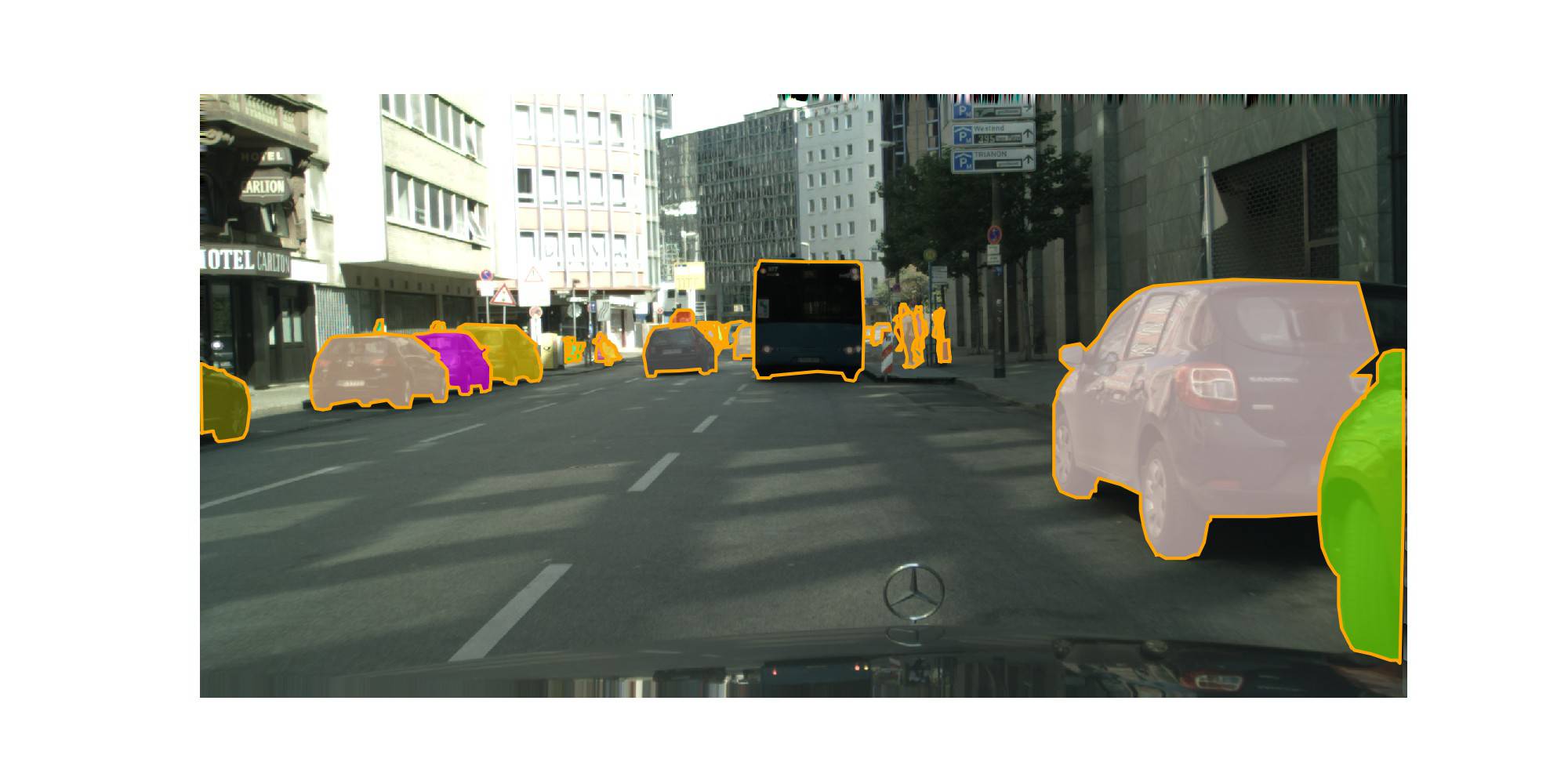}
\end{tabular}
\caption{{\bf Full-image instance segmentation}: Qualitative results of full image prediction using Polygon-RNN++ with boxes from Faster-RCNN}
\label{fig:human_rcnn_cityscapes}
\end{figure*}

In Figure~\ref{fig:human_rcnn_cityscapes}, we show more qualitative results of our full instance segmentation model (i.e. with boxes from Faster-RCNN).
Results on the Cityscapes Instance Segmentation Benchmark are reported in Table~\ref{tbl_performance_cityscapes_test}.

\vspace{-3mm}
\paragraph{Interactive mode in Cityscapes.} Figure~\ref{fig:hist} shows the histogram of the number of corrections (clicks) for different values of $T_2$ and $T$. It also shows the average IoU and the average number of clicks for the given thresholds. We can see that most of the predictions can be successfully corrected with 5 clicks.
Figure~\ref{fig:human_interactive_cityscapes1} shows a few qualitative examples of our interactive simulation on the Cityscapes dataset. The automatically predicted polygons are shown in the first column while the second column shows the result after a certain number of corrections. The last one depicts the ground-truth polygons.
For all instances, we show the required number of clicks and the achieved IoU.
\vspace{-3mm}
\paragraph{Automatic Mode in Out-of-Domain Imagery.} In Figures ~\ref{fig:human_auto_ade20k}, \ref{fig:human_auto_aerial}, \ref{fig:human_auto_kitti}, \ref{fig:human_auto_medical1}, \ref{fig:human_auto_medical2} we analyze the qualitative performance of our model on different datasets that capture both shifts in environment KITTI \cite{kitti} and domain (general scenes~\cite{ade20k}, aerial~\cite{Sun14}, medical~\cite{medical1,medical1b,medical2}).
We emphasize that here we use our model trained on Cityscapes \textbf{without any fine-tuning} on these datasets. 
The first column shows prediction in automatic mode while the second column visualizes the GT instances.
Note that in some of these datasets we do not have GT polygons as  only segmentation masks are provided.
In those cases, the GT image is labeled as \emph{Mask} and the number of clicks is not shown.
\vspace{-3mm}
\paragraph{Automatic Mode and Online Fine-Tuning.} Figure~\ref{fig:online_finet} illustrates the performance of the proposed Online Fine-tuning algorithm on different datasets.
We first show the prediction of the model without any fine-tuning. We then illustrate the automatic predictions of the fine-tuned model. The GT instances are shown in the last column.
In all cases, the fine-tuned predictions are generated after the last chunk has been seen (illustrated in Figure 9 of the main paper).

\begin{figure*}[ht]
\addtolength{\tabcolsep}{-2pt}
	\begin{tabular}{c c}
	\hspace{-2mm}\includegraphics[height=0.4\textheight,width=0.51\linewidth,trim=0 0 0 0,clip]{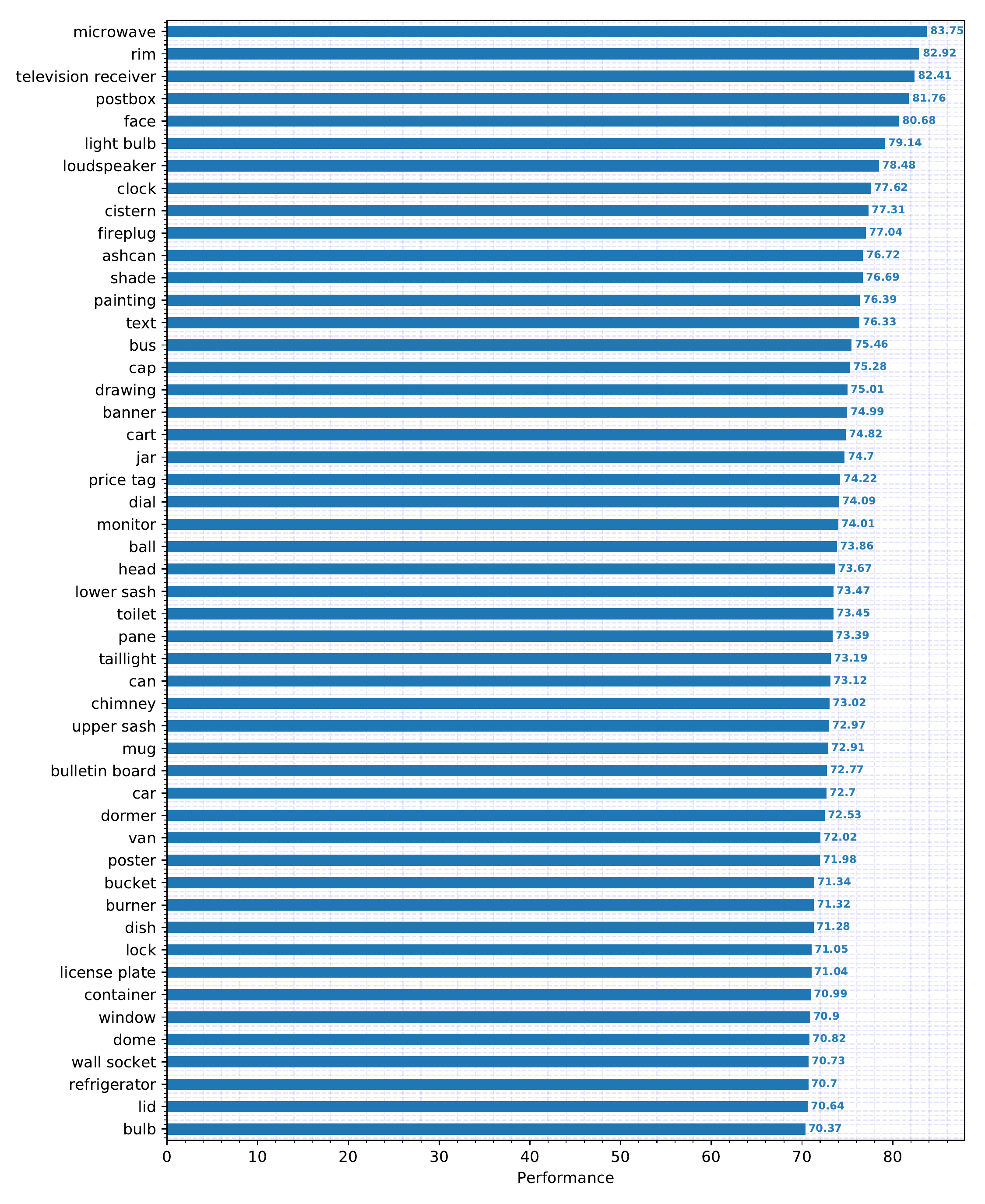} & \includegraphics[height=0.4\textheight,width=0.51\linewidth,trim=0 0 0 0,clip]{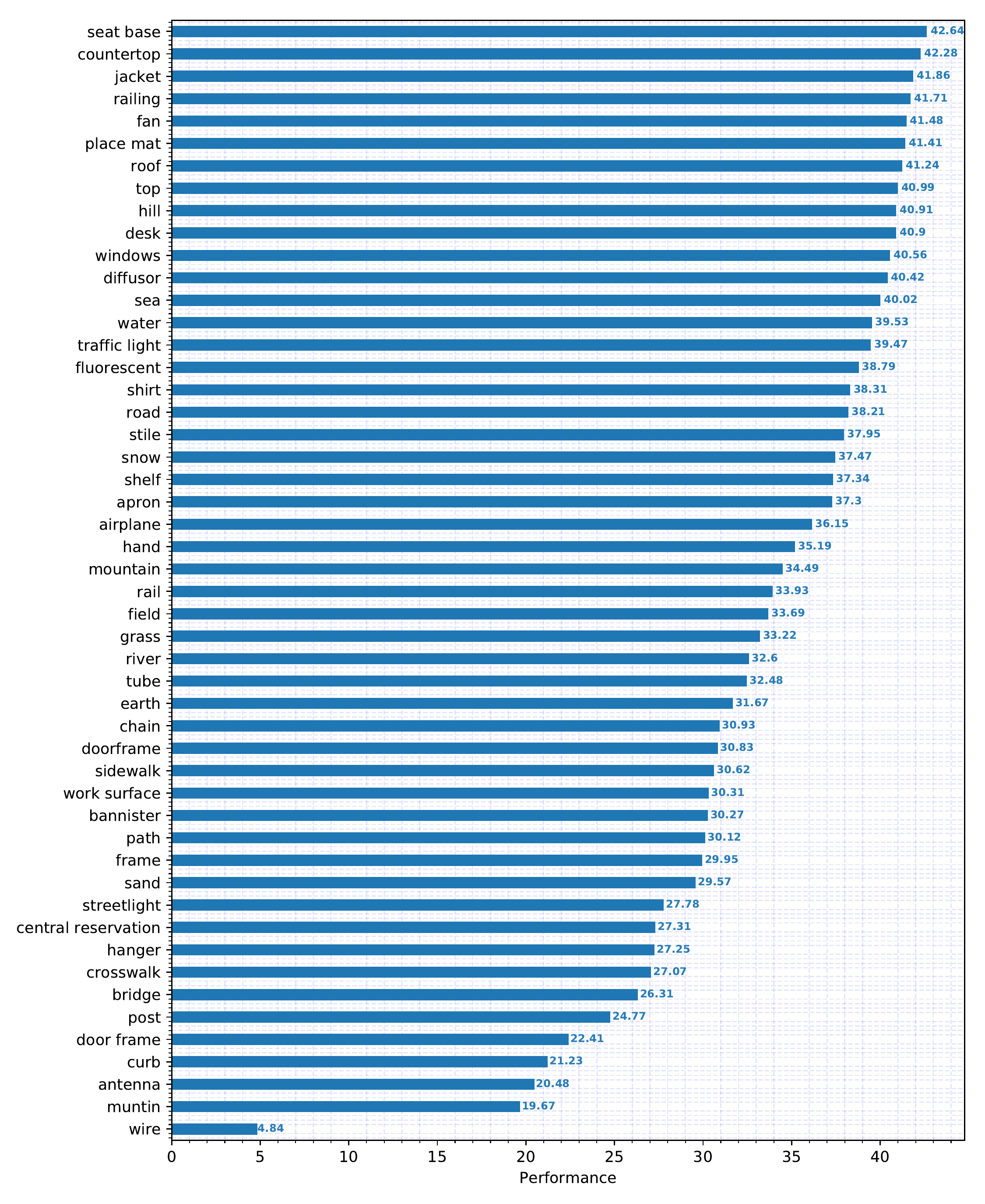}\\
	
\end{tabular}
\caption{{\bf Automatic mode on ADE val}: Performance of PolygonRNN++ {\bf without fine-tuning} on ADE validation set. Only categories with more than 20 instances are shown.    \textbf{Left:} Top 50 categories in terms of IoU performance.  \textbf{Right:} Bottom 50 categories in terms of IoU performance.}
\label{fig:full_ade_plots}
\end{figure*}

\begin{figure*}[ht]
	\includegraphics[width=0.496\linewidth,trim=20 30 20 0,clip]{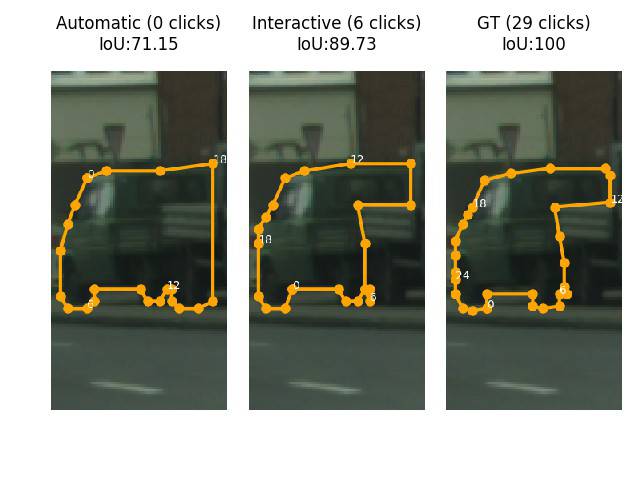} \hspace{2mm} 
	\includegraphics[width=0.496\linewidth,trim=20 30 20 0,clip]{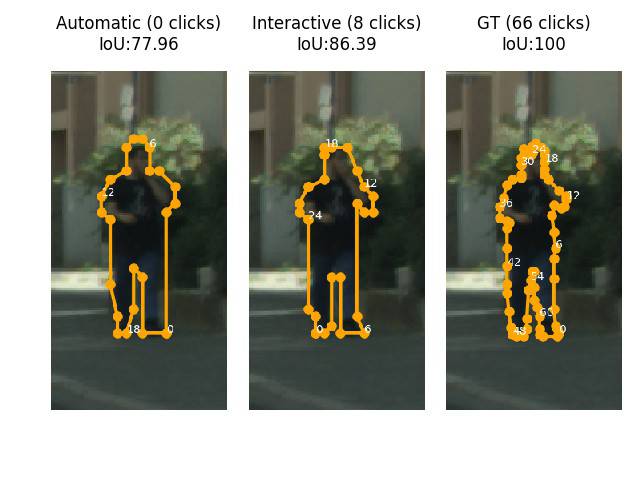}\\
	\includegraphics[width=0.496\linewidth,trim=20 37 20 12,clip]{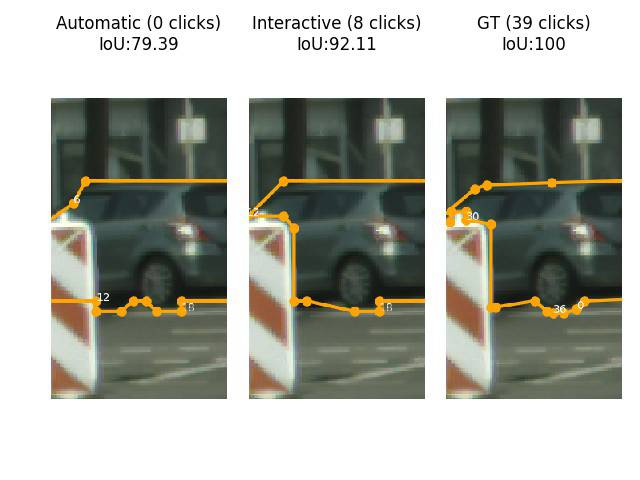} \hspace{2mm}    \includegraphics[width=0.496\linewidth,trim=20 30 20 0,clip]{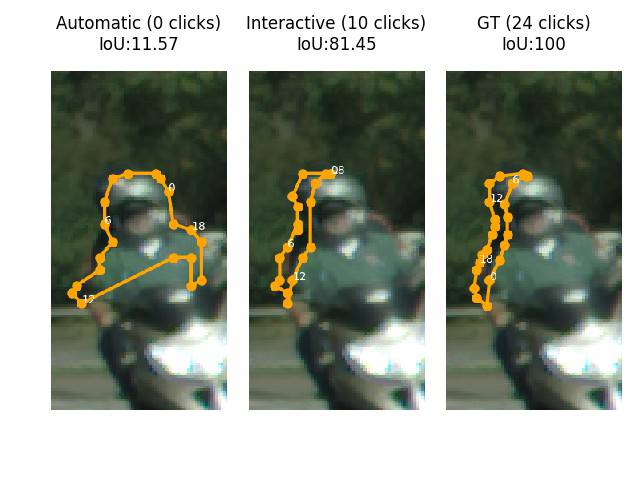}\\
	\includegraphics[width=0.496\linewidth,trim=20 30 20 0,clip]{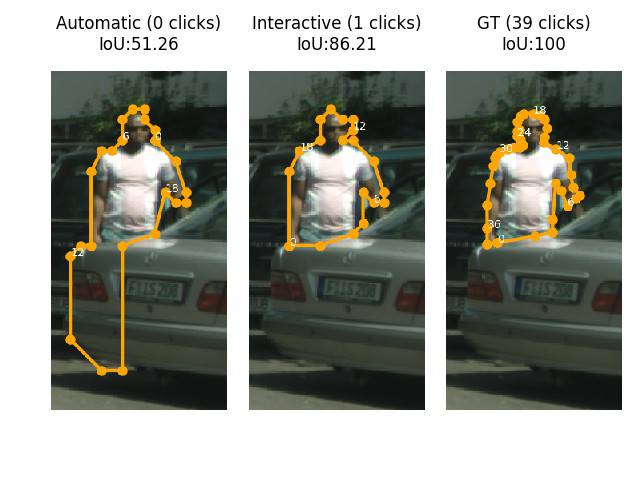} \hspace{2mm} 
	\includegraphics[width=0.496\linewidth,trim=20 30 20 0,clip]{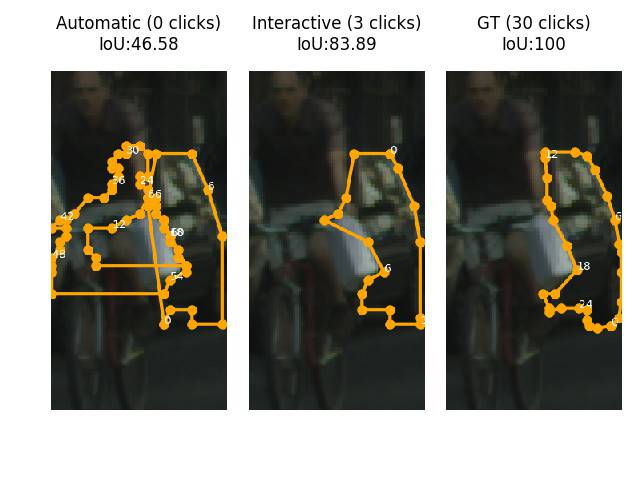}\\
\caption{{\bf Interactive mode on the Cityscapes dataset}. Here we show only predicted and corrected vertices at the $28\times 28$ resolution (no GGNN is used here). Notice a major failure case in automatic mode in the bottom right example (where self-intersection occurs), which however gets quickly corrected. Note that the annotator is simulated (we correct a vertex if it deviates from the ground-truth vertex by a threshold $T$).}
\label{fig:human_interactive_cityscapes1}
\end{figure*}

\begin{figure*}[ht]
	\includegraphics[width=0.496\linewidth,trim=20 30 10 0,clip]{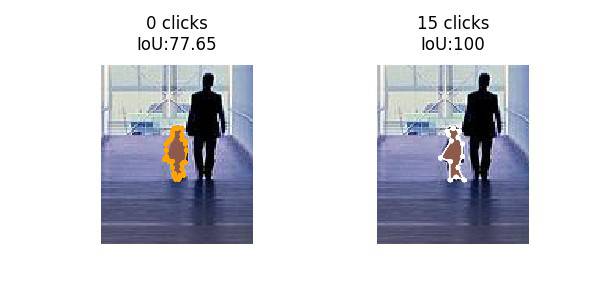} \hspace{2mm}
	\includegraphics[width=0.496\linewidth,trim=20 30 10 0,clip]{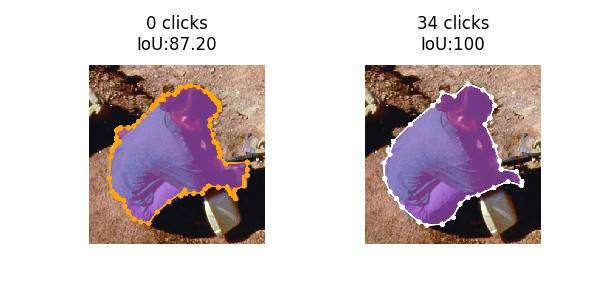}\\[1mm]
	\includegraphics[width=0.496\linewidth,trim=20 30 10 0,clip]{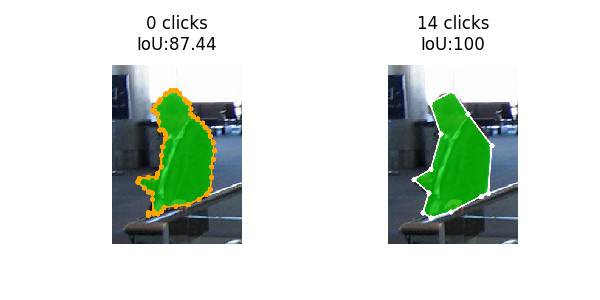} \hspace{2mm} \includegraphics[width=0.496\linewidth,trim=20 30 10 0,clip]{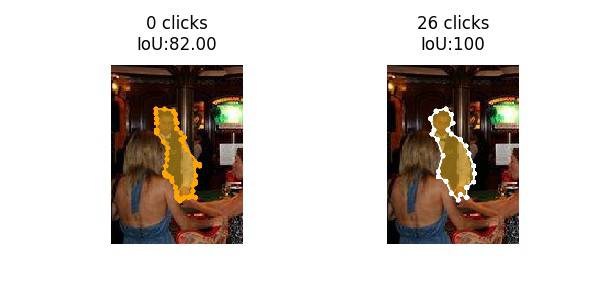}\\[1mm]
	\includegraphics[width=0.496\linewidth,trim=20 30 10 0,clip]{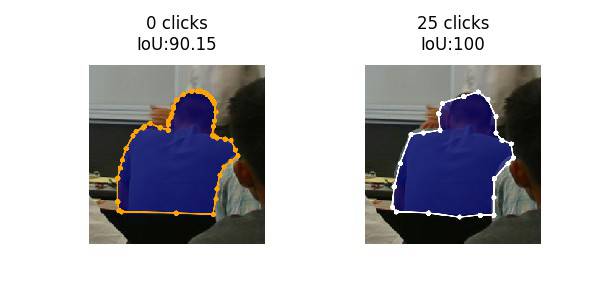} \hspace{2mm} 
	\includegraphics[width=0.496\linewidth,trim=20 30 10 0,clip]{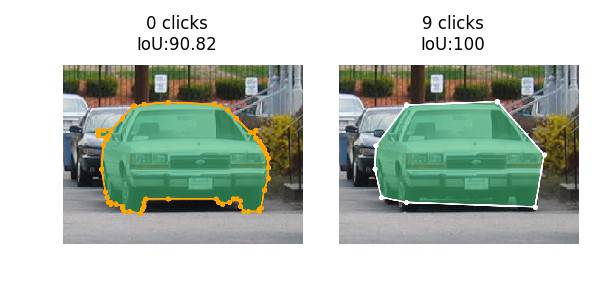}\\[1mm]
	\includegraphics[width=0.496\linewidth,trim=20 30 10 0,clip]{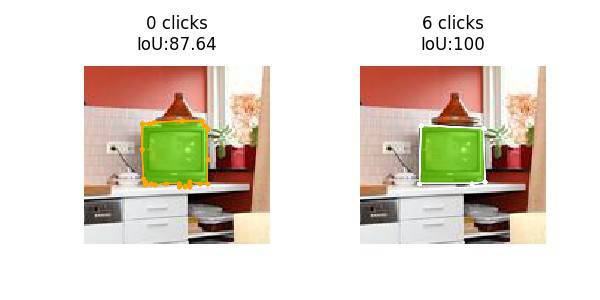} \hspace{2mm} 
	\includegraphics[width=0.496\linewidth,trim=20 30 10 0,clip]{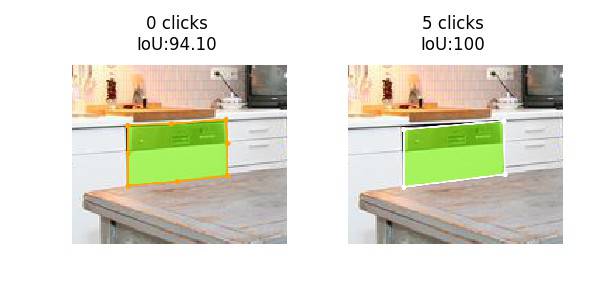}\\[1mm]
	\includegraphics[width=0.496\linewidth,trim=20 30 10 0,clip]{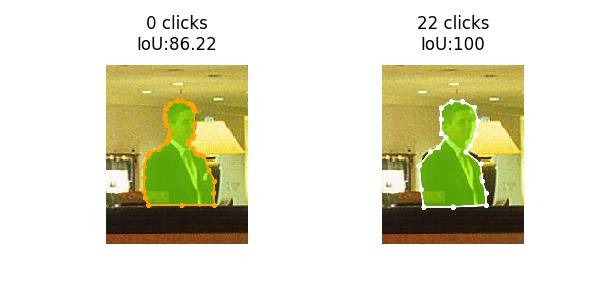} \hspace{2mm} 
	\includegraphics[width=0.496\linewidth,trim=20 30 10 0,clip]{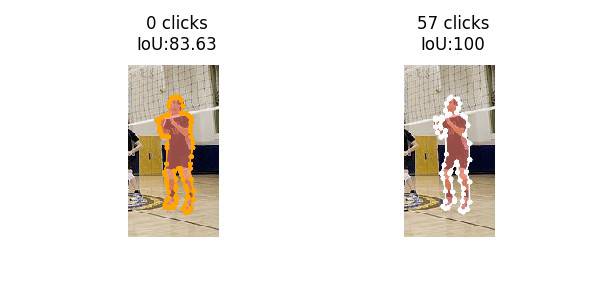}\\[1mm]
\caption{{\bf Cross-domain {\color{magenta}{without}} fine-tuning: trained on Cityscapes $\rightarrow$ tested on ADE20k, automatic mode}.  Qualitative comparison between a human annotator vs PolygonRNN++ in automatic mode in ADE20K \textbf{without fine-tuning}.  Note that our model relies on bounding boxes. Notice also that in some cases our predictions achieve a much higher level of detail than human provided annotations.}
\label{fig:human_auto_ade20k}
\end{figure*}

\begin{figure*}[ht]
	\includegraphics[width=0.496\linewidth,trim=0 30 0 0,clip]{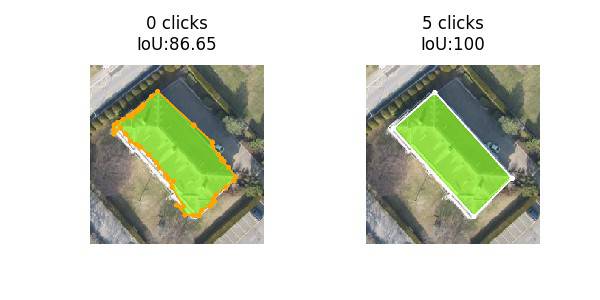} \hspace{2mm}
	\includegraphics[width=0.496\linewidth,trim=0 30 0 0,clip]{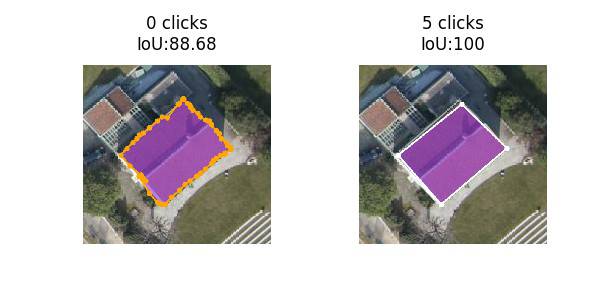}\\
	\includegraphics[width=0.496\linewidth,trim=0 170 0 0,clip]{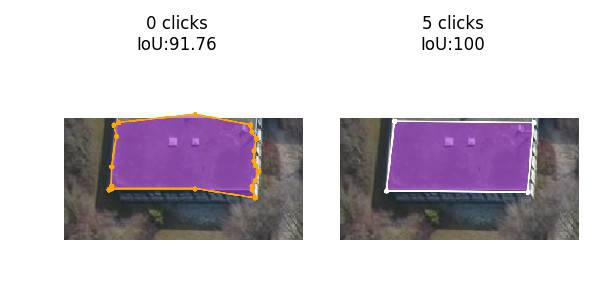} \hspace{2mm} \includegraphics[width=0.496\linewidth,trim=0 170 0 0,clip]{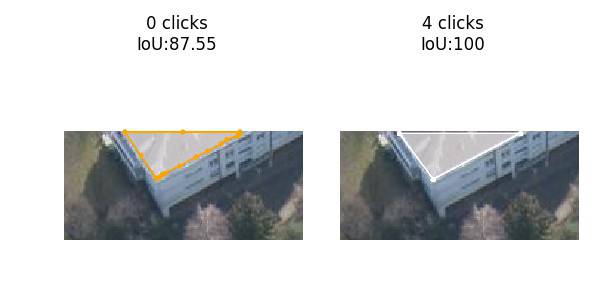}\\
		\includegraphics[width=0.496\linewidth,trim=0 30 0 80,clip]{figures/cross_domain/aerial/IMG_0595_4_stack_iou_0_92_corrections_0_gt_polylen_5} \hspace{2mm} \includegraphics[width=0.496\linewidth,trim=0 30 0 94,clip]{figures/cross_domain/aerial/IMG_0574_11_stack_iou_0_88_corrections_0_gt_polylen_4}\\
	\includegraphics[width=0.496\linewidth,trim=0 174 0 0,clip]{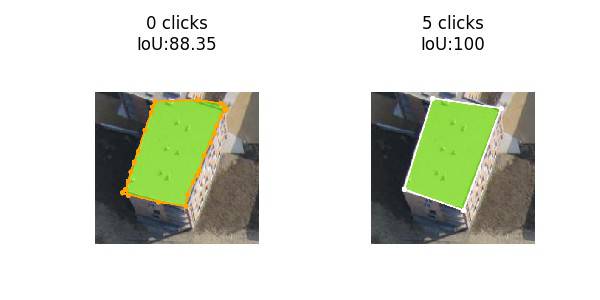} \hspace{2mm} 
	\includegraphics[width=0.496\linewidth,trim=0 174 0 0,clip]{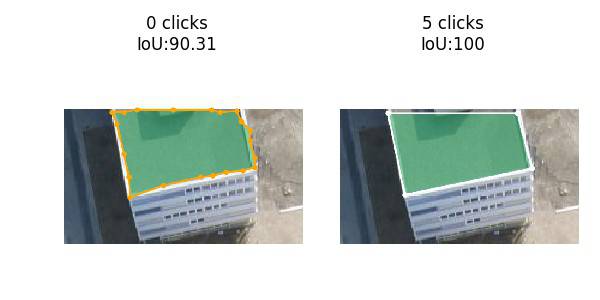}\\
		\includegraphics[width=0.496\linewidth,trim=0 30 0 60,clip]{figures/cross_domain/aerial/IMG_0688_1_stack_iou_0_88_corrections_0_gt_polylen_5} \hspace{2mm} 
	\includegraphics[width=0.496\linewidth,trim=0 30 0 74,clip]{figures/cross_domain/aerial/IMG_0653_6_stack_iou_0_90_corrections_0_gt_polylen_5}\\[-4mm]
\caption{{\bf Cross-domain {\color{magenta}{without}} fine-tuning: trained on Cityscapes $\rightarrow$ tested on Rooftop-Aerial, automatic mode}. Qualitative comparison between a human annotator and PolygonRNN++ in automatic mode in the Rooftop-Aerial dataset \textbf{without fine-tuning}.  Note that our model relies on ground-truth bounding boxes. }
\label{fig:human_auto_aerial}
\end{figure*}

\begin{figure*}[ht]
	\includegraphics[width=0.496\linewidth,trim=0 30 0 0,clip]{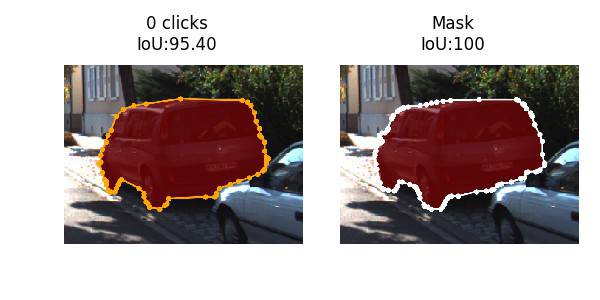} \hspace{2mm}
	\includegraphics[width=0.496\linewidth,trim=0 30 0 0,clip]{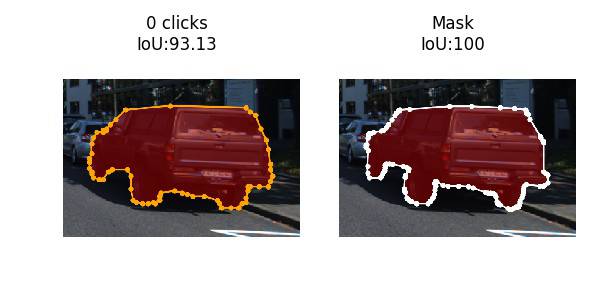}\\
	\includegraphics[width=0.496\linewidth,trim=0 30 0 0,clip]{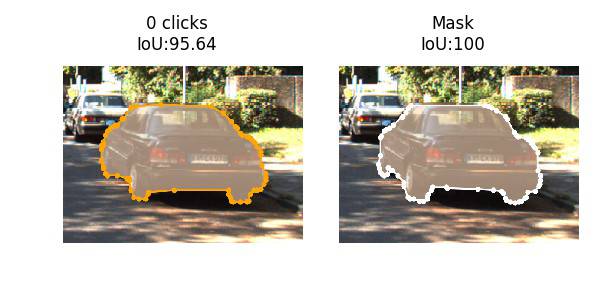} \hspace{2mm} %
	\includegraphics[width=0.496\linewidth,trim=0 30 0 0,clip]{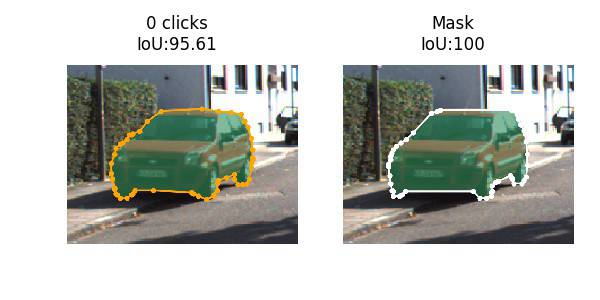} 
\caption{{\bf Cross-domain {\color{magenta}{without}} fine-tuning: trained on Cityscapes $\rightarrow$ tested on KITTI, automatic mode}. Qualitative comparison between a human annotator and PolygonRNN++ in automatic mode in KITTI \textbf{without fine-tuning}.  Note that our model relies on ground-truth bounding boxes. }
\label{fig:human_auto_kitti}
\end{figure*}

\begin{figure*}[ht]
\vspace{-3mm}
	\includegraphics[width=0.496\linewidth,trim=30 30 15 0,clip]{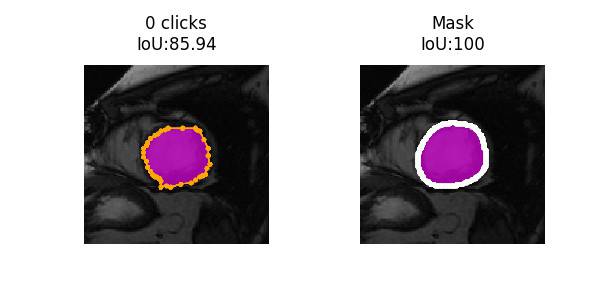} \hspace{2mm}
	\includegraphics[width=0.496\linewidth,trim=30 30 15 0,clip]{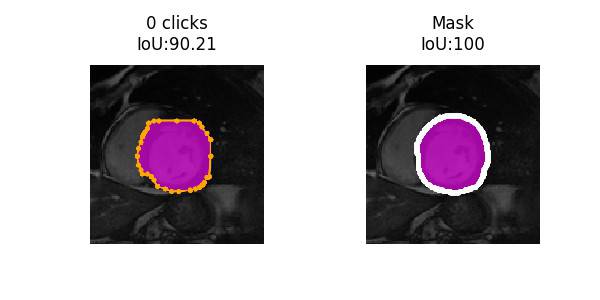}\\
	\includegraphics[width=0.496\linewidth,trim=30 30 15 0,clip]{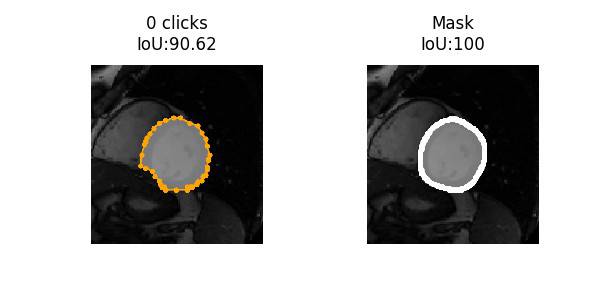} \hspace{2mm} \includegraphics[width=0.496\linewidth,trim=30 30 15 0,clip]{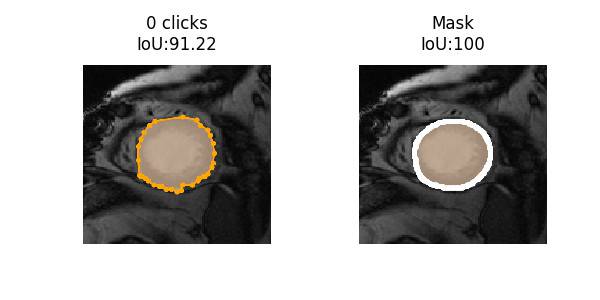}\\[-7mm]
\caption{{\bf Cross-domain {\color{magenta}{without}} fine-tuning: trained on Cityscapes $\rightarrow$ tested on Cardiac MR, automatic mode}. Qualitative comparison of ground-truth masks vs PolygonRNN++ in automatic mode in the Cardiac MR dataset \cite{medical1,medical1b} \textbf{without fine-tuning}.  Note that our model relies on bounding boxes.}
\label{fig:human_auto_medical1}
\end{figure*}

\begin{figure*}[ht]
	\includegraphics[width=0.496\linewidth,trim=20 30 15 0,clip]{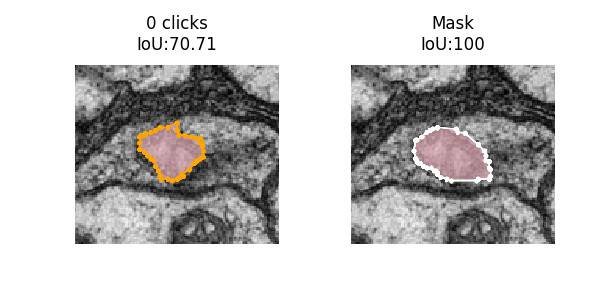} \hspace{2mm}
	\includegraphics[width=0.496\linewidth,trim=20 30 15 0,clip]{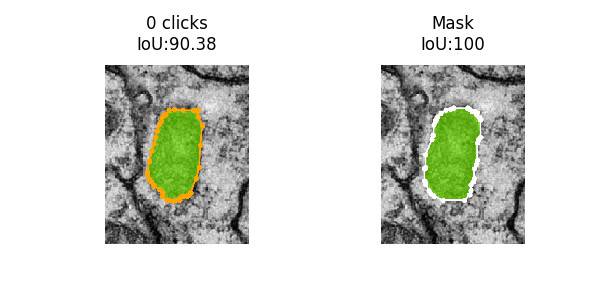}\\[-1mm]
	\includegraphics[width=0.496\linewidth,trim=20 30 15 0,clip]{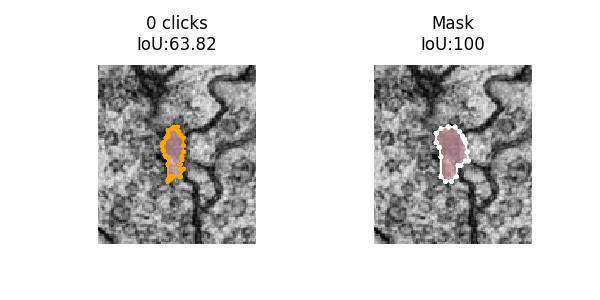} \hspace{2mm}
	\includegraphics[width=0.496\linewidth,trim=20 30 15 0,clip]{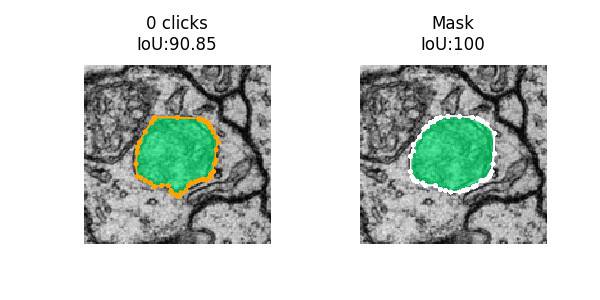}\\[-1mm]
	\includegraphics[width=0.496\linewidth,trim=20 30 15 0,clip]{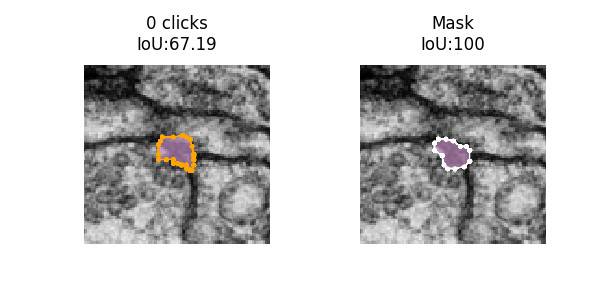} \hspace{2mm} \includegraphics[width=0.496\linewidth,trim=20 30 15 0,clip]{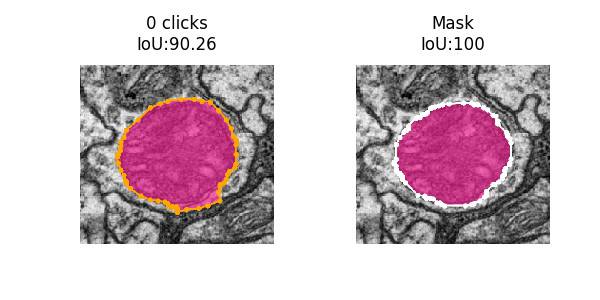}\\[-6mm]
\caption{{\bf Cross-domain {\color{magenta}{without}} fine-tuning: trained on Cityscapes $\rightarrow$ tested on ssTEM, automatic mode}. Qualitative comparison of ground-truth masks vs PolygonRNN++ in automatic mode in the ssTEM dataset\cite{medical2} \textbf{without fine-tuning}.  Note that our model relies on ground-truth bounding boxes. }
\label{fig:human_auto_medical2}
\end{figure*}

\begin{figure*}[ht]
	\includegraphics[width=0.496\linewidth,trim=15 30 0 0,clip]{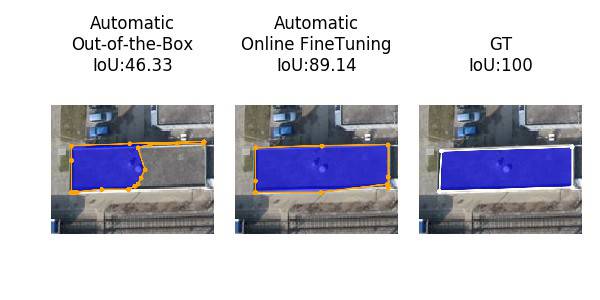} \hspace{2mm}
	\includegraphics[width=0.496\linewidth,trim=15 30 0 0,clip]{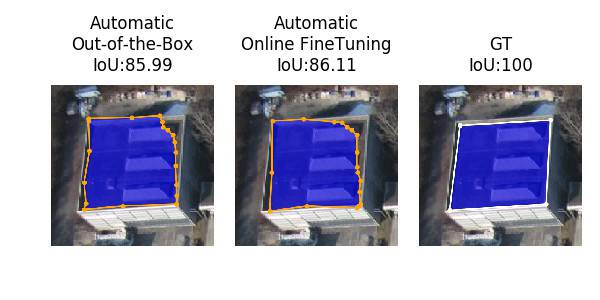}\\
	\includegraphics[width=0.496\linewidth,trim=15 30 0 0,clip]{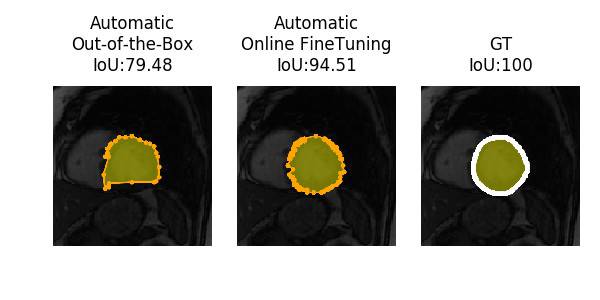} \hspace{2mm}
	\includegraphics[width=0.496\linewidth,trim=15 30 0 0,clip]{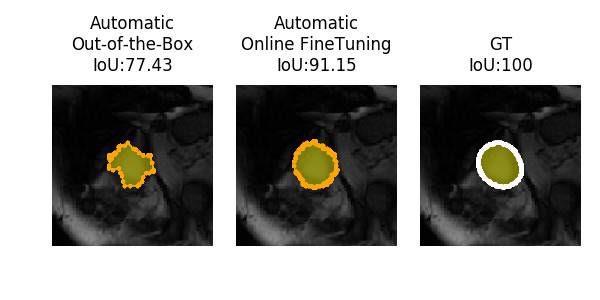}\\
	\includegraphics[width=0.496\linewidth,trim=15 30 0 0,clip]{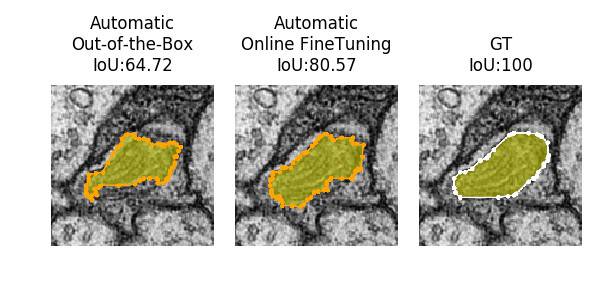} \hspace{2mm} \includegraphics[width=0.496\linewidth,trim=15 30 0 0,clip]{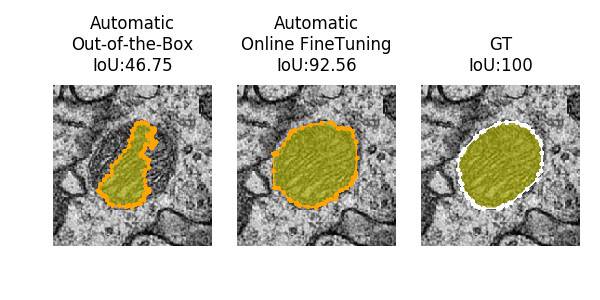}\\
	\includegraphics[width=0.496\linewidth,trim=15 30 0 0,clip]{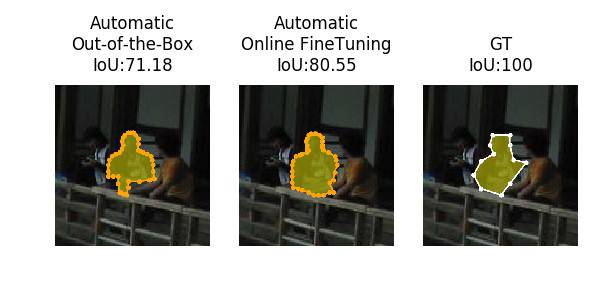} \hspace{2mm} \includegraphics[width=0.496\linewidth,trim=15 30 0 0,clip]{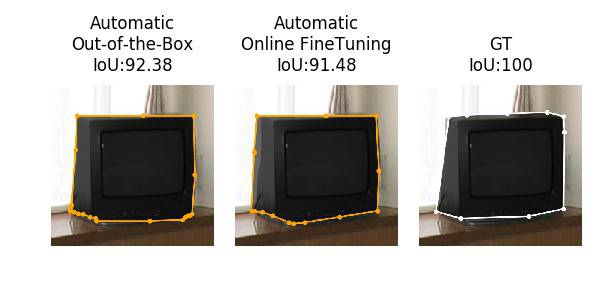}\\
	\includegraphics[width=0.496\linewidth,trim=15 30 0 0,clip]{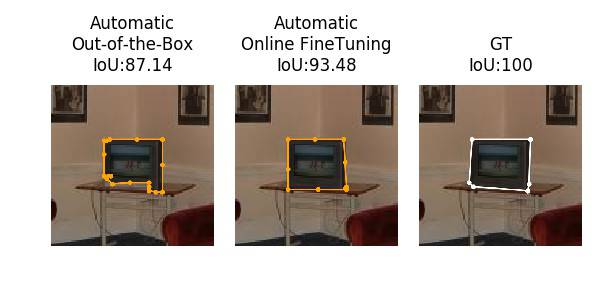} \hspace{2mm} \includegraphics[width=0.496\linewidth,trim=15 30 0 0,clip]{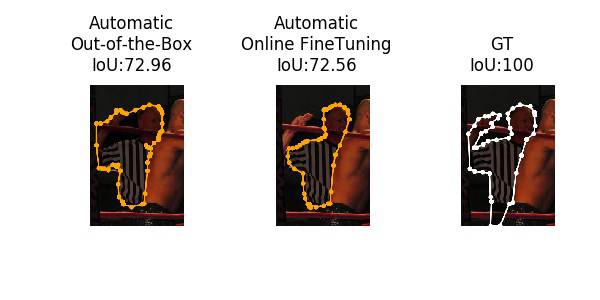}\\[-5mm]
\caption{{\bf Cross-domain {\color{magenta}{with}} fine-tuning, automatic mode}. Qualitative results on different out-of-domain datasets after using the proposed Online Fine-tuning algorithm.}
\label{fig:online_finet}
\end{figure*}

\paragraph{Extended Evaluation on ADE Val Set.} In Figure~\ref{fig:full_ade_plots}, we report the performance of PolygonRNN++ on the ADE validation set without any fine-tuning, and running in automatic mode (with ground-truth boxes). Note that, for the ease of visualization we only illustrate the top and bottom 50 categories, sorted by performance. Only instances with more than 20 categories are shown.

\begin{table}[t!]
\begin{center}
\small
\begin{tabular}{|l|c|c|}
\hline
Model & AP & AP 50 \% \\
\hline 
PANet & 36.4 & 63.1 \\
Mask R-CNN & 32.0 &	58.1 \\
SegNet & 29.5 & 55.6 \\
GMIS & 27.6 & 44.6 \\
\hline 
{\bf PolygonRNN++} & 25.5 & 45.5 \\
\hline
SGN	& 25.0 &	44.9	\\
\hline
\end{tabular}
\vspace{0.1mm}
\caption{Performance on official Cityscapes Instance Labeling benchmark (test). We report best result for each method.}
\label{tbl_performance_cityscapes_test}
\end{center}
\vspace{-2mm}
\end{table}

\begin{figure*}[ht]
	\includegraphics[width=0.496\linewidth,trim=15 30 0 0,clip]{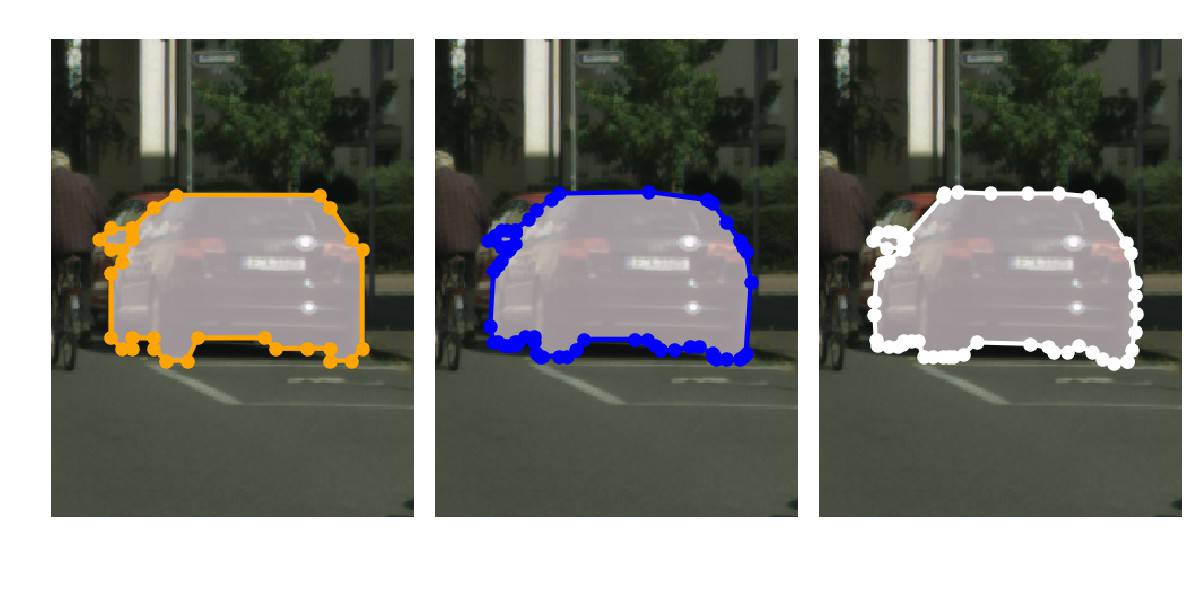} \hspace{2mm}
	\includegraphics[width=0.496\linewidth,trim=15 30 0 0,clip]{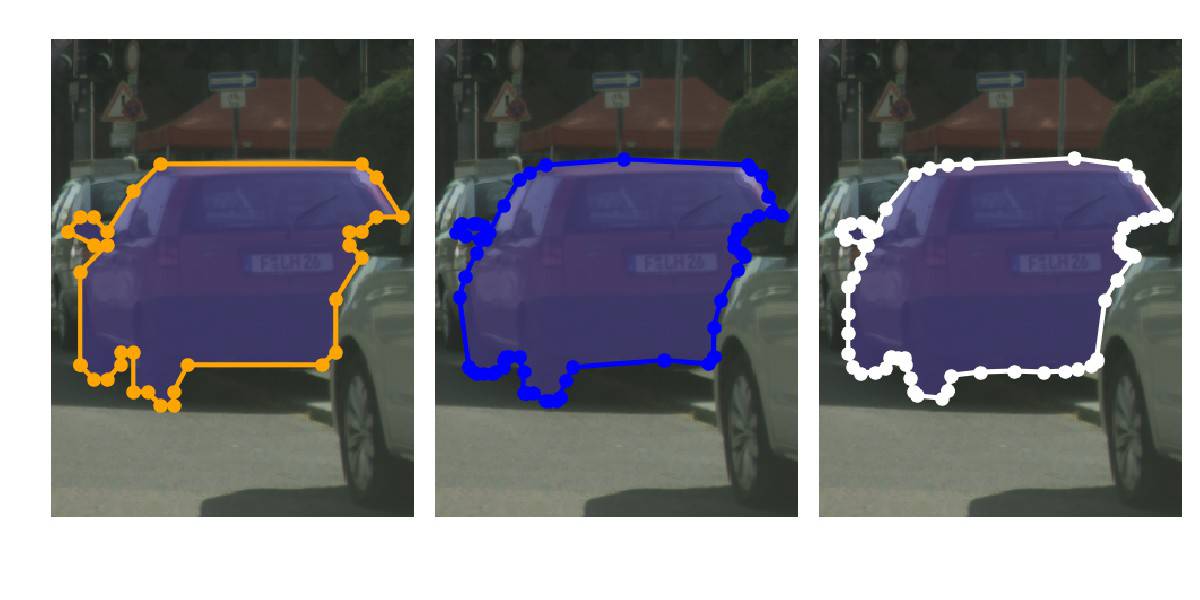}\\
	\includegraphics[width=0.496\linewidth,trim=15 30 0 0,clip]{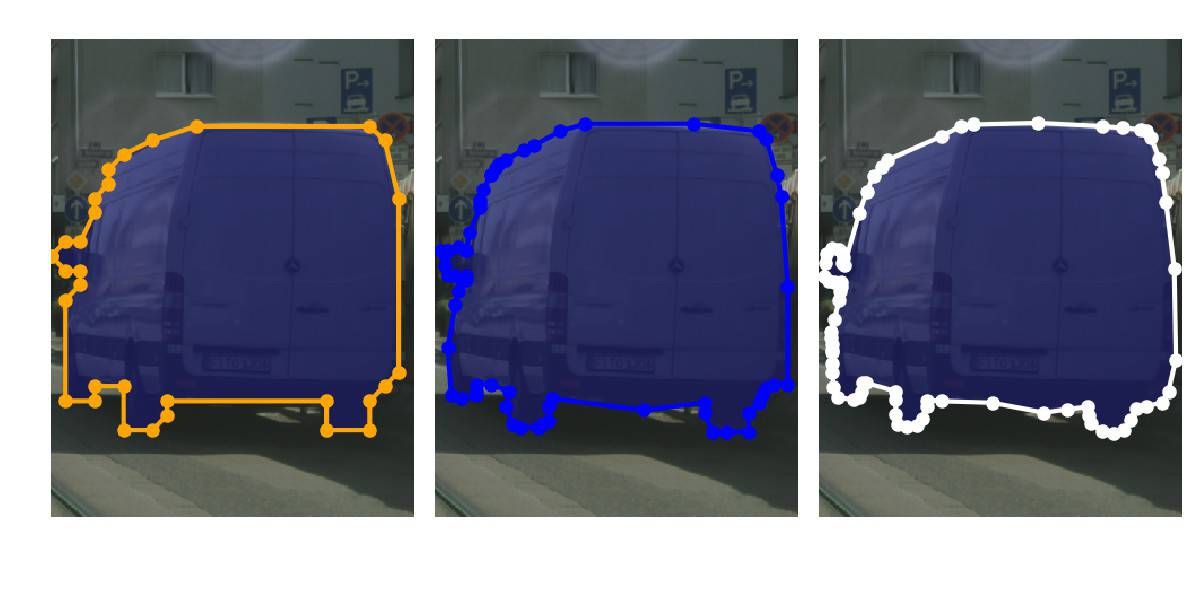} \hspace{2mm}
	\includegraphics[width=0.496\linewidth,trim=15 30 0 0,clip]{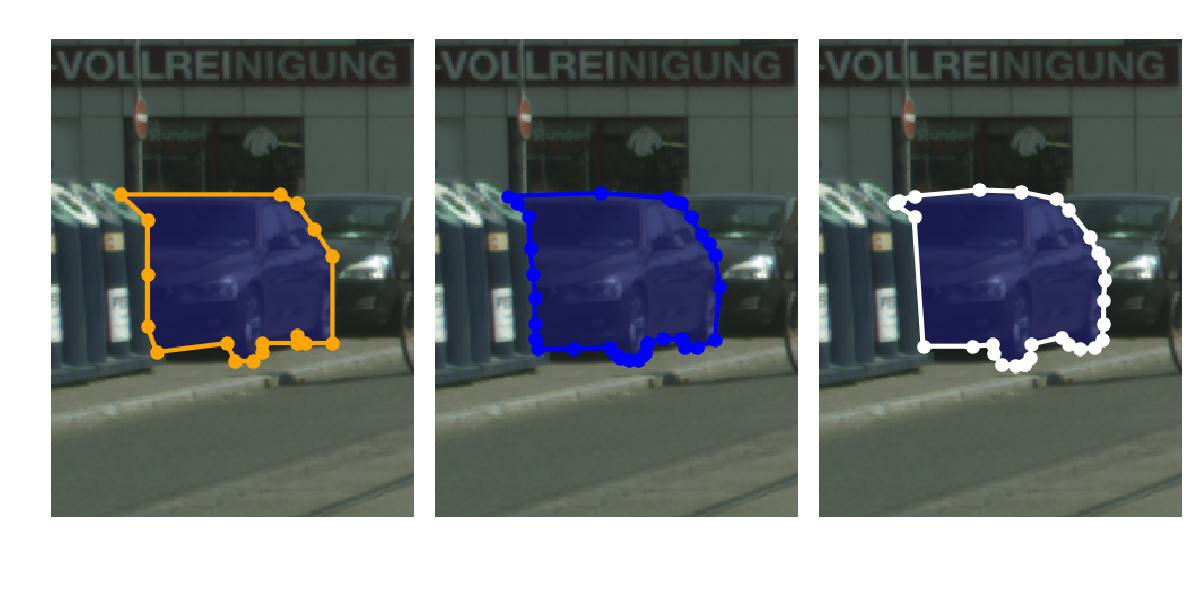}\\
	\includegraphics[width=0.496\linewidth,trim=15 30 0 0,clip]{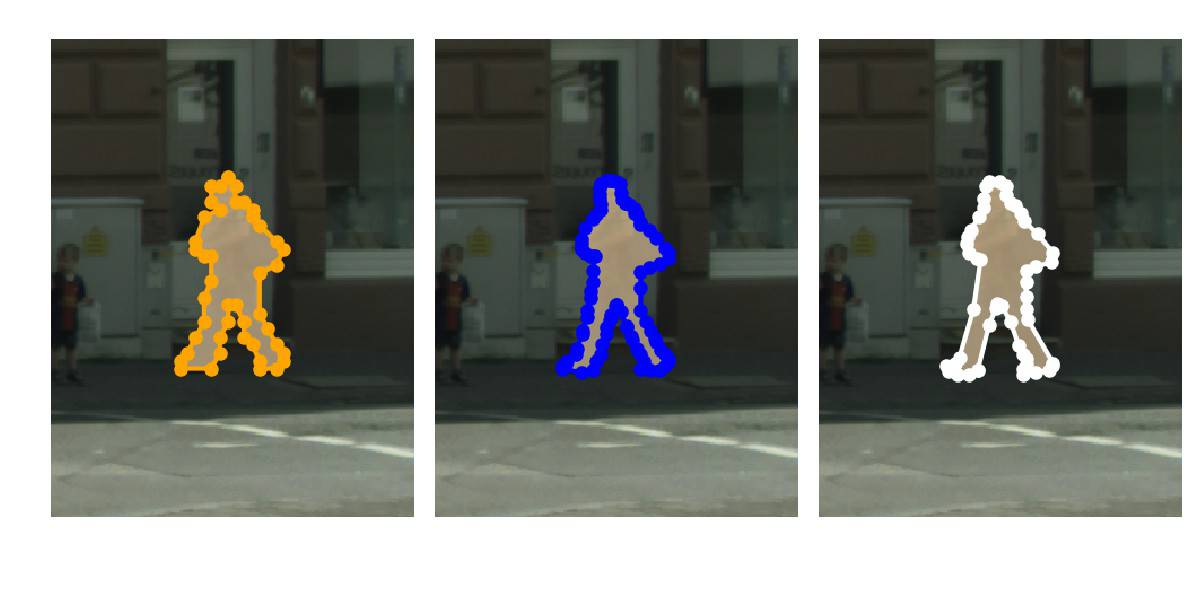} \hspace{2mm}
	\includegraphics[width=0.496\linewidth,trim=15 30 0 0,clip]{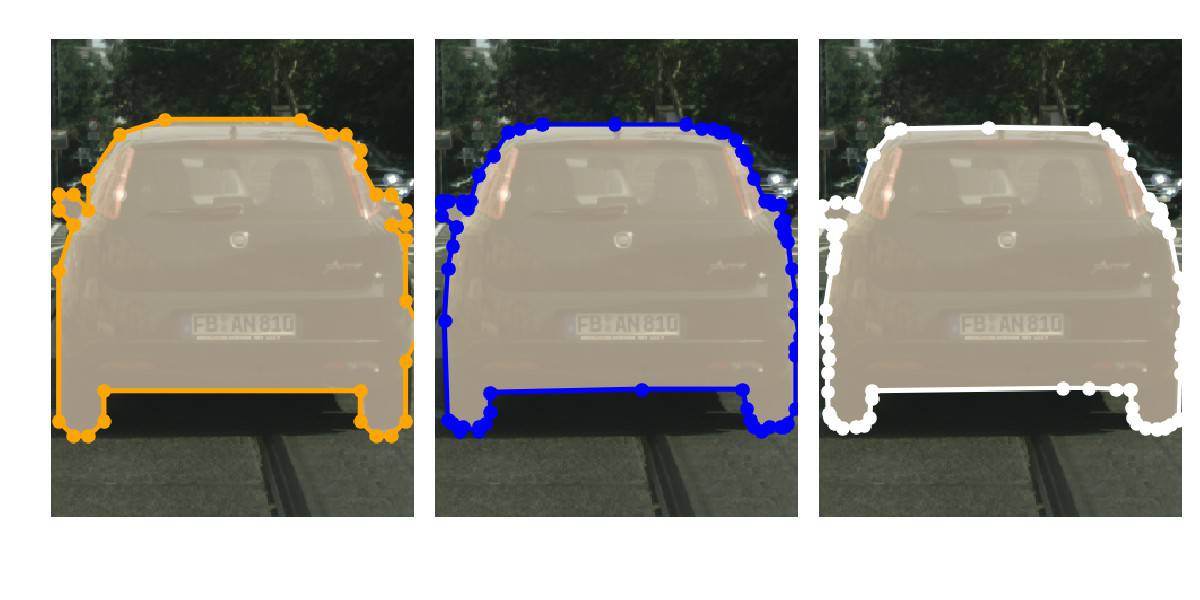}\\
	\includegraphics[width=0.496\linewidth,trim=15 30 0 0,clip]{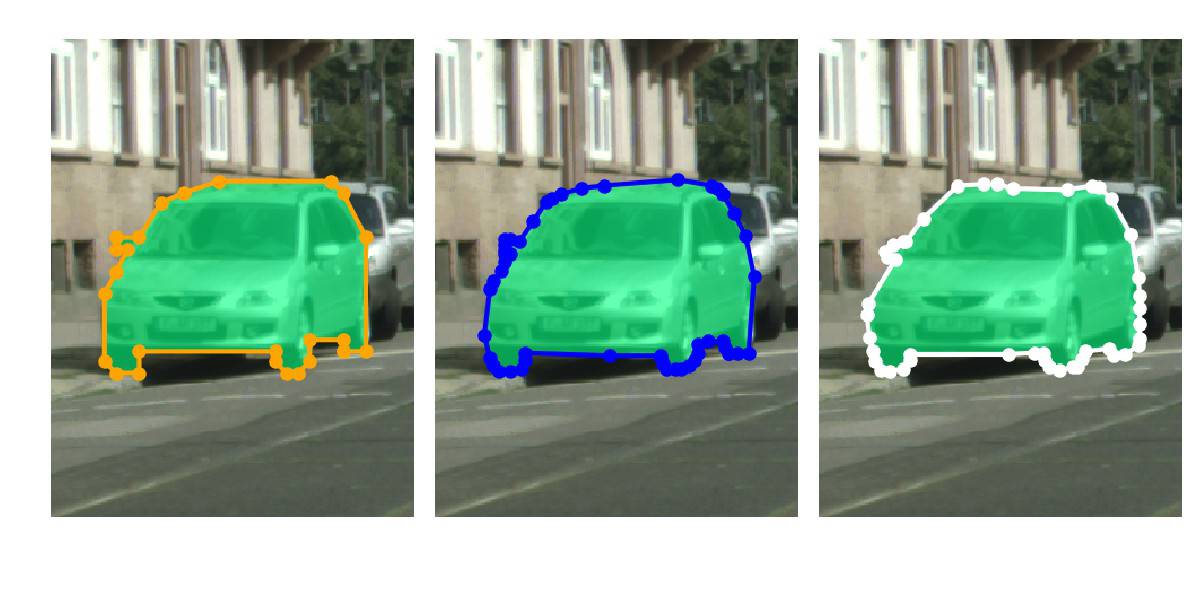} \hspace{2mm}
	\includegraphics[width=0.496\linewidth,trim=15 30 0 0,clip]{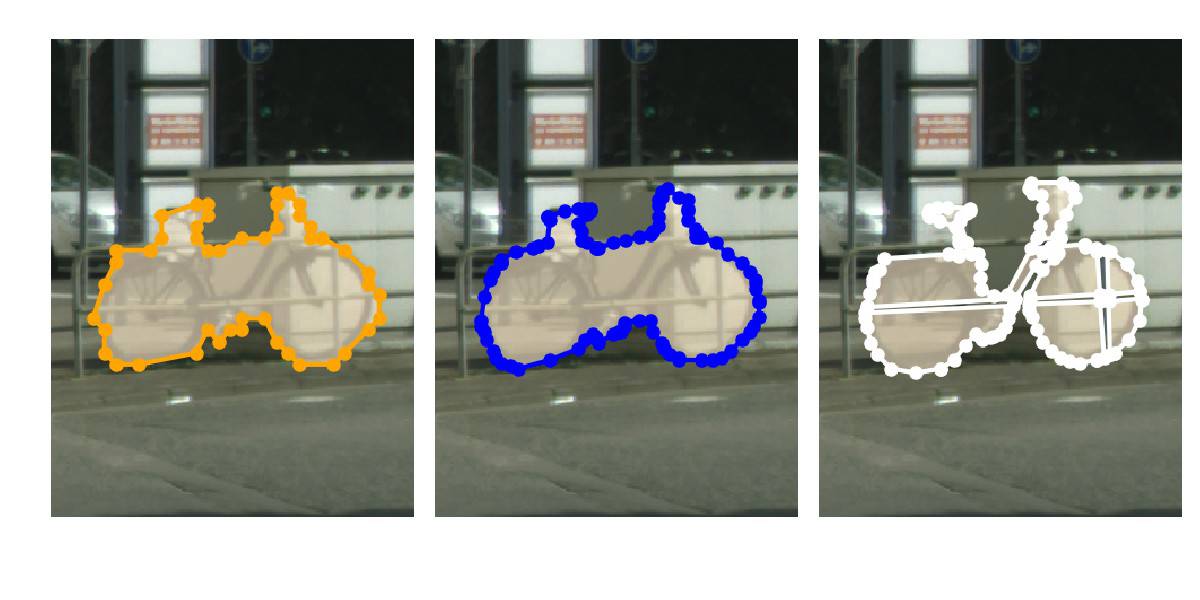}\\
	\includegraphics[width=0.496\linewidth,trim=15 30 0 0,clip]{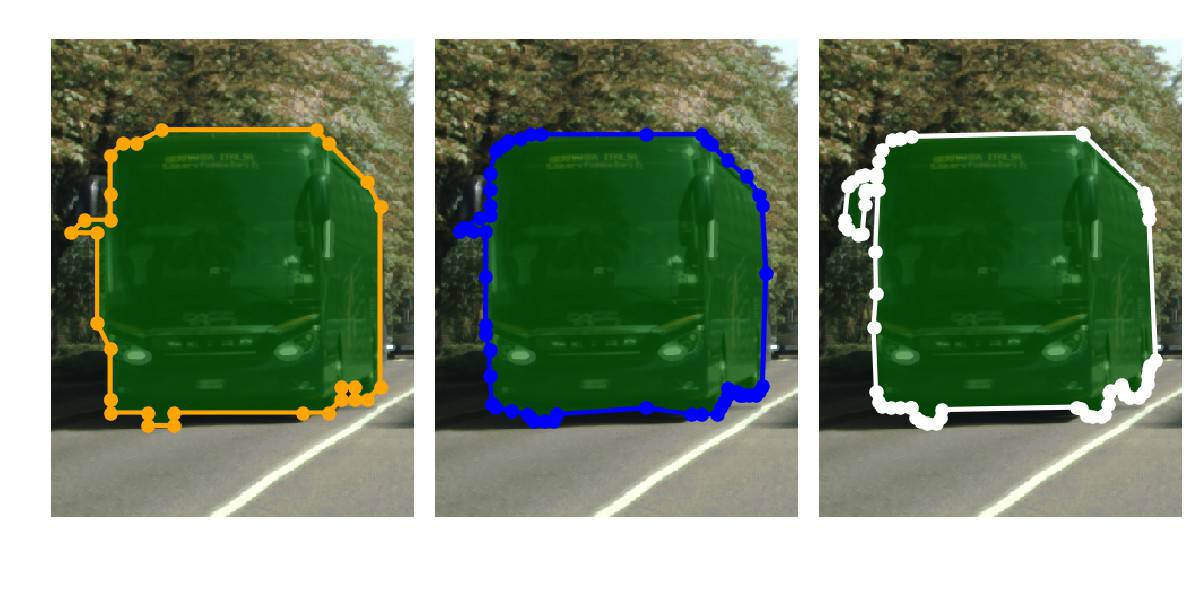} \hspace{2mm}
	\includegraphics[width=0.496\linewidth,trim=15 30 0 0,clip]{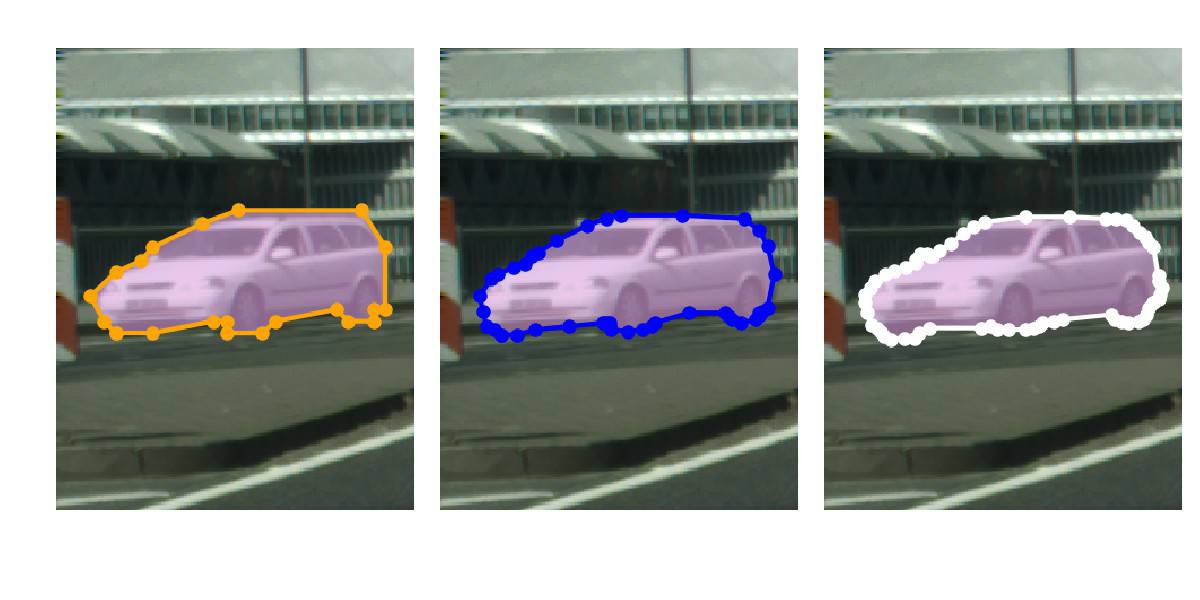}\\[-5mm]
\caption{{\bf Upscaling with the GGNN: } Performance of PolygonRNN++ after and before upscaling the output polygon with GGNN \textbf{Left:} Output of PolygonRNN++ before upscaling  \textbf{Center:} Output of PolygonRNN++ after GGNN \textbf{Right:} GT}
\label{fig:ggnn_role}
\end{figure*}

\end{document}